\PassOptionsToPackage{numbers, compress}{natbib}
\documentclass{article}

\usepackage[main, final]{neurips_2026}

\usepackage[utf8]{inputenc} 
\usepackage[T1]{fontenc}    
\usepackage{hyperref}       
\usepackage{url}            
\usepackage{booktabs}       
\usepackage{amsfonts}       
\usepackage{nicefrac}       
\usepackage{microtype}      
\usepackage{xcolor}         
\usepackage{multirow}
\usepackage{graphicx}
\usepackage{subcaption}
\usepackage{algorithm}
\usepackage{algpseudocode}
\usepackage{placeins}
\usepackage{wrapfig}
\usepackage{multirow}
\usepackage{amsmath} 
\usepackage{ulem}
\usepackage[dvipsnames, table]{xcolor}
\usepackage{xcolor} 
\usepackage{pifont}

\newcommand{\best}[1]{\underline{\textbf{#1}}}
\newcommand*{\doi}[1]{\href{http://dx.doi.org/#1}{doi: #1}}

\definecolor{mydarkgreen}{rgb}{0.0, 0.5, 0.0}
\definecolor{mydarkred}{rgb}{0.75, 0.0, 0.0}

\newcommand{\cmark}{\textcolor{mydarkgreen}{\ding{51}}}
\newcommand{\xmark}{\textcolor{mydarkred}{\ding{55}}}

\newcommand{\nt}[1]{{\color{black}#1}}

\title{Geometry Meets Physics: Data-Efficient Pre-Training for Unstructured Neural PDE Solvers}

\author{%
  \textbf{Luis Medrano-Navarro}$^1$ \quad
  \textbf{Giacomo Baldan}$^1$ \quad
  \textbf{Qiang Liu}$^1$ \quad
  \textbf{Benjamin Holzschuh}$^1$ \\
  \textbf{Jan Hagnberger}$^2$ \quad
  \textbf{Mathias Niepert}$^2$ \quad
  \textbf{Nils Thuerey}$^1$ \\[0.6em]
  $^1$Technical University of Munich \qquad $^2$University of Stuttgart \\
  \texttt{\{luis.medrano, nils.thuerey\}@tum.de}
}

\begin{document}

\maketitle

\begin{abstract}
  Neural surrogate models for Partial Differential Equations (PDEs) on unstructured 3D geometries are often limited by poor generalization and the high cost of generating large-scale training datasets. Consequently, pre-training on massive datasets of related PDE dynamics has emerged as a critical alternative to enhance the robustness and scalability of these models. However, this strategy is neither compute- nor data-efficient, as it relies on massive pre-computed data that is very costly to generate.
  In this work, we introduce a disk-data-free pre-training framework tailored to both steady-state and transient regimes. For steady-state problems, we propose a geometry-driven strategy that leverages intrinsic shape descriptors to learn representations of complex 3D domains. For transient problems, we introduce a physics-driven approach based on online generation of synthetic PDE data, enabling scalable pre-training without reliance on expensive datasets. Across multiple experiments, our approach achieves faster convergence, greater data efficiency, and higher accuracy during fine-tuning, particularly under realistic low-data regimes. This methodology provides a practical pathway toward data-efficient neural emulators for large-scale simulations. Code and models are available at \url{https://github.com/tum-pbs/GMP}.
\end{abstract}

\section{Introduction}\label{introduction}

Engineering workflows have long depended on high-fidelity numerical solvers to resolve intricate physical systems~\citep{cfd1, particle, grid, cfd2}. Nevertheless, their immense computational overhead severely limits their feasibility in time-sensitive settings, iterative design exploration, and closed-loop control. Data-driven surrogate models provide a compelling pathway toward high-throughput inference~\citep{dl4p_mario, dl4p2, dl4p3,pbdl_book}. Yet, practical adoption of these emulators remains critically constrained by an underlying bottleneck: their performance hinges on massive, pre-computed simulation datasets whose generation is itself computationally prohibitive.
Motivated by advances in Large Language Models ~\citep[LLMs]{llms1, llms2}, foundation models for PDEs have adapted pre-training and fine-tuning paradigms to physical systems~\citep{tadpole, foundation_physix, foundation_poseidon, foundation_walrus, dpot}. 
Yet, these approaches still rely heavily on extensive offline datasets and predominantly target structured grids without geometric variation. Although recent work on unstructured meshes explores geometric pre-training~\citep{geopt, cheap_geometry}, capturing the coupled dynamics of geometry and physics remains challenging without costly labeled simulations. This motivates our central question: \textit{to what extent can \nt{unstructured} pre-training be realized efficiently without any precomputed simulation data?}

We categorize unstructured physical problems into two regimes with distinct dynamics: \textit{steady-state} and \textit{transient}. Steady-state problems correspond to equilibrium solutions where physical fields, e.g., surface pressure and shear stress in aerodynamics, are fully dictated by the geometry, boundary conditions and operating conditions. Consequently, pre-training purely on geometric descriptors can yield representations directly transferable to downstream flow prediction. In contrast, transient simulations evolve states autoregressively, with dynamics conditioned jointly on prior physical states in addition to geometric information. This distinction motivates specialized pre-training strategies for each regime to capture structural and physical inductive biases. 

To circumvent the prohibitive, often intractable burden of generating massive pre-computed datasets, we develop an entirely 
storage-free 
pre-training paradigm that constructs supervisory data on-the-fly. For steady-state regimes, we generate random assemblies of geometric primitives
online and train the model to reconstruct essential differential descriptors, including surface normals, curvature, and Vector Distance Fields (VDF). For transient autoregressive settings, we avoid offline trajectory storage by generating solutions to simple, canonical PDEs online during training. Together, these online strategies instill robust geometric and physical inductive biases at reduced computational overhead, bypassing the prohibitive cost of large-scale offline simulation repositories.

 Our main contributions are: 
(\textbf{I}) We present \textit{Geometry Meets Physics} (\textit{GMP}), a modular pre-training framework with two distinct variants for steady-state and transient physics simulations on unstructured meshes that bypasses the need to store precomputed PDE solutions and expensive CAD-level geometries. 
(\textbf{II}) We demonstrate its effectiveness to enable faster convergence, data savings and superior accuracy during fine-tuning, particularly in realistic low-data regimes. Additionally, we show that it is inherently architecture-agnostic.
(\textbf{III}) We propose \textit{GMP-Net}, a modified SOTA architecture specifically designed to efficiently pre-train for both transient and steady-state problems and to boost performance.
In summary, Table~\ref{tab:features_strategies} shows the key structural advantages of GMP. Relative to prior work, GMP requires zero offline data storage, dramatically accelerates pre-training time and supports multiple downstream tasks from a single pre-training stage.

\begin{figure}[t]
     \centering
     \begin{subfigure}[c]{0.49\textwidth}
         \centering
         \includegraphics[width=\textwidth]{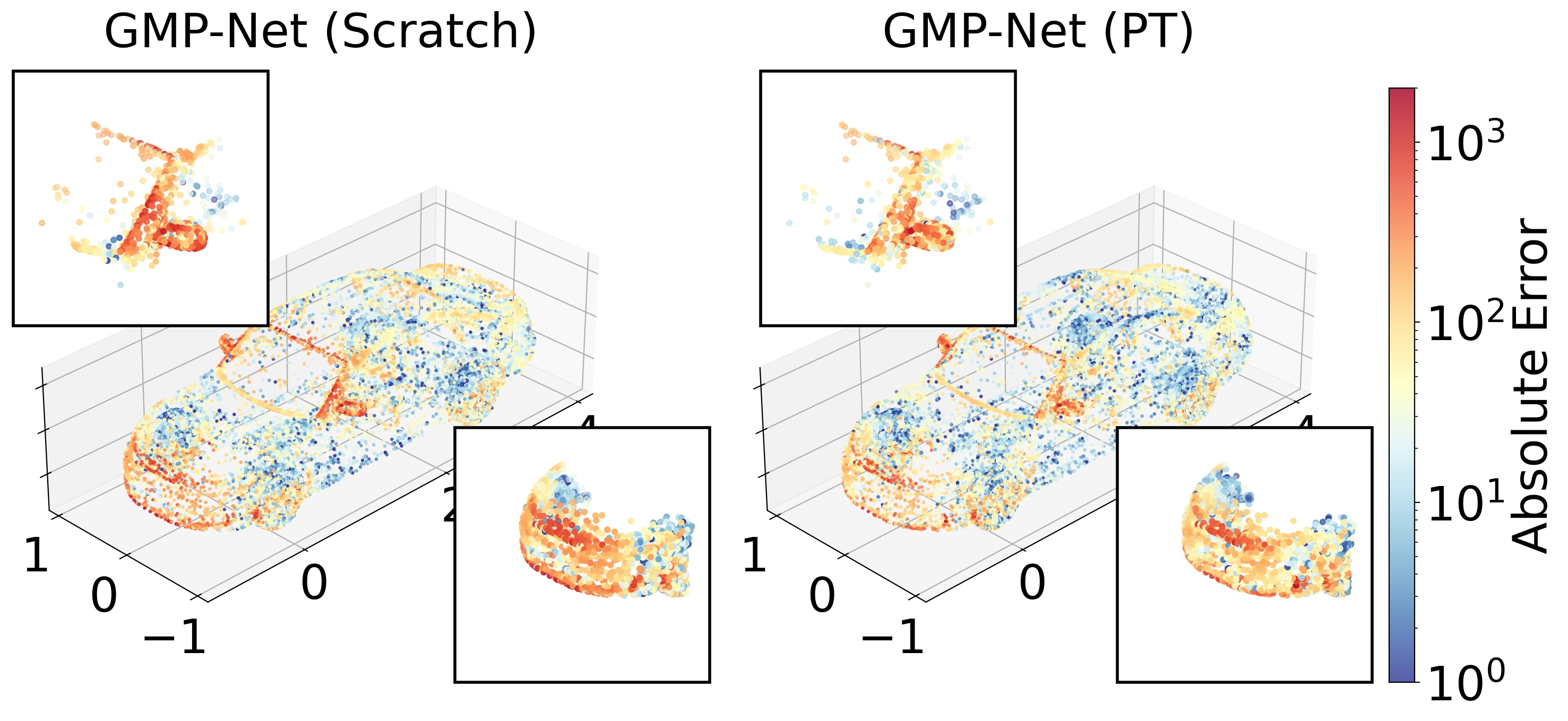}
         \caption{DrivAerML surface pressure errors}
         \label{fig:car_64_scatter}
     \end{subfigure}
     \hfill
     \begin{subfigure}[c]{0.49\textwidth}
         \centering
         \includegraphics[width=\textwidth]{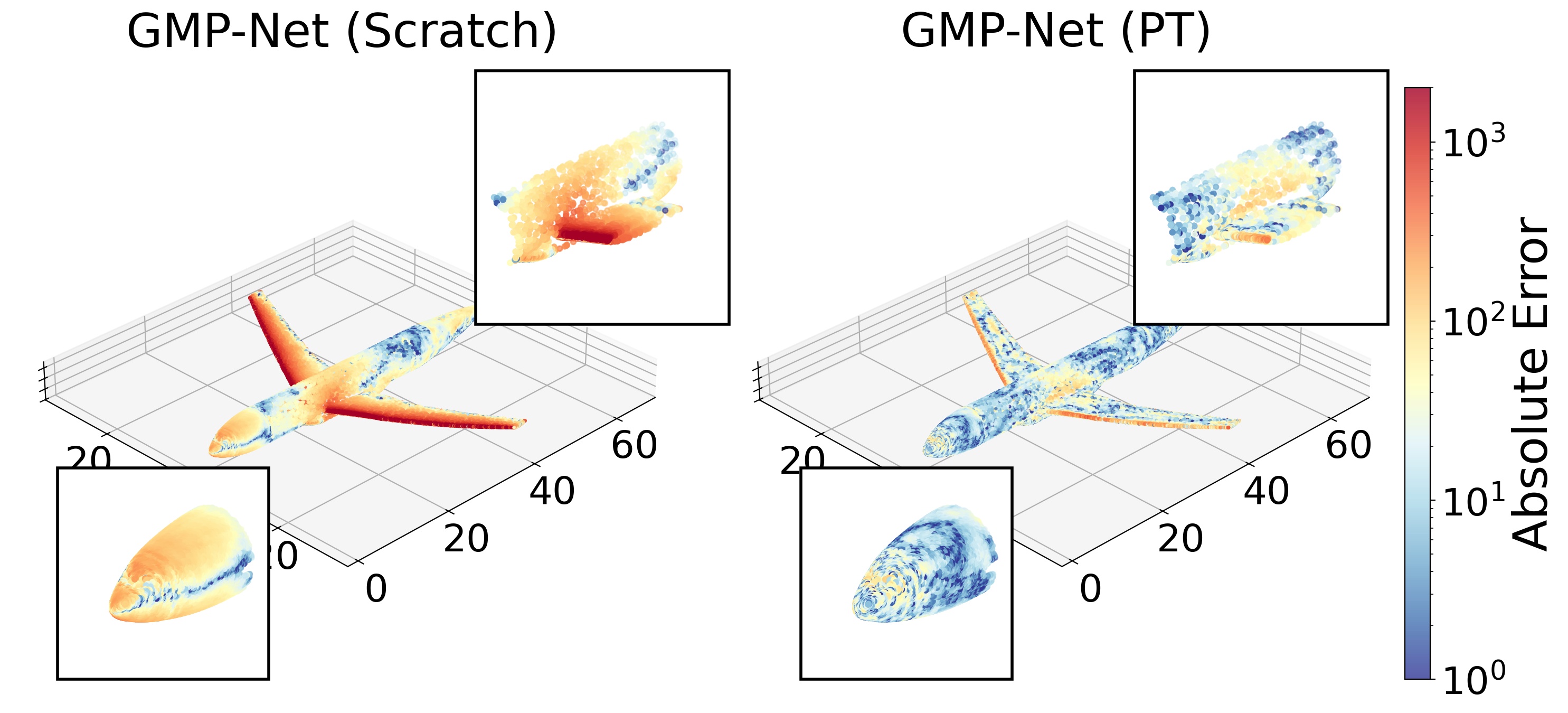}
         \caption{SHIFT-Wing surface pressure errors}
         \label{fig:wing_32_scatter}
     \end{subfigure}
     \caption{Point-wise absolute error comparison between trained from scratch and pre-trained models (PT). Our GMP pre-training shows increased accuracy in critical regions.}
     \label{fig:initial_plot}
\end{figure}

\begin{table}[t]
\centering
\caption{Comparison of pre-training paradigms across geometric and physical constraints.}
\label{tab:features_strategies}
\small
\begin{tabular}{lcccc}
\toprule
\textbf{Property} & \textbf{Traditional FM} & \textbf{\citet{cheap_geometry}} & \textbf{GeoPT} & \textbf{GMP (Ours)} \\
\midrule
Physics-label-free           & \xmark       & \cmark   & \cmark     & \cmark \\
Low pre-training cost     & \xmark & \cmark     & \xmark & \cmark \\
Unique pre-training       & \cmark     & \xmark     & \cmark     & \cmark \\
No high-detail CAD geometries    & \xmark       & \xmark     & \cmark     & \cmark \\
Zero disk storage  & \xmark & \xmark     & \xmark       & \cmark \\
\bottomrule
\end{tabular}
\end{table}

\section{Related work}\label{related_work}

\paragraph{Neural PDE solvers} Early efforts in neural physics surrogate modeling mainly employed convolutional neural networks (CNNs) over regular 
Cartesian grids~\citep{Thuerey_airfoil,fukami2019superres,wang2020tfnet}. 
Concurrently, architectures such as DeepONet~\citep{deeponet} and Fourier Neural Operators (FNO)~\citep{fno} emerged as alternative architectures with varying advantages and disadvantages. Motivated by the success of attention-based architectures in LLMs~\citep{attention_google} and computer vision (CV, ~\citep{vit, swin, dit}), recent research has adapted Transformers to PDEs, yielding architectures such as FactFormer~\citep{factformer}, scOT~\citep{foundation_poseidon}, and PDE-Transformer~\citep{pdetransformer, p3d}.
Some of these even include physics biases in the form of hard constraints, such as rotational equivariance \citep{revit, rotation_gnn}. 
Nonetheless, conventional CNNs and vision Transformers remain inherently constrained to Euclidean structured grids. To support irregular computational domains, Graph Neural Networks (GNNs) rapidly gained traction~\citep{meshgraphnets, gnnlino, gnn_original, gnncueto}. Early GNNs relied primarily on local message passing and frequently encountered scalability bottlenecks on large-scale 3D meshes comprising millions of vertices. Subsequent frameworks have addressed these constraints by combining graph-based topological structures with attention mechanisms~\citep{gnn_attention}, extending Transformers to unstructured point clouds and meshes, including OctFormer~\citep{octformer}, OFormer~\citep{oformer}, RIGNO~\citep{rigno}, UPT~\citep{upt}, and GIST~\citep{gist}. 
A persistent challenge is the quadratic complexity $\mathcal{O}(N^2)$ of standard attention, which becomes prohibitive for high-resolution 3D simulations. Recent models deploy diverse efficiency strategies: restricting attention to localized geometric neighborhoods (e.g., GAOT~\citep{gaot}, Erwin~\citep{erwin}, PTV3~\citep{ptv3}) or the use of subsampled anchor points e.g., AB-UPT~\citep{abupt} and SMART~\citep{smart}. In parallel, Transolver~\citep{transolver} and its adaptations, such as GeoTransolver~\citep{geotransolver}, address this scaling bottleneck by utilizing physics-informed sliced attention.

\paragraph{Pre-training physics surrogate models} While physics foundation models still trail progress in LLMs and CV, several recent frameworks have narrowed this gap~\citep{fluid_intelligence}. DPOT~\citep{dpot}, Poseidon~\citep{foundation_poseidon}, PhysiX~\citep{foundation_physix}, and Walrus~\citep{foundation_walrus} represent recent examples, scaling up to one billion parameters across domains ranging from astrophysics to plasma physics. Others, such as \citet{yunjia_fm}, explore particular applications in aerodynamic design.
However, data-driven physics foundation models still require massive precomputed datasets. While frameworks like Tadpole~\citep{tadpole} utilize online data generation via GPU-based spectral solvers, they remain restricted to regular grids. Furthermore, the approaches above are typically designed for autoregressive temporal rollouts rather than supporting direct \textit{geometry-to-physics} mappings, such as predicting steady-state aerodynamics around complex geometries. For this latter setting, recent research emphasizes leveraging intrinsic geometric features through specialized pre-training. For instance, the masked autoencoder (MAE) paradigm~\citep{masked_ae_clouds} has been adapted to physics~\citep{masked_ae_pde}, while GeoPT~\citep{geopt} introduced geometric pre-training via lifting. Similarly, others set up pre-train architectures to predict the occupancy of complex shapes without corresponding physics labels~\citep{cheap_geometry}. To date, no prior work has presented a modular framework for geometric and physics pre-training, nor demonstrated the effectiveness of storage-free online pre-training for unstructured simulations.

\section{Method}\label{method}

This section outlines our pre-training method. 
\nt{As a starting point we consider a unified formulation for steady-state and transient problems.} 
Let a spatial domain $\Omega(\mathcal{S}) \subset \mathbb{R}^n$ be defined by a geometry surface $\mathcal{S}$, and let $\mathbf{u}(\mathbf{x}, t) \in \mathbb{R}^m$ represent the physical target fields. The governing system obeys
\begin{equation}
\frac{\partial \mathbf{u}}{\partial t} + \mathcal{N}(\mathbf{x}, \mathbf{u}; \mathcal{S}, p, t) = 0,
\quad \mathbf{x} \in \Omega(\mathcal{S}), \quad t > 0,
\end{equation}
where $\mathcal{N}$ is a differential operator parametrized by the geometry $\mathcal{S}$ and global parameters $p$. If we discretize $\Omega(\mathcal{S})$ via a point cloud $\mathbf{x} = \{\mathbf{x}_i\}_{i=1}^N$, the discretized fields can now be written as $\mathbf{u}_t \in \mathbb{R}^{N \times m}$. We want a surrogate $f_\theta : (\mathbf{x}_t, \mathbf{u}_t, \mathcal{S}, p) \mapsto \mathbf{u}_{t+1}$ for the temporal evolution optimized via error minimization of trajectory data
\begin{equation}
\min_\theta \sum_{j,t} \big| f_\theta(\mathbf{x}_t^{(j)}, \mathbf{u}_t^{(j)}, \mathcal{S}^{(j)}, p^{(j)}) - \mathbf{u}_{t+1}^{(j)} \big|^2,
\end{equation}
over several available training samples. This formulation seamlessly encompasses steady-state regimes, where solutions $\mathbf{\bar u}(\mathcal{S}, p)$ correspond to fixed points satisfying  $\mathcal{N}(\mathbf{x}, \mathbf{\bar u}; \mathcal{S}, p) = \mathbf{0}$. By decoupling from past temporal states, the model reduces to the direct mapping $f_\theta : (\mathbf{x}, \textbf{0}, \mathcal{S}, p) \mapsto \mathbf{\bar u}$, trained via
\begin{equation}
\min_\theta \sum_j \big| f_\theta(\mathbf{x}^{(j)}, \textbf{0}, \mathcal{S}^{(j)}, p^{(j)}) - \mathbf{\bar u}^{(j)} \big|^2.
\end{equation}
\nt{For neural network architectures below, we distinguish geometry representations $\mathcal{S}$, surface physical fields $\mathbf{u}_{S}$ and volumetric physical fields $\mathbf{u}_{V}$.}

\begin{figure}[t]
     \centering
     \begin{subfigure}[c]{0.49\textwidth}
         \centering
         \includegraphics[height=3.8cm]{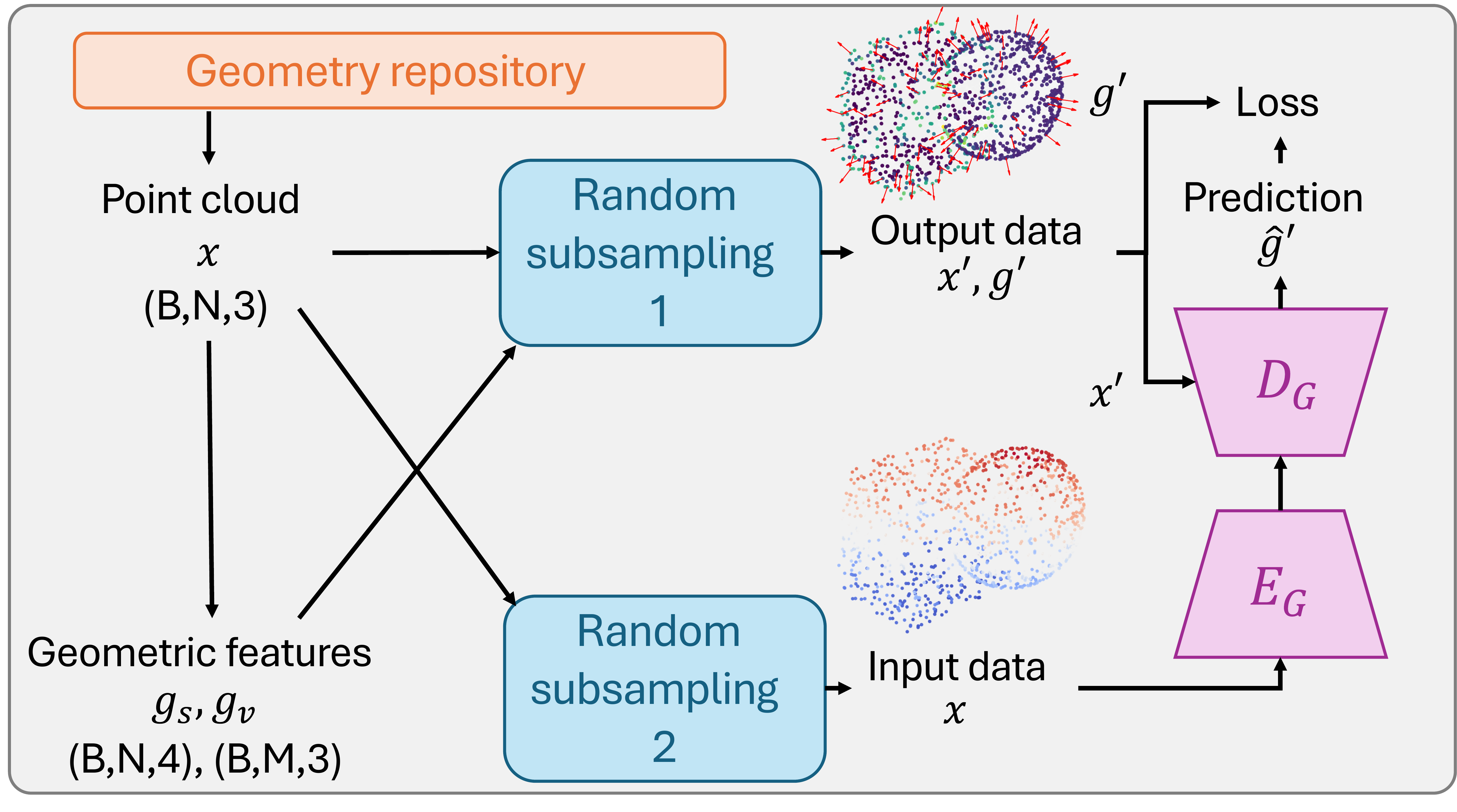}
         \caption{Geometric pre-training}
         \label{fig:sketch_geo}
     \end{subfigure}
     \hfill
     \begin{subfigure}[c]{0.49\textwidth}
         \centering
         \includegraphics[height=3.8cm]{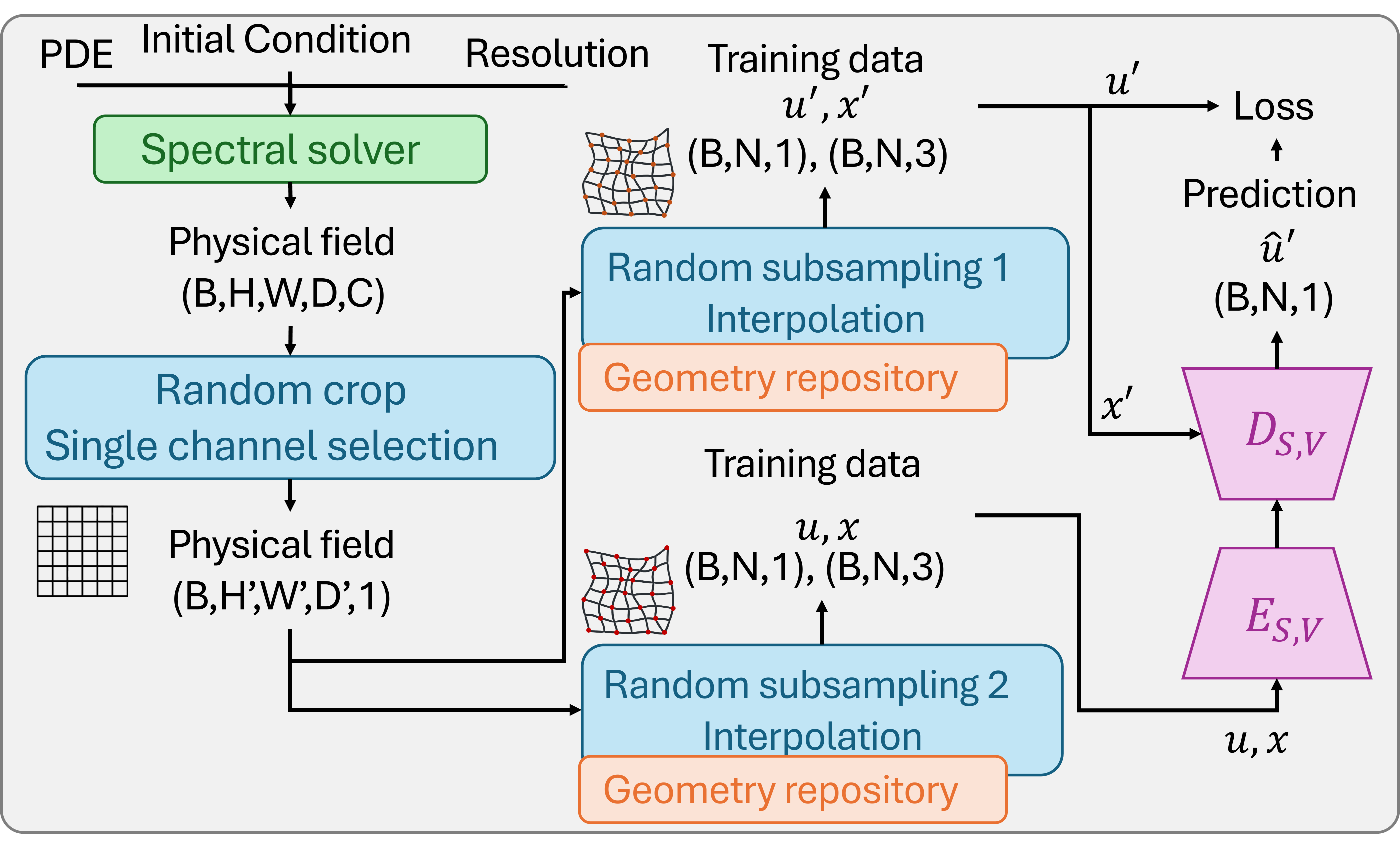}
         \caption{Physics pre-training}
         \label{fig:sketch_physics}
     \end{subfigure}
     \caption{Sketches for our two pre-training strategies. For steady predictions, we use geometric pre-training (left). The model learns to predict geometric features of randomized geometries, like the normals and curvature on the surface, and the Vector Distance Field (VDF) in the surrounding volume. For transient problems, we use physics pre-training (right). The model learns to autoencode online generated solutions to 
     canonical, synthetic PDEs at randomized locations. 
     }
     \label{fig:sketch_pretraining}
\end{figure}

\newpage
\subsection{Geometric pre-training}\label{geometric_pretraining}

For geometric pre-training we fully rely on online generated geometries and
construct randomized assemblies of \nt{cubes, spheres and tetrahedra as core primitives.} 
Mathematically, we define shapes $\mathcal{S}(\alpha) \subset \mathbb{R}^3, \quad \alpha \sim \mu(\alpha),$ where $\alpha \in \mathbb{R}^k$ are shape parameters and $\mu$ defines their sampling distribution. This construction yields a controlled distribution of geometries with varied, well-defined surface properties.
Each surface $\mathcal{S}(\alpha)$ is discretized as a point cloud $\{ {\mathbf{x}_S^{(i)}} \} \subset \mathcal{S}(\alpha)$, $i=1 \dots N$ and for each we compute a set of geometric descriptor fields
$g = \big( n, \kappa, v \big)$. Here, $n : \mathcal{S} \to \mathbb{S}^2$ is the unit normal field, $\kappa : \mathcal{S} \to \mathbb{R}$ denotes the curvature, and $v : \Omega \to \mathbb{R}^3$ is the Vector Distance Field (VDF) defined on a volumetric neighborhood $\Omega \supset \mathcal{S}$. In particular, as $\kappa$ we use the surface variation (or PCA curvature index~\citep{curvature_computation}). To probe volumetric structure, we sample query locations $\{\mathbf{x}_V^{(j)}\} \subset \Omega$ , $j=1 \dots M$ from a parametrized volumetric deformation within a bounding domain enclosing $\mathcal{S}(\alpha)$. This induces a joint learning problem over surface and volume representations.

For architectures with continuous queryable outputs, we formulate geometric pre-training as a masked operator regression problem: given a subset of locations $\mathbf{x}=(\mathbf{x}_S, \mathbf{x}_V)$, the model predicts descriptor values $g(\mathbf{x}')$ at disjoint query points $\mathbf{x}'$  with $\mathbf{x} \cap \mathbf{x}' = \emptyset$, via $\hat{g}'_\theta = D_\theta(E_\theta(\mathbf{x}), \mathbf{x}')$, where $E$ and $D$ represent the model's encoder and decoder. For architectures with fixed output positions, we drop $\mathbf{x}'$, and pre-training is formulated in the same fixed positions $\hat{g}'_\theta = NN_\theta (\mathbf{x})$. This objective enforces the learning of a continuous latent representation with the model's encoder $z_\theta = E_\theta(\mathbf{x})$ approximating the underlying geometric fields across both $\mathcal{S}$ and $\Omega$. The corresponding learning objective is 
\begin{equation}
    \mathcal{L} = \omega_{\kappa} \mathcal{L}_{\kappa} + \omega_n \mathcal{L}_\text{n} + \omega_V \mathcal{L}_\text{V},
\end{equation} 
where each component of the loss is $\mathcal{L}_i=||\hat{g}'_{i, \theta} - g'_i||$, and $\omega_i$ denote weights for each loss term. To avoid conflicting directions between the three gradients, we use ConFIG~\citep{config} as our optimization algorithm (more details in Appendix~\ref{app:ablation_studies}).

\begin{wrapfigure}[6]{r}{0.29\textwidth}
\vspace{-23pt}
  \begin{center}
    \includegraphics[width=0.29\textwidth]{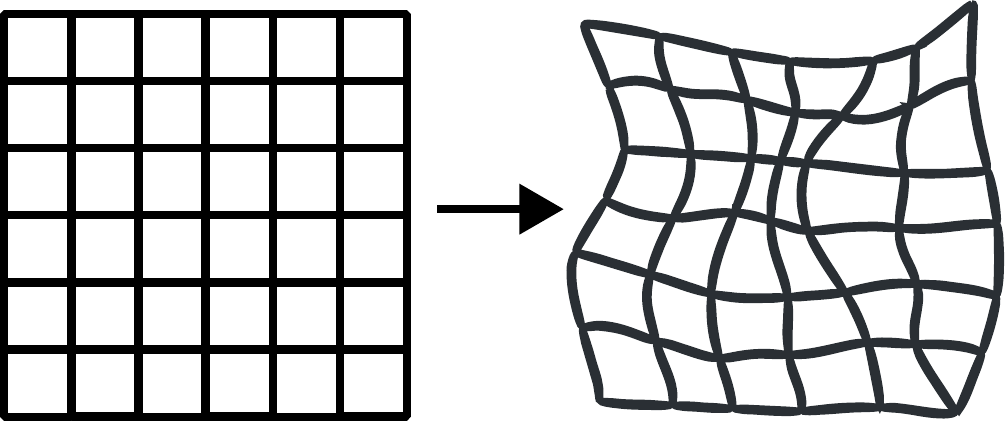}
  \end{center}
  \caption{Deformed grid.}
  \label{fig:3d_shapes}
\end{wrapfigure}

The sketch of geometric pre-training is provided in Figure \ref{fig:sketch_geo}. For the volumetric discretization of $\Omega \supset \mathcal{S}$, we use smooth, and intersection-free Gaussian deformed grids (Figure \ref{fig:3d_shapes}).
To do so, we build Gaussian displacements $\mathbf{d}(\mathbf{x})$ (Appendix~\ref{app:gaussianDefo}) and then $\Omega(\mathbf{x},\alpha) = \mathbf{x} + \alpha \mathbf{d}(\mathbf{x})$ with $\mathbf{x}$ sampled from regular grids, and $\alpha \le 1$.

While previous approaches have considered engineering-level CAD designs \citep{cheap_geometry} or lifted trajectories \citep{geopt} requiring large amounts of storage, GMP offers significant advantages in versatility and data efficiency, with zero data storage needed and everything generated on-the-fly. Additionally, we show below that including advanced surface features like normals or curvature proves to be critical to capture complex PDE behavior.

\subsection{Physics pre-training}\label{physics_pretraining}

For transient tasks, GMP focuses on masked autoencoding of PDE data \nt{that is generated online at training time}.
\nt{Extending the} Tadpole~\citep{tadpole} approach for structured datasets, we make use of spectral-based solvers for randomized canonical PDE tasks for rapid, online data generation.
The models are exclusively pre-trained on static 3D single-channel data and can be effectively fine-tuned for novel multi-channel scenarios or lower dimensions, offering flexibility in channel depth and cross-dimensional transferability. A sketch of the data generation is provided in Figure~\ref{fig:sketch_physics}. The process begins by randomly selecting a PDE type, its physical parameters $p$, an initial condition $IC$, and the numerical resolution \text{Res}. Then, the spectral solver $\mathcal{P}$ computes the solution trajectory, from which a single snapshot is extracted
\begin{equation}
    u_\text{GRID} = \mathcal{P}(t, p, IC, \text{Res}).
\end{equation}
Given the varying resolutions and field counts of the raw simulations, the domain is cropped to a uniform fixed-resolution grid and a single random channel. Then, we select a random geometry $x_S$ and a deformed volume grid $x_V$ (from the geometric pre-training framework), and interpolate the 3D regular data at these coordinates $x=(x_S, x_V)$ to obtain the field values $u$. This point cloud serves as the model input. For neural operators that support queryable decoders, the sampling process is repeated to generate a different set $(x', u')$ as the target for a masked reconstruction
\begin{equation}
    u=\mathcal{I}(u_\text{GRID},x) , \hspace{1cm}  u'=\mathcal{I}(u_\text{GRID},x') , \hspace{1cm} \text{s.t. } x \cap x' = \emptyset,
\end{equation}
where $\mathcal{I}$ is the interpolation operator. Then, the model's encoder $E_\theta$ and decoder $D_\theta$ are used to predict $\hat{u}'_\theta = D_\theta(E_\theta(x, u), x'),$ and the final reconstruction task is defined as $\mathcal{L} = ||\hat{u}'_\theta - u'||_2^2$.

\subsection{Fine-tuning}\label{finetuning}

\begin{wrapfigure}[6]{r}{0.49\textwidth}
\vspace{-16pt}
  \begin{center}
    \includegraphics[width=0.49\textwidth]{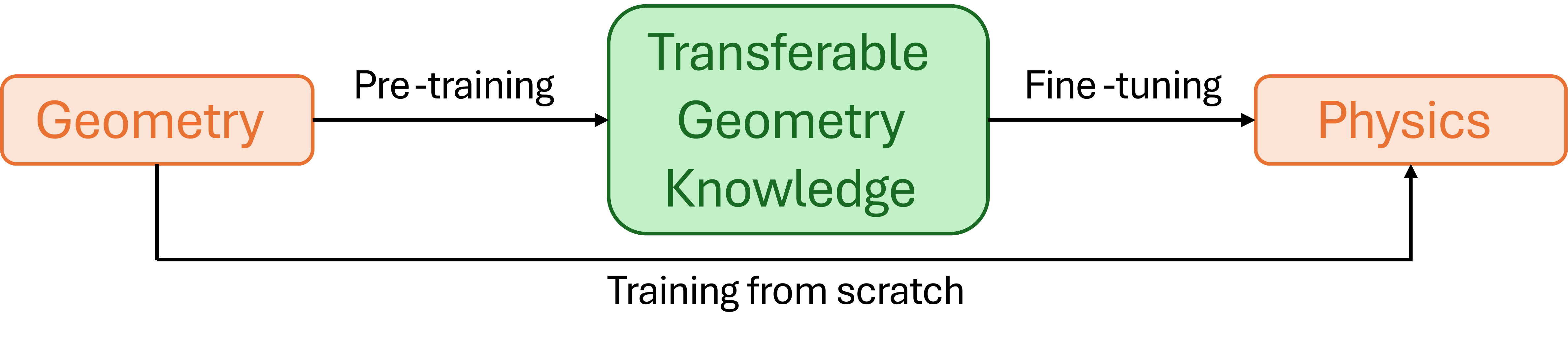}
  \end{center}
  \vspace{-10pt}
  \caption{Geometric pre-training sketch.}
  \label{fig:intuition}
\end{wrapfigure}

We evaluate pre-trained models against trained-from-scratch baselines across diverse steady-state and transient autoregressive downstream tasks. Note that the description of the pre-training method does not involve details of any particular architecture, making it model-agnostic. During fine-tuning, transient models transition from pre-training masked reconstruction to single-step dynamics. During pre-training, the model learns transferable spatial representations for static physics, and once this is learned, the fine-tuned model learns how to perform dynamics between these learned states.

Steady-state models map directly from geometry to physical target fields. Because in this setup the geometry fully determines the physical outputs, pre-learning to understand geometries helps to transfer knowledge into how fluids flow around these geometries. Both setups utilize full-parameter fine-tuning initialized from their respective pre-trained checkpoints. This process is sketched in Figure~\ref{fig:intuition}.

Because downstream tasks often present differing input and output channel dimensions, the first and last projection layers of the models are initialized from scratch. To safeguard pre-trained representations, backbone weights remain frozen during initial warmup epochs while optimizing only the projection layers. We enforce this identical fine-tuning protocol across all pre-trained baselines, including SMART, AB-UPT and Transolver++~\citep{transolver_pp}.

\begin{figure}[t]
     \centering
     \begin{subfigure}[c]{0.49\textwidth}
         \centering
         \includegraphics[width=0.88\textwidth]{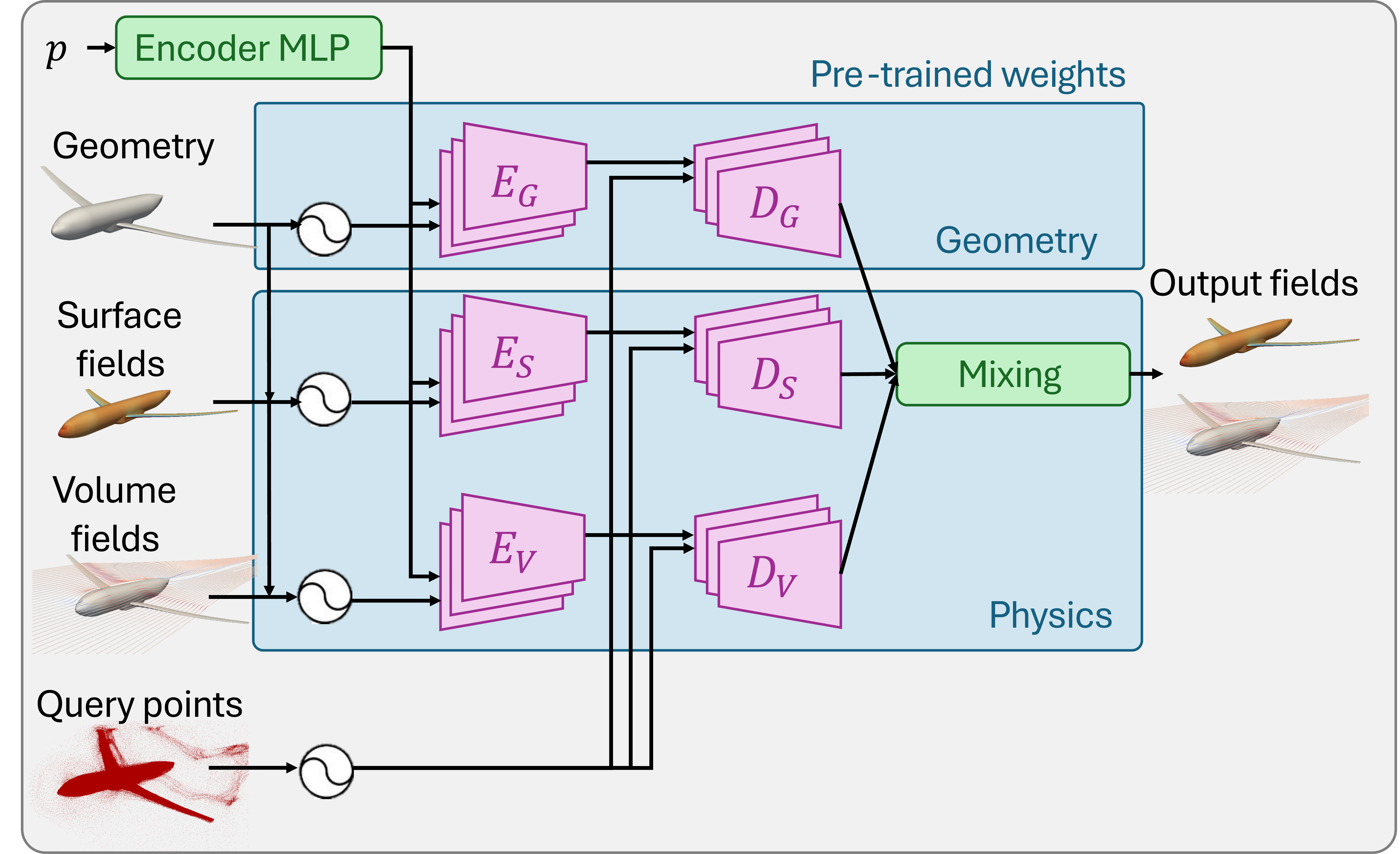}
         \caption{General overview}
         \label{fig:sketch_model_overview}
     \end{subfigure}
     \hfill
     \begin{subfigure}[c]{0.49\textwidth}
         \centering
         \includegraphics[width=1.0\textwidth]{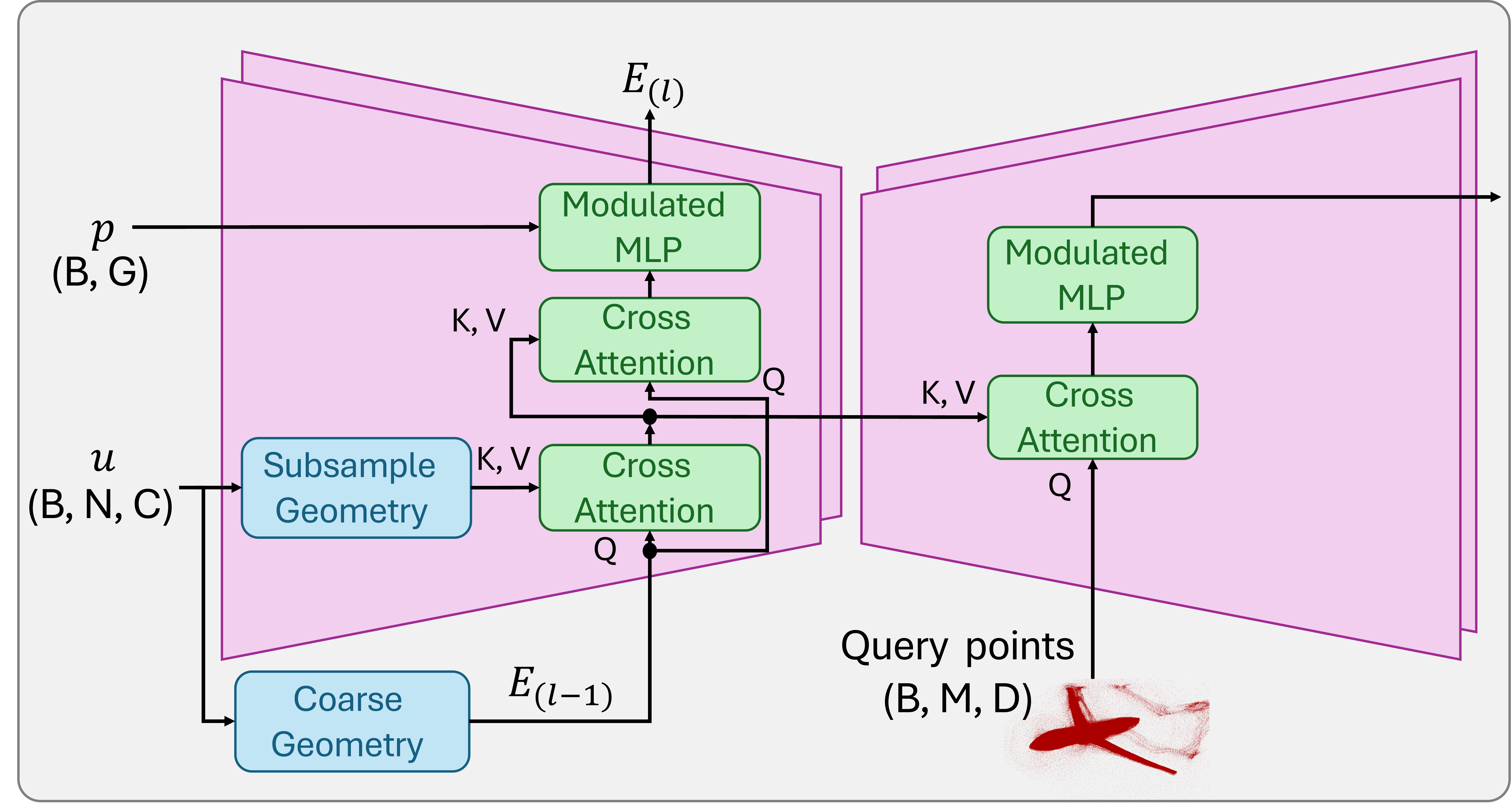}
         \caption{Details of each encoder-decoder block}
         \label{fig:sketch_block}
     \end{subfigure}
     \caption{Sketch of GMP-Net combining three encoders: one processing the general geometry of the object if available, and the other two the autoregressive physical fields. A decoder cross-attends the query positions to the output of the encoders. Finally, we compute a weighted sum using three learnable weights. Each encoder contains cross-attention from a set of anchor points to the rest of the point cloud, and an MLP modulated with external or simulation parameters.}
     \label{fig:sketch_model}
\end{figure}

\subsection{Architecture}\label{models}

\nt{While the pre-training design described so far is agnostic to the specific architecture, assessing the most competitive architectures remains crucial for achieving optimal performance.}
Hence, we benchmark several modern neural operators: AB-UPT, Erwin, PTV3, Transolver++, GeoTransolver and SMART across three test cases DrivAerML~\citep{drivaerml}, SHIFT-Wing~\citep{shift_wing}, and SHIFT-Crash~\citep{shift_crash}  (Appendix~\ref{app:model_selection}).
\nt{Our evaluation shows that SMART consistently achieves the highest standalone accuracy. As such, it presents the most challenging baseline to improve upon via pre-training and is chosen as the central architecture for the experiments below.}

\paragraph{Architectural extensions for transient dynamics} 
Our ultimate goal is to design a modular architecture that excels across both steady-state and transient regimes while maximizing the benefits of pre-training.
As the regular SMART architecture cannot perform autoregressive rollouts, we generalize it into a modular triple-encoder framework (Figure~\ref{fig:sketch_model}). A geometric encoder $E_G$ embeds boundary shapes, while dedicated physical encoders $E_S$ and $E_V$ process surface and volumetric physical states $u_t$, respectively. Each branch applies sinusoidal positional encodings and compresses point clouds via anchor-based cross-attention into latent tokens $z_G, z_S, z_V$. Decoders then cross-attend target query coordinates $x_Q$ to these latents to predict $u_{t+1}$, balanced by learned task weights $w_i$:
\begin{equation}\label{eq:smart2}
    u_{t+1} = \sum_{i \in \{G, S, V\}} w_i \cdot D_i(z_i, x_Q), \qquad z_G = E_G(x_G, p);\quad z_{S,V} = E_{S,V}(x_{S,V}, u_{S,V}, p).
\end{equation}

This decoupled formulation allows modular execution: branches are activated selectively based on problem requirements. For steady-state regimes, only the geometric branch ($E_G, D_G$) is active, naturally recovering the baseline SMART model. For transient problems, the surface ($E_S, D_S$) and volumetric ($E_V, D_V$) branches are toggled according to data availability. Crucially, this modularity facilitates decoupled pre-training: the geometric branch is pre-trained using geometric objectives (Section~\ref{geometric_pretraining}), while the physical branches are pre-trained online without stored simulation data (Section~\ref{physics_pretraining}). We denote this new extended model as \textbf{GMP-Net}.

\subsection{Downstream application datasets}

\nt{To show pre-training impact for real-world scenarios,}  we evaluate on the following downstream engineering benchmarks unseen during pre-training.

\begin{figure}[t]
    \centering
    \includegraphics[width=\textwidth]{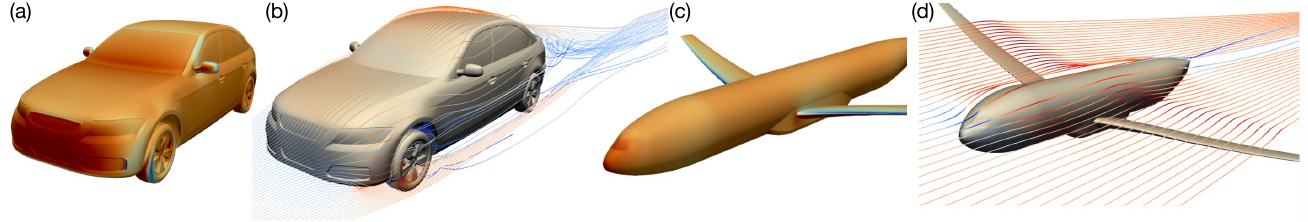}
    \caption{Samples showing surface (pressure in a,c) and volume (velocity streamlines in b,d) fields: DrivAerML (car, left) and SHIFT-Wing (aircraft, right).}
    \label{fig:four_samples}
\end{figure}

\paragraph{Steady-State Datasets} 
To rigorously assess steady-state operator performance, we evaluate on two primary benchmarks: DrivAerML~\citep{drivaerml}, an open-source automotive benchmark comprising 500 geometric variations of the DrivAer notchback model, and SHIFT-Wing~\citep{shift_wing}, an aerospace testbed featuring thousands of parametrically perturbed NASA Common Research Model (CRM) configurations. While DrivAerML isolates geometric variation under fixed inlet velocities, SHIFT-Wing systematically varies free-stream operating conditions. 
Representative geometries and flow fields are illustrated in Figure~\ref{fig:four_samples}.
For both datasets, our model predicts surface pressure $p_s$ and wall shear stress $\tau$, as well as volumetric pressure $p_v$ and velocity fields $\textbf{u}$. Finally, to evaluate transferability to fundamentally distinct physical regimes, we evaluate on SHIFT-Crash~\citep{shift_crash}, an automotive structural mechanics benchmark.

\paragraph{Transient Datasets} 
Given that transient benchmarks on unstructured meshes are relatively scarce, we employ a combination of synthetic experiments and established open datasets. First, we construct a synthetic testbed by interpolating planar Kuramoto--Sivashinsky (KS) solutions onto deformed, irregular grids. While pre-training leverages static 3D fields, this setup verifies whether the architecture generalizes across spatial dimensions and captures unfamiliar temporal dynamics. For instance, to solve 2D downstream tasks, coordinates are zero-padded to $(x,y,0)$. Second, we adopt the 2D flow past an ellipse from DGN~\citep{mario_diffusion_gnn} to predict time-dependent pressure and velocity fields. Lastly, we evaluate the dynamic response of SHIFT-Crash over the given 12 discrete time steps. For autoregressive rollout experiments, models unroll predictions across the entire temporal sequence; under the steady-state formulation, the model directly maps initial geometries and crash parameters to the final deformed state. Comprehensive benchmark details are provided in Table~\ref{tab:datasets}, where \textit{Type} indicates volumetric (V) or surface-based (S) meshes, and the transient sample count denotes total trajectories multiplied by the temporal resolution per rollout.

\begin{table}[t]
    \centering
    \small
    \caption{Overview of the fine-tuning datasets used for downstream tasks evaluation.}
    \label{tab:datasets}
    \begin{tabular}{@{}lllllll@{}}
        \toprule
        Dataset & Time & Dim. & Type & \# Samples & \nt{\# Points (S;V)} & \# Fields \\ \midrule
        DrivAerML \citep{drivaerml} & Steady & 3D & S; V & 500 & $\sim$ 9M ; 160M & 4+4 \\
        SHIFT-Wing \citep{shift_wing} & Steady & 3D & S; V & 6000 & $\sim$ 3M ; 6M & 4+4 \\
        KS (ours) & Transient & 2D & V & 630 $\times$ 30 & - ; 4096 & 1 \\
        Ellipse \citep{mario_diffusion_gnn} & Transient & 2D & V & 1200 $\times$ 100 & - ; 1024 & 3\\
        SHIFT-Crash \citep{shift_crash} & Both & 3D & S & 768 $\times$ 12 & $\sim$ 400K ; - & 4 \\ \bottomrule
    \end{tabular}
\end{table}

\begin{table}[t]
    \centering
    \caption{Comparison between training from scratch and pre-training (PT) on steady-state tasks: relative $L_2$ error and pre-training improvement. Tests for different numbers of training samples. 
    A Transolver++ variant pre-trained with GeoPT~\citep{geopt}  is also shown for comparison.
    }
    \label{tab:steady_final_results}
    \scriptsize
    \setlength{\tabcolsep}{2.5pt}
    \resizebox{\columnwidth}{!}{%
    \begin{tabular}{l *{12}{c}} 
    \toprule
    & \multicolumn{4}{c}{DrivAerML} & \multicolumn{4}{c}{SHIFT-Wing} & \multicolumn{4}{c}{SHIFT-Crash} \\
    \cmidrule(lr){2-5} \cmidrule(lr){6-9} \cmidrule(lr){10-13}
    Training samples & 16 & 32 & 64 & 128 & 16 & 32 & 64 & 128 & 16 & 32 & 64 & 128 \\ 
    \midrule

    Transolver++ (scratch) $\downarrow$    & 0.685 & 0.559 & 0.329 & 0.180 & 0.603 & 0.403 & 0.235 & 0.209 & 0.611 & 0.601 & 0.548 & 0.328 \\
    Transolver++ (GeoPT) $\downarrow$    & 0.520 & 0.443 & 0.309 & 0.218 & 0.308 & 0.266 & 0.233 & 0.204 & 0.488 & 0.428 & 0.410 & 0.388 \\
    Transolver++ (GMP ours) $\downarrow$ & 0.503 & 0.453 & 0.302 & 0.178 & 0.318 & 0.265 & 0.232 & 0.205 & 0.530 & 0.563 & 0.356 & 0.322 \\
    Rel. \hspace{0.07cm}Improvement $\uparrow$ & \cellcolor{Green!27}27\% & \cellcolor{Green!19}19\% & \cellcolor{Green!8}8\% & \cellcolor{Green!1}1\% & \cellcolor{Green!47}47\% & \cellcolor{Green!34}34\% & \cellcolor{Green!1}1\% & \cellcolor{Green!2}2\% & \cellcolor{Green!13}13\% & \cellcolor{Green!6}6\% & \cellcolor{Green!34}34\% & \cellcolor{Green!2}2\% \\

    \midrule

    AB-UPT (scratch) $\downarrow$    & 0.439 & 0.313 & 0.244 & 0.184 & 0.569 & 0.456 & 0.379 & 0.298 & 0.396 & 0.363 & 0.340 & 0.326 \\
    AB-UPT (GMP ours) $\downarrow$ & 0.450 & 0.316 & 0.218 & 0.155 & 0.596 & 0.432 & 0.347 & 0.273 & 0.389 & 0.362 & 0.309 & 0.310 \\
    Rel. \hspace{0.07cm}Improvement $\uparrow$ & \cellcolor{Green!0}-2\% & \cellcolor{Green!0}-1\% & \cellcolor{Green!20}10\% & \cellcolor{Green!32}16\% & \cellcolor{Green!0}-5\% & \cellcolor{Green!8}4\% & \cellcolor{Green!16}8\% & \cellcolor{Green!14}7\% & \cellcolor{Green!4}2\% & \cellcolor{Green!0}0\% & \cellcolor{Green!18}9\% & \cellcolor{Green!10}5\% \\

    \midrule
    
    GMP-Net (scratch) $\downarrow$    & 0.412 & 0.348 & 0.187 & 0.099 & 0.274 & 0.236 & 0.234 & 0.217 & 0.432 & 0.398 & 0.359 & 0.326 \\
    GMP-Net (PT ours) $\downarrow$ & 0.346 & 0.299 & 0.178 & 0.097 & 0.253 & 0.224 & 0.214 & 0.196 & 0.395 & 0.358 & 0.327 & 0.300 \\
    Rel. \hspace{0.07cm}Improvement $\uparrow$ & \cellcolor{Green!32}16\% & \cellcolor{Green!28}14\% & \cellcolor{Green!10}5\% & \cellcolor{Green!6}3\% & \cellcolor{Green!14}7\% & \cellcolor{Green!5}5\% & \cellcolor{Green!18}9\% & \cellcolor{Green!20}10\% & \cellcolor{Green!18}9\% & \cellcolor{Green!20}10\% & \cellcolor{Green!18}9\% & \cellcolor{Green!16}8\% \\

    \bottomrule
    \end{tabular}%
    }
\end{table}

\begin{figure}[t]
    \centering
    \includegraphics[width=0.8\textwidth]{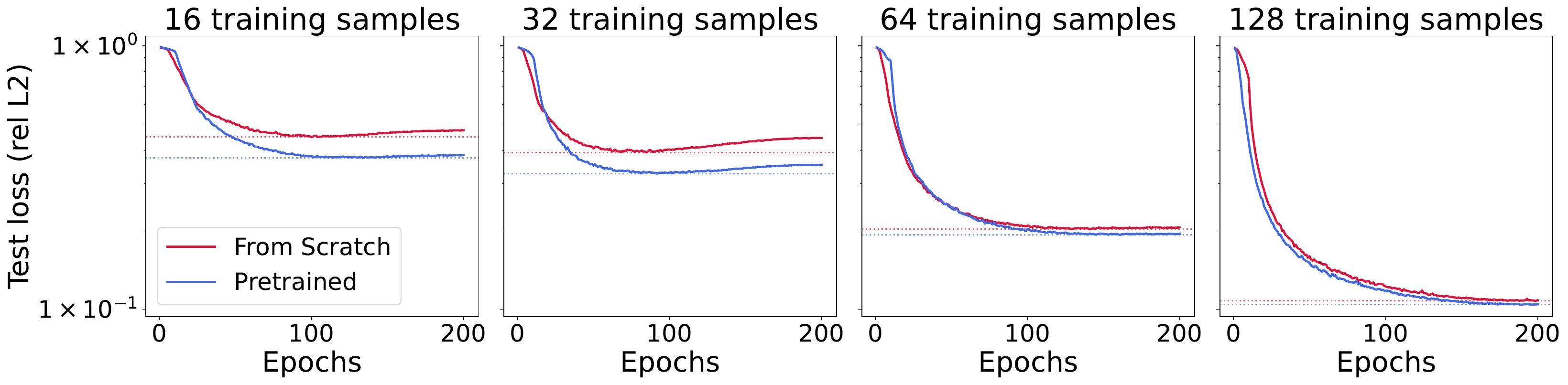}
    \caption{DrivAerML test loss of GMP-Net trained from scratch and with pre-training. Horizontal lines indicate the minimum error. Pre-trained models achieve lower error and faster convergence.}
    \label{fig:combined_loss_plots}
\end{figure}

\section{Experiments}\label{experiments}

\subsection{Results for steady-state problems}

\begin{figure}[t]
    \centering
    \begin{subfigure}[c]{0.32\textwidth}
        \includegraphics[width=\textwidth]{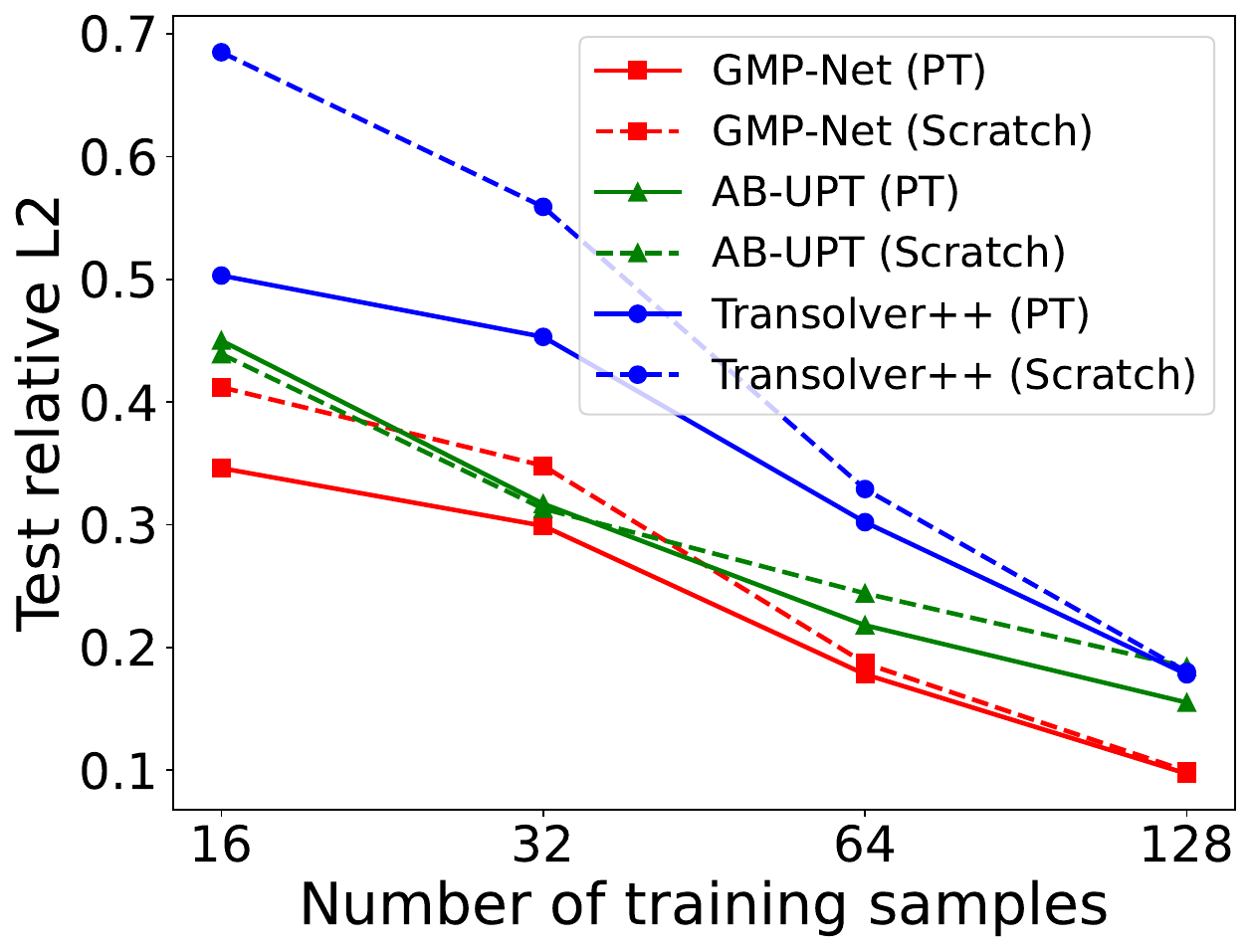}
        \caption{DrivAerML.}
    \end{subfigure}
    \hfill
    \begin{subfigure}[c]{0.32\textwidth}
        \centering
        \includegraphics[width=\textwidth]{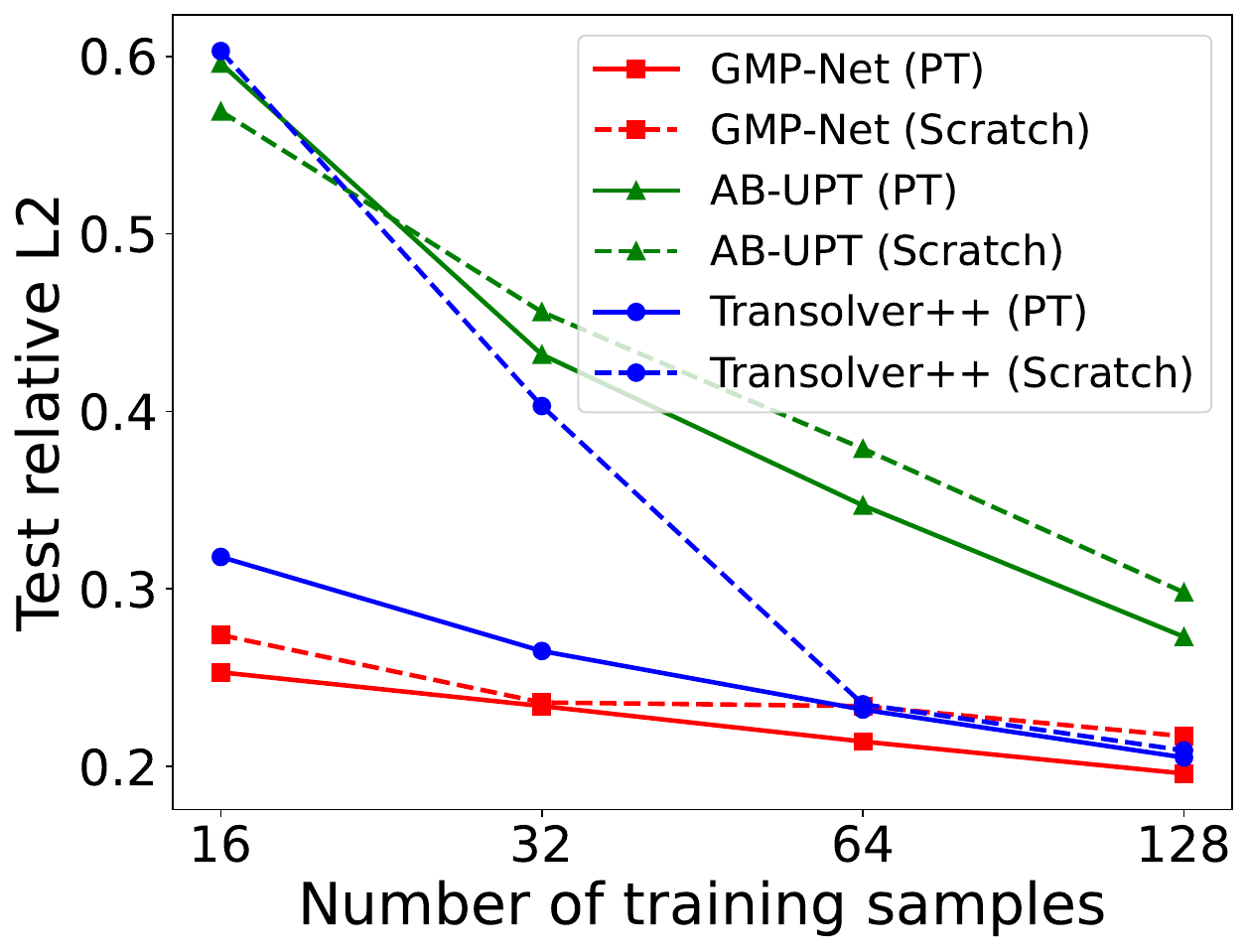}
        \caption{SHIFT-Wing.}
    \end{subfigure}
    \hfill
    \begin{subfigure}[c]{0.32\textwidth}
        \centering
        \includegraphics[width=\textwidth]{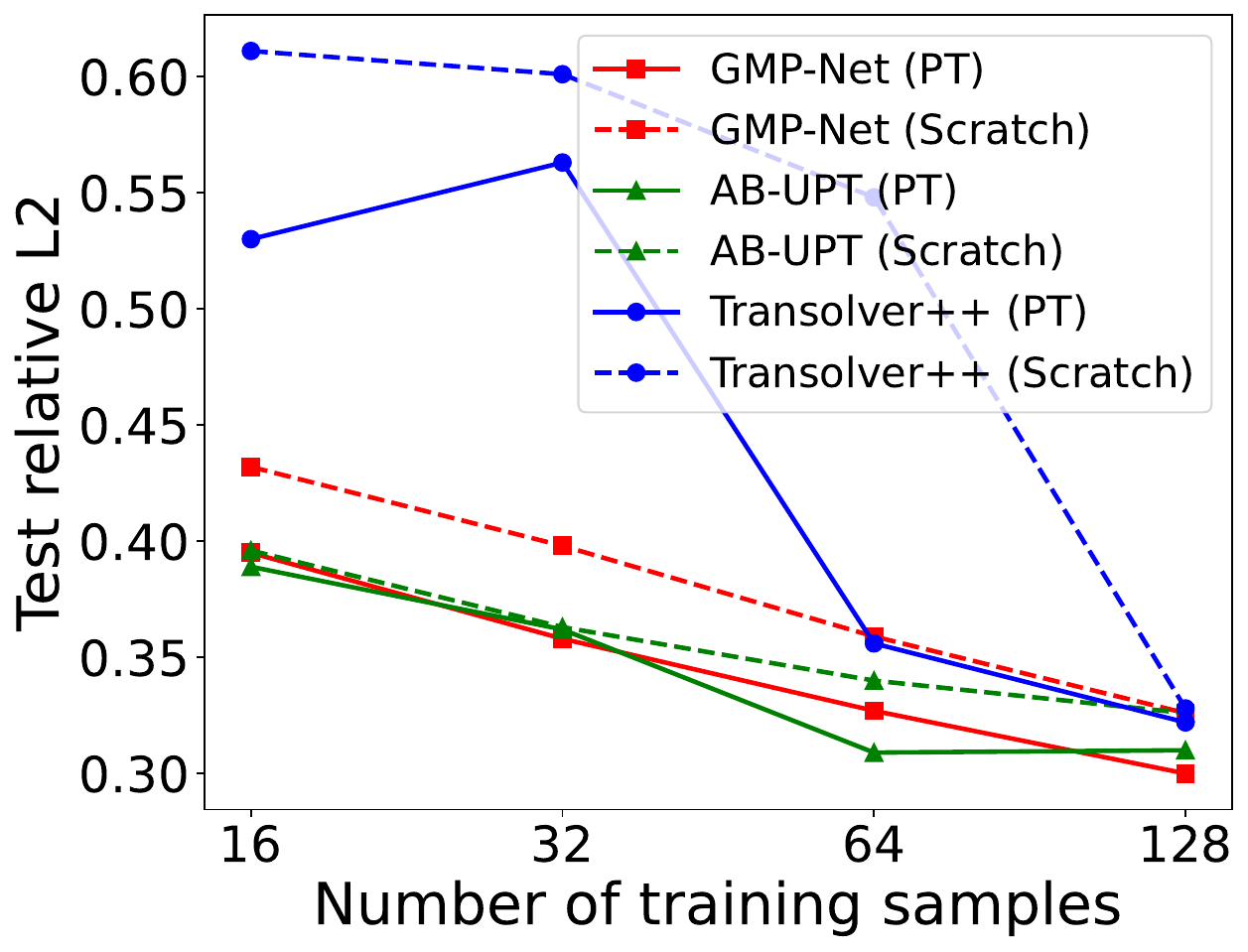}
        \caption{SHIFT-Crash.}
    \end{subfigure}
    \hfill
    \caption{Summary of model performance in steady problems. We show the final relative $L_2$ error for different fine-tuning dataset sizes.}
    \label{fig:data_scaling}
\end{figure}

\begin{figure}[t]
    \centering
    \begin{subfigure}[c]{0.32\textwidth}
        \includegraphics[width=\textwidth]{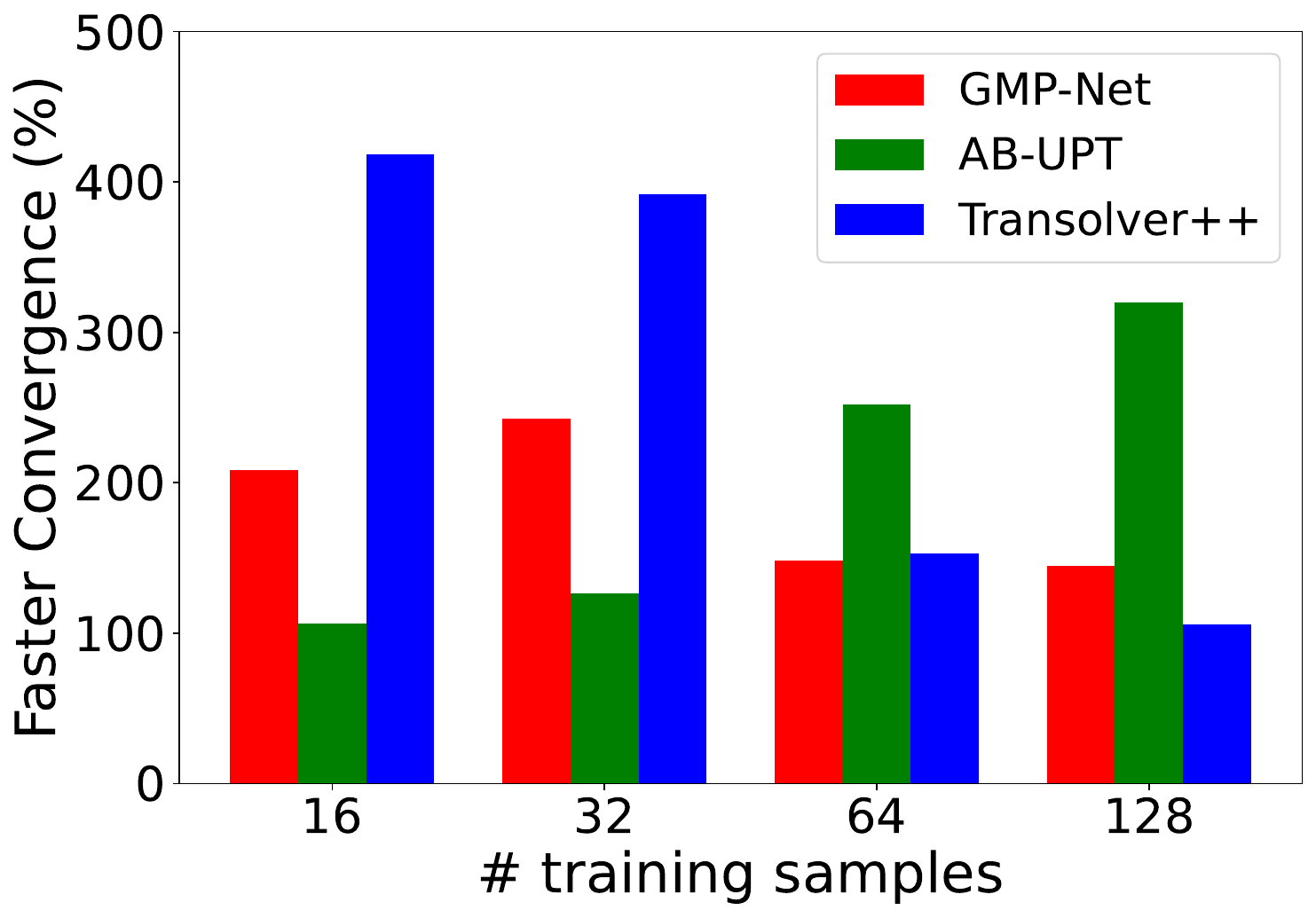}
        \caption{Convergence (DrivAerML).}
        \label{fig:faster_convergence}
    \end{subfigure}
    \hfill
    \begin{subfigure}[c]{0.32\textwidth}
        \centering
        \includegraphics[width=\textwidth]{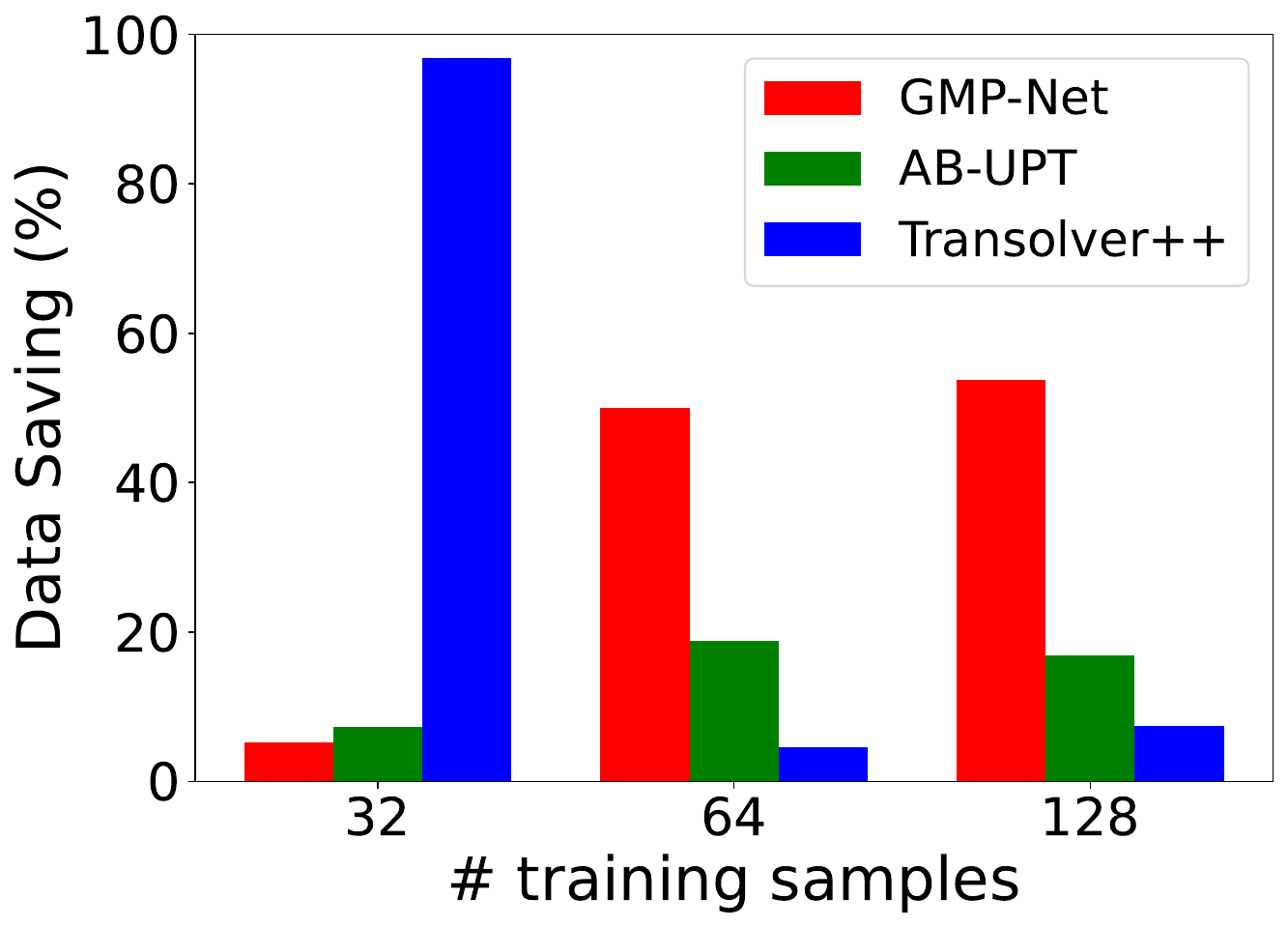}
        \caption{Data efficiency (SHIFT-Wing).}
        \label{fig:data_efficiency}
    \end{subfigure}
    \hfill
    \begin{subfigure}[c]{0.32\textwidth}
        \centering
        \includegraphics[width=\textwidth]{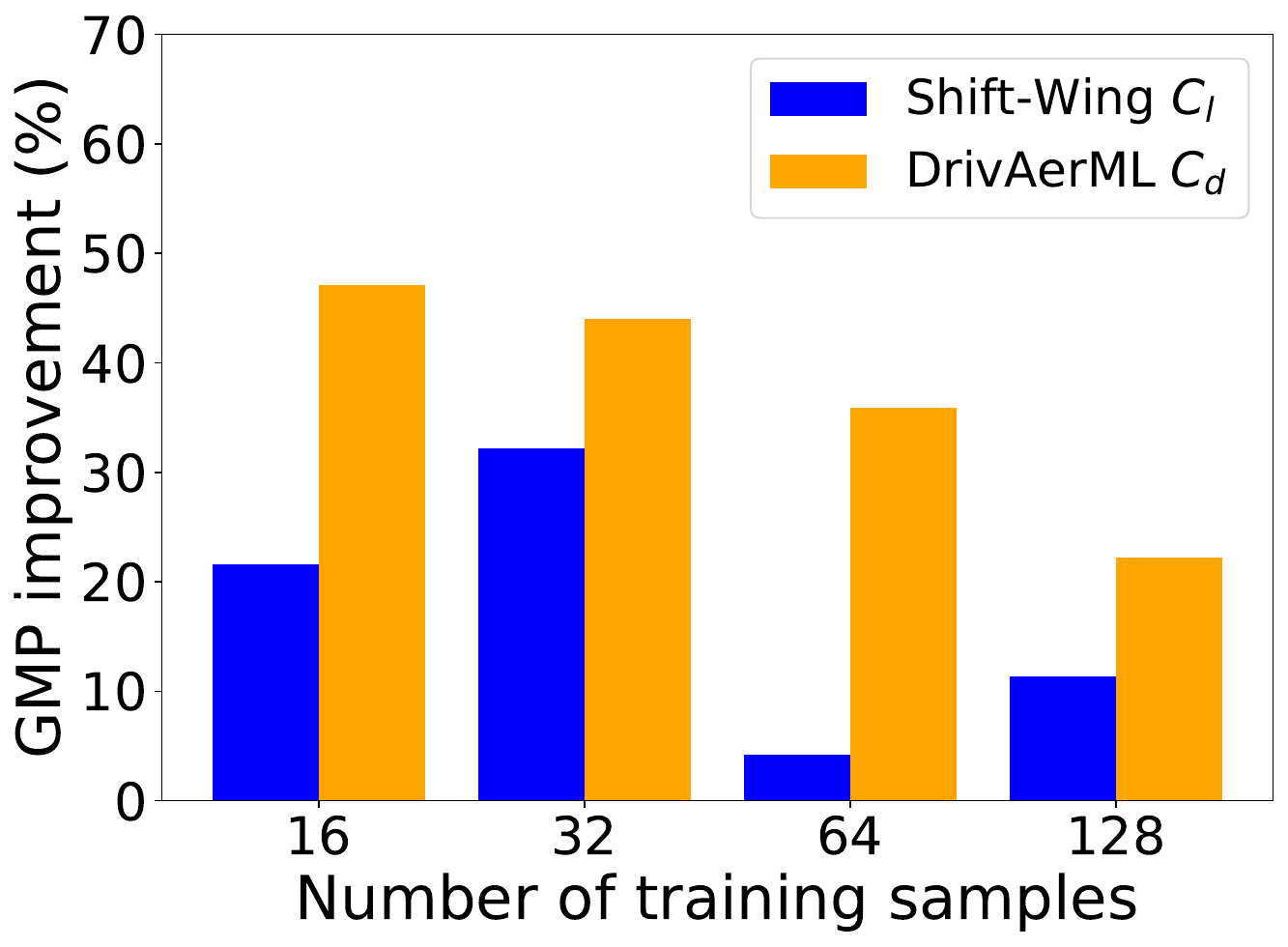}
        \caption{Aerodynamic forces.}
        \label{fig:aero_forces}
    \end{subfigure}
    \hfill
    \caption{Summary of model performance in steady problems. a) Faster convergence analysis shows how much faster pre-trained models converge to the accuracy observed by scratch versions. b) Data efficiency showing how much data can be saved. c) $R^2$ improvement on the computation of the aerodynamic coefficients thanks to pre-training.}
    \label{fig:miscelanea}
\end{figure}

\paragraph{Accuracy} To demonstrate the practical benefits of our pre-training, we adopt a low-data regime analysis. For all datasets, we utilize a subset of 144 samples, reserving a fixed random set of 16 samples for testing, and the remaining 128 samples are used to create nested training sets of varying sizes, allowing us to analyze how accuracy scales with the volume of training data. The results, summarized in Table~\ref{tab:steady_final_results}, consistently demonstrate that our pre-training method enhances accuracy across all models in most data regimes. However, the behaviour is different across them.
For GMP-Net and Transolver++ we observe a clear and expected trend: the accuracy gain is most pronounced in the sparsest data regimes, with a maximum of 47\% improvement for 16 training samples. In addition, GMP-Net shows a more stable improvement across the experiments. For AB-UPT, we observe no measurable gains in highly data-sparse regimes due to severe overfitting: both the scratch and pre-trained variants rapidly converge to nearly identical error levels. With limited training data, model performance appears bounded by an accuracy plateau that both approaches reach almost immediately. Overall, our GMP-Net, derived from SMART, achieves the lowest errors.

\paragraph{Convergence and data saving} 
Figure~\ref{fig:combined_loss_plots} illustrates the test loss trajectories on DrivAerML for our GMP-Net model. Comparing the number of optimization steps required for the pre-trained model to match the performance of the model trained from scratch highlights a substantial convergence speedup (Figure~\ref{fig:faster_convergence}). Furthermore, data-scaling experiments (Figure~\ref{fig:data_scaling}) show that fine-tuning on only 16 samples outperforms training from scratch with 32 samples, delivering significant sample efficiency gains as summarized in Figure~\ref{fig:data_efficiency}. Finally, evaluating lift and drag coefficients ($C_L, C_D$) confirms that GMP pre-training improves the agreement of predicted and reference lift coefficients, with the largest gains in the low-data regime (Figure~\ref{fig:aero_forces}).

\paragraph{Computational efficiency}
GMP geometric pre-training requires only 4 hours on a single NVIDIA A100, negligible compared to the 80 hours required to train a baseline like SMART on SHIFT-Wing from scratch for 200 epochs. Thus, pre-training can be integrated as a light pre-processing step.

\paragraph{Comparison against other pre-training methods} 
We compare our pre-training framework against GeoPT on Transolver++. Although official checkpoints are public, we pre-train and fine-tune GeoPT from scratch under a controlled reimplementation, maintaining identical model configurations across all experiments while evaluating different downstream datasets (original published numbers available in Appendix~\ref{app:geopt_comparison} for completeness). Under identical hardware, GeoPT pre-training requires $\sim$12 hours per epoch (taking $\sim$10 days to reach a test MSE of $0.02$), whereas our disk-data-free pre-training converges in just 4 hours.
As reported in Table~\ref{tab:steady_final_results} (under the Transolver++ setting), our model matches GeoPT's performance in low-data regimes, except for SHIFT-Crash, and outperforms it as the downstream training set scales. In contrast to our framework, \citet{cheap_geometry} relies on high-fidelity CAD geometries for pre-training, less practical in workflows where such meshes are proprietary or unavailable. Nonetheless, their reported $10\text{--}20\%$ accuracy gains remain comparable to the improvements achieved by our method.

\begin{wraptable}[11]{r}{6.8cm}
\vspace{-13pt}
    \centering
    \caption{Effect of geometric descriptors on GMP pre-training accuracy gains. Mean rel $L_2$ error.}
    \label{tab:ablation}
    \small
    \setlength{\tabcolsep}{4pt}
    \begin{tabular}{lcccc} 
    \toprule
    & \multicolumn{4}{c}{DrivAerML average $\downarrow$} \\
    \cmidrule(lr){2-5}
    Training samples & 16 & 32 & 64 & 128 \\ 
    \midrule
    GMP-Net (scratch) & 0.412 & 0.348 & 0.187 & \textbf{0.099} \\
    GMP (occupancy)       & 0.424 & 0.379 & 0.243 & 0.138 \\
    GMP (VDF)       & 0.402 & 0.333 & 0.189 & 0.103 \\
    GMP (N and C)   & \textbf{0.387} & \textbf{0.311} & \textbf{0.182} & 0.101 \\
    GMP (all)      & \best{0.346} & \best{0.299} & \best{0.178} & \best{0.097} \\
    \bottomrule
    \end{tabular}
\end{wraptable}

\paragraph{Geometric descriptors ablation} We examine the contribution of individual geometric descriptors during pre-training. While prior works rely primarily on VDF~\citep{geopt} or binary occupancy~\citep{cheap_geometry}, GMP integrates surface normals, curvature, and volumetric VDF. Table~\ref{tab:ablation} evaluates different pre-training variants: surface features only (N and C), volumetric VDF alone, and their combination (all). In addition, we test pre-training using only binary occupancy (second row), the pre-training objective from \citet{cheap_geometry}, but this fails to show any improvement in GMP and is therefore omitted from our final setup. Notably, VDF alone provides the weakest downstream transfer, whereas surface normals and curvature substantially improve representations. This aligns with aerodynamic intuition, as resolving steep boundary-layer gradients necessitates accurate differential surface geometry.

\begin{table}[h!]
    \centering
    \caption{NRMSE for transient problems. Comparison of from-scratch models and GMP pre-training.}
    \small
    \label{tab:transient_results_models}
    \setlength{\tabcolsep}{4pt}
    \begin{tabular}{l ccc p{0.05cm} ccc p{0.05cm} ccc}
        \toprule
        & \multicolumn{3}{c}{2D KS $\downarrow$} && \multicolumn{3}{c}{2D Ellipse $\downarrow$} && \multicolumn{3}{c}{3D SHIFT-Crash $\downarrow$} \\
        \cmidrule{2-4} \cmidrule{6-8} \cmidrule{10-12}
        Timesteps & $1$ & $5$ & $10$ && $1$ & $5$ & $10$ && $1$ & $5$ & $10$ \\ 
        \midrule
        Erwin         & 0.143 & 0.713 & 1.281 && 0.016 & 0.083 & 0.438 && 0.039 & 0.131 & 0.245 \\
        Transolver++  & 0.203 & 0.779 & 1.028 && \best{0.006} & \textbf{0.021} & \textbf{0.042} && 0.020 & 0.072 & 0.072 \\
        \midrule
        GMP-Net (scratch)        & \textbf{0.030} & \textbf{0.196} & \textbf{0.615} && 0.008 & 0.025 & 0.044 && \textbf{0.009} & \textbf{0.030} & \textbf{0.040} \\
        GMP-Net (PT)           & \best{0.024} & \best{0.148} & \best{0.479} && \best{0.006} & \best{0.020} & \best{0.033} && \best{0.008} & \best{0.026} & \best{0.034} \\
        \midrule
        Rel. Improvement  & \cellcolor{Green!20} 20\% & \cellcolor{Green!24} 24\% & \cellcolor{Green!22} 22\% && \cellcolor{Green!25} 25\% & \cellcolor{Green!20} 20\% & \cellcolor{Green!23} 23\% && \cellcolor{Green!11} 11\% & \cellcolor{Green!13} 13\% & \cellcolor{Green!15} 15\% \\
        \bottomrule
    \end{tabular}
\end{table}

\begin{figure}[t]
    \centering
    \includegraphics[width=\textwidth]{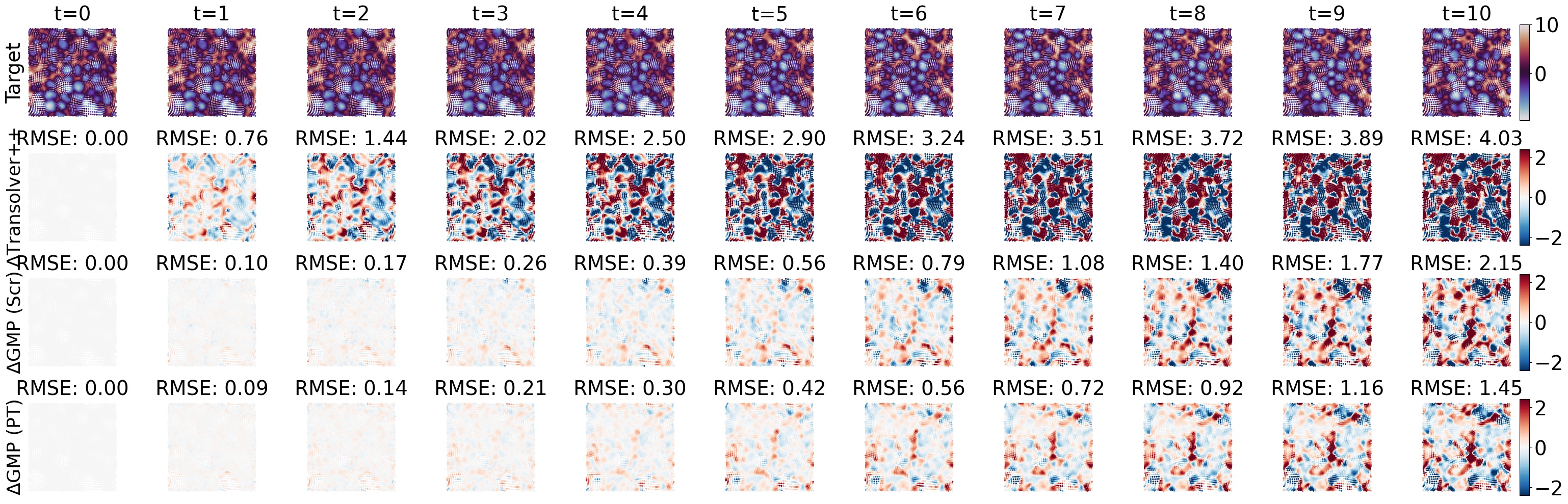} 
    \caption{2D KS equation rollout error comparison. GMP-Net shows state-of-the-art accuracy compared to other baselinse. Physics pre-trained GMP-Net exhibits improved performance over the model trained from scratch.}
    \label{fig:rollout_ks_error}
\end{figure}

\subsection{Results for time-dependent problems}

Table~\ref{tab:transient_results_models} compares rollout NRMSE against baselines, first showing that GMP-Net (scratch) already achieves the lowest error among the evaluated baselines for 2D KS and 3D SHIFT-Crash, and comparable to Transolver++ in the 2D Ellipse. Most importantly, across all scenarios, pre-training reduces error by $10\text{--}25\%$, demonstrating that inductive biases learned on simple PDEs transfer effectively to more complex systems. This transfer also bridges spatial dimensions and channel configurations, successfully mapping from 3D single-channel pre-training to 2D multi-channel tasks. Qualitative rollouts are illustrated in Figure~\ref{fig:rollout_ks_error} and Appendix~\ref{app:extra_plots}.

\begin{wraptable}[8]{r}{6.0cm}
\vspace{-13pt}
    \centering
    \caption{1-step NMSE test loss and GMP pre-training reduction.}
    \label{tab:transient_results_training_size}
    \small
    \setlength{\tabcolsep}{2pt}
    \begin{tabular}{lccc}
        \toprule
        \# Trajectories & 16 & 32 & 64 \\ 
        \midrule
        GMP-Net (scratch) $\times 10^{-3}$ $\downarrow$ & 8.8 & 6.9 & 5.4 \\
        GMP-Net (PT) $\times 10^{-3}$ $\downarrow$    & 6.2 & 4.8 & 3.9 \\
        \midrule
        Improvement $\uparrow$           & 29\% & 30\% & 28\% \\
        \bottomrule
    \end{tabular}
\end{wraptable}

Finally, we analyze the accuracy gains as a function of the amount of training data. While the results in Table \ref{tab:transient_results_models} use the complete datasets, results in Table \ref{tab:transient_results_training_size} use only 16, 32, and 64 training trajectories. We compare the final 1-step test loss (NMSE) averaged across the three datasets, and we observe consistent accuracy gains, with larger absolute gains in the smaller-data regimes, but around 30\% relatively in all three cases.

\section{Conclusions, limitations and future work}\label{conclusion}

We have introduced \textit{GMP}, an innovative pre-training framework for unstructured simulations based on online generation of physics data and advanced geometric descriptors. GMP enhances the accuracy of diverse downstream tasks, achieving greater data efficiency and faster convergence. Since the pre-training setup is computationally lightweight, it can be integrated as a standard pre-processing step to improve performance without a substantial increase in total training time. A remaining challenge is mitigating overfitting in highly complex, low-data scenarios.
Regarding geometric pre-training, interesting questions remain concerning the selection of geometric descriptors, and in the future we plan to explore whether additional features enhance model robustness and generalization. 
Importantly, unlike training from scratch, GMP amortizes initial computational costs by accelerating downstream convergence and reducing data requirements, ultimately making the design cycle more efficient. This paves the way for faster and more accurate surrogate models in science and engineering.

\begin{ack}

The authors gratefully acknowledge the scientific support and HPC resources provided by the Erlangen National High Performance Computing Center (NHR@FAU) of the Friedrich-Alexander-Universität Erlangen-Nürnberg (FAU) under the NHR project b278bb. NHR funding is provided by federal and Bavarian state authorities.

We want to thank Michael Emory and the team at Luminary Cloud for the access to the SHIFT datasets. This work would not have been possible without their support and high-fidelity data.
Special thanks goes to our colleague Rene Winchenbach for the many helpful discussions we had throughout the development of this project.

\end{ack}

\small 
\bibliographystyle{unsrtnat} 
\bibliography{references}

\newpage
\appendix

\section*{Technical appendices and supplementary material}

These appendices provide additional details on the following topics:

\begin{itemize}
    \item \ref{app:ablation_studies}: Ablation studies.
    \item \ref{app:geopt_comparison}: Details on the comparison against GeoPT.
    \item \ref{app:hyperparameters}: Training hyperparameters.
    \item \ref{app:models}: Model and baselines hyperparameters.
    \item \ref{app:model_selection}: Baseline architectures comparison.
    \item \ref{app:aero_forces}: Computation of the aerodynamic coefficients.
    \item \ref{app:gaussianDefo}: Intersection-free meshes via Gaussian random fields.
    \item \ref{app:additional_results}: Additional results: regional breakdown.
    \item \ref{app:extra_plots}: Additional plots and figures.
\end{itemize}

\section{Ablation studies for pre-training}\label{app:ablation_studies}

We begin this appendix section by presenting ablation studies on key design choices across the two pre-training regimes.

\subsection{Number of points during pre-training}

Owing to the discretization-invariant nature of modern unstructured neural operators, pre-training does not strictly require the dense spatial resolutions utilized during downstream fine-tuning, where target meshes routinely scale to millions of vertices. This decoupling enables an advantageous trade-off between computational efficiency (at lower point densities) and representational fidelity (at higher geometric resolutions). Table~\ref{tab:ablation_pt_resolution} compares three candidate point cloud resolutions for pre-training, with our default configuration occupying the intermediate operating point. As demonstrated by the empirical trends, scaling the point budget further yields diminishing accuracy gains on downstream tasks while introducing prohibitive computational overhead during the pre-training phase.

\begin{table}[h]
  \caption{Performance and fine-tuning relative $L_2$ error on DrivAerML across different volume and surface point resolutions during pre-training (PT).}
  \label{tab:ablation_pt_resolution}
  \centering
  \begin{tabular}{cccc}
    \toprule
    Vol. Points & Surf. Points & PT Time (h) & Rel $L_2$ Error FT (64 samples) \\
    \midrule
    1024 & 512  & 2.5 & 0.194 \\
    \textbf{2048} & \textbf{1024} & \textbf{4.0} & \textbf{0.178} \\
    4096 & 2048 & 9.0 & 0.175 \\
    \bottomrule
  \end{tabular}
\end{table}

\subsection{Loss weighting during pre-training}

During geometric pre-training, the model jointly predicts three distinct descriptors: local curvature $\kappa$, surface normals $\mathbf{n}$, and the vector distance function (VDF). A central question is how to properly balance these heterogeneous objective terms. We evaluate two distinct multi-task weighting strategies:

\begin{itemize}
    
    \item \textbf{Dynamic Relative Weighting:} Rebalances the objective dynamically according to the running magnitude of each loss term, preventing dominant tasks from starving weaker gradients:
    \begin{equation}
        \mathcal{L} = w_\kappa^{(t)} \mathcal{L}_\kappa + w_n^{(t)} \mathcal{L}_n + w_V^{(t)} \mathcal{L}_V,
    \end{equation}
    where normalized weights are computed at iteration $t$ as
    \begin{equation}
        w_k^{(t)} = K \cdot \frac{\tilde{w}_k^{(t)}}{\sum_{j \in \mathcal{T}} \tilde{w}_j^{(t)}}, 
        \qquad 
        \tilde{w}_k^{(t)} = \frac{1}{\bar{\mathcal{L}}_k^{(t)} + \epsilon},
    \end{equation}
    with $\mathcal{T} = \{\kappa, n, V\}$, task count $K = |\mathcal{T}| = 3$, numerical stabilizer $\epsilon > 0$, and $\bar{\mathcal{L}}_k^{(t)}$ denoting the exponential moving average (EMA) of the $k$-th loss component.

    \item \textbf{Gradient Conflict resolution (ConFIG):} Simple scalarization approaches may suffer from destructive gradient interference across competing objectives. Rather than collapsing the multi-task objectives into a single scalar loss, we leverage the ConFIG algorithm~\citep{config}, which performs gradient surgery at each optimization step to project task gradients onto a unified, conflict-free descent direction.
\end{itemize}

\begin{figure}[htbp]
    \centering
    \begin{subfigure}[b]{0.32\textwidth}
        \centering
        \includegraphics[width=\textwidth]{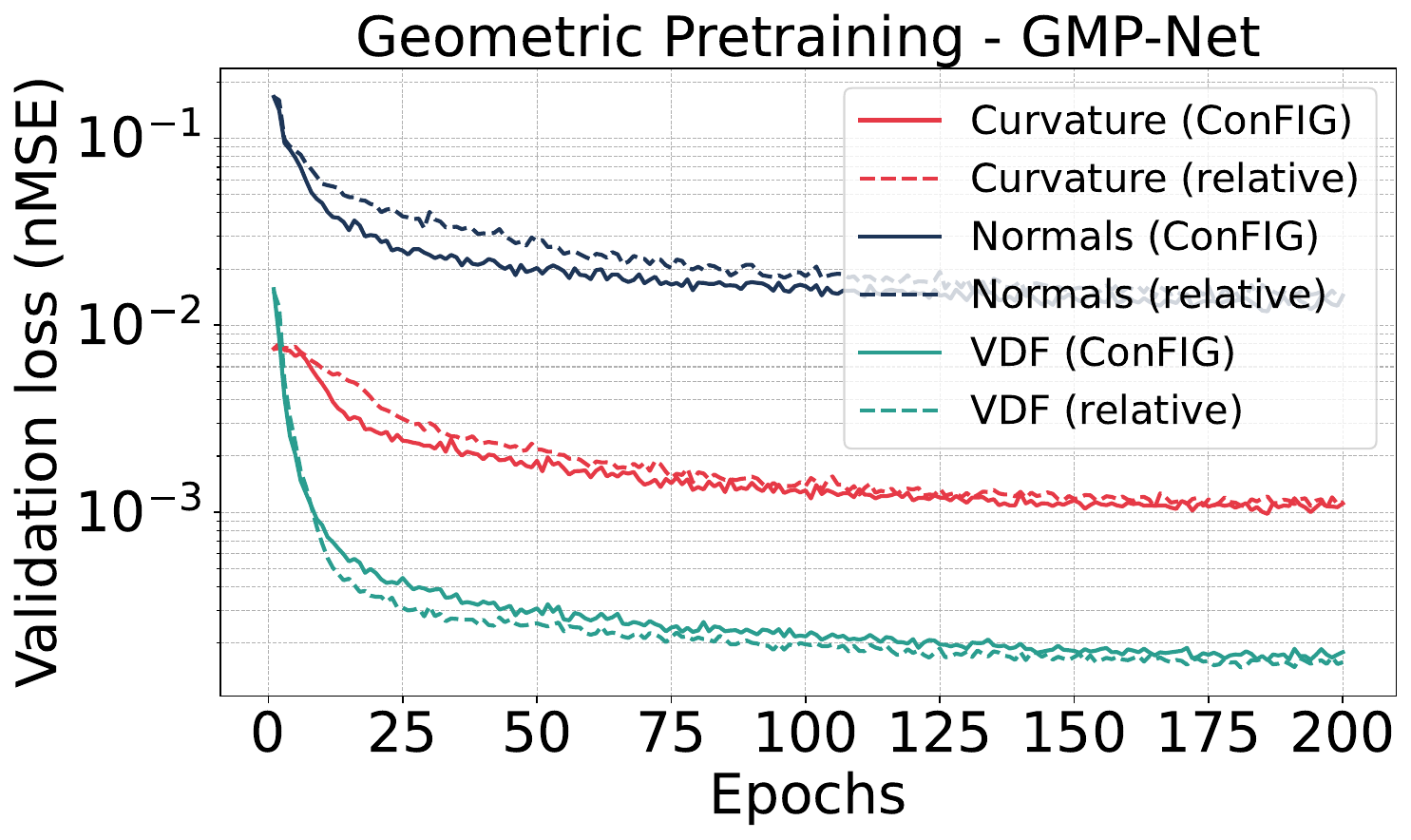}
        \caption{GMP-Net.}
    \end{subfigure}
    \hfill 
    \begin{subfigure}[b]{0.32\textwidth}
        \centering
        \includegraphics[width=\textwidth]{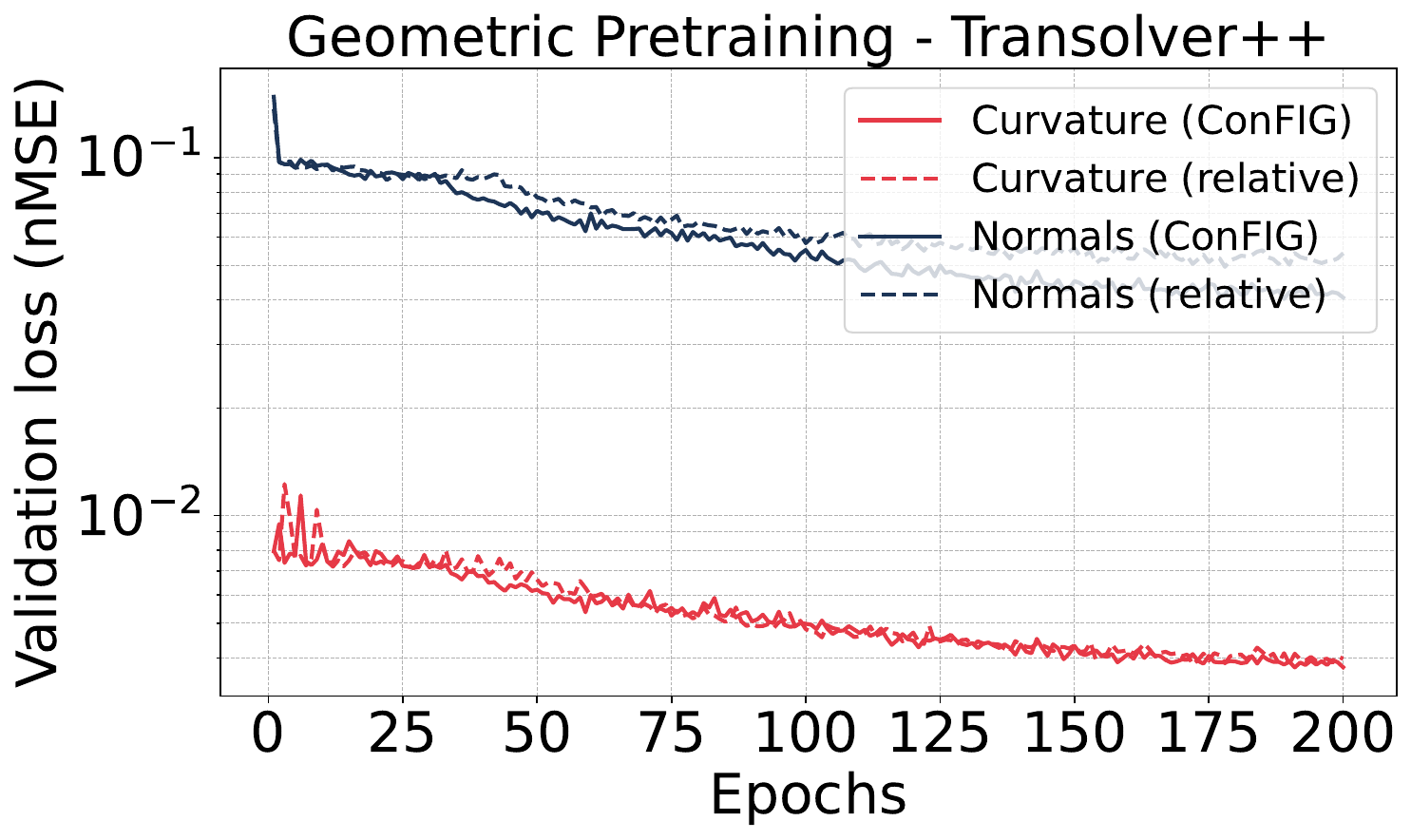}
        \caption{Transolver++.}
    \end{subfigure}
    \hfill
    \begin{subfigure}[b]{0.32\textwidth}
        \centering
        \includegraphics[width=\textwidth]{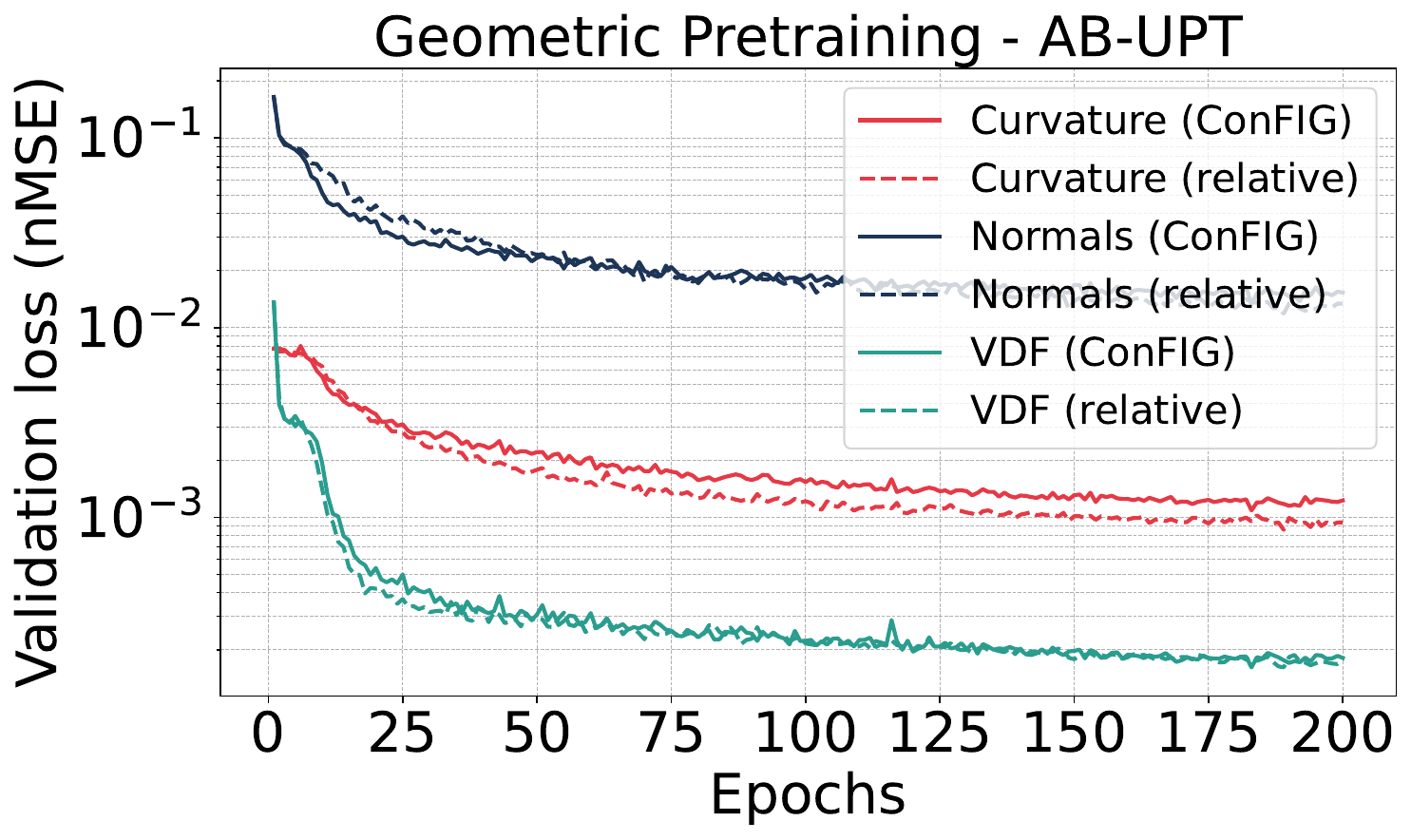}
        \caption{AB-UPT.}
    \end{subfigure}
    \caption{Geometry pre-training.}
    \label{fig:app_geo_pt}
\end{figure}

Figure~\ref{fig:app_geo_pt} illustrates the evolution of each loss component throughout geometric pre-training, showing that ConFIG generally achieves superior convergence across all objectives. A critical question is whether these optimization gains translate into downstream fine-tuning performance. As detailed in Table~\ref{tab:ablation_weights}
, the representations pre-trained with ConFIG consistently yield the lowest prediction errors upon downstream fine-tuning, validating that mitigating gradient conflict during multi-task pre-training produces more transferable geometric features.

\begin{table}[h]
  \caption{Performance evaluation (Relative $L_2$ error) on DrivAerML (64 samples).}
  \label{tab:ablation_weights}
  \centering
  \begin{tabular}{lccc}
    \toprule
    \textbf{Setting} & \textbf{Transolver++} & \textbf{AB-UPT} & \textbf{GMP-Net} \\
    \midrule
    Scratch & 0.329 & 0.244 & 0.187 \\
    GMP relative     & 0.325 & 0.238 & 0.185 \\
    GMP ConFIG  & 0.302 & 0.218 & 0.178 \\
    \midrule 
    Max. Imp    & 8\% & 17\% & 6\% \\
    \bottomrule
  \end{tabular}
\end{table}

\subsection{Family of PDEs}

Following the design philosophy of the Tadpole framework for regular grids \citep{tadpole}, our physics pre-training strategy relies on representation learning via spatial autoencoding rather than learning multi-channel temporal trajectories. Consequently, the primary objective of PDE selection is not to match downstream dynamics, but to expose the spatial autoencoder to a comprehensive set of spatial features, including smooth diffusive gradients, sharp shock interfaces, and multi-scale turbulent vortices. 

\begin{figure}[h!]
     \centering
         \includegraphics[width=0.95\textwidth]{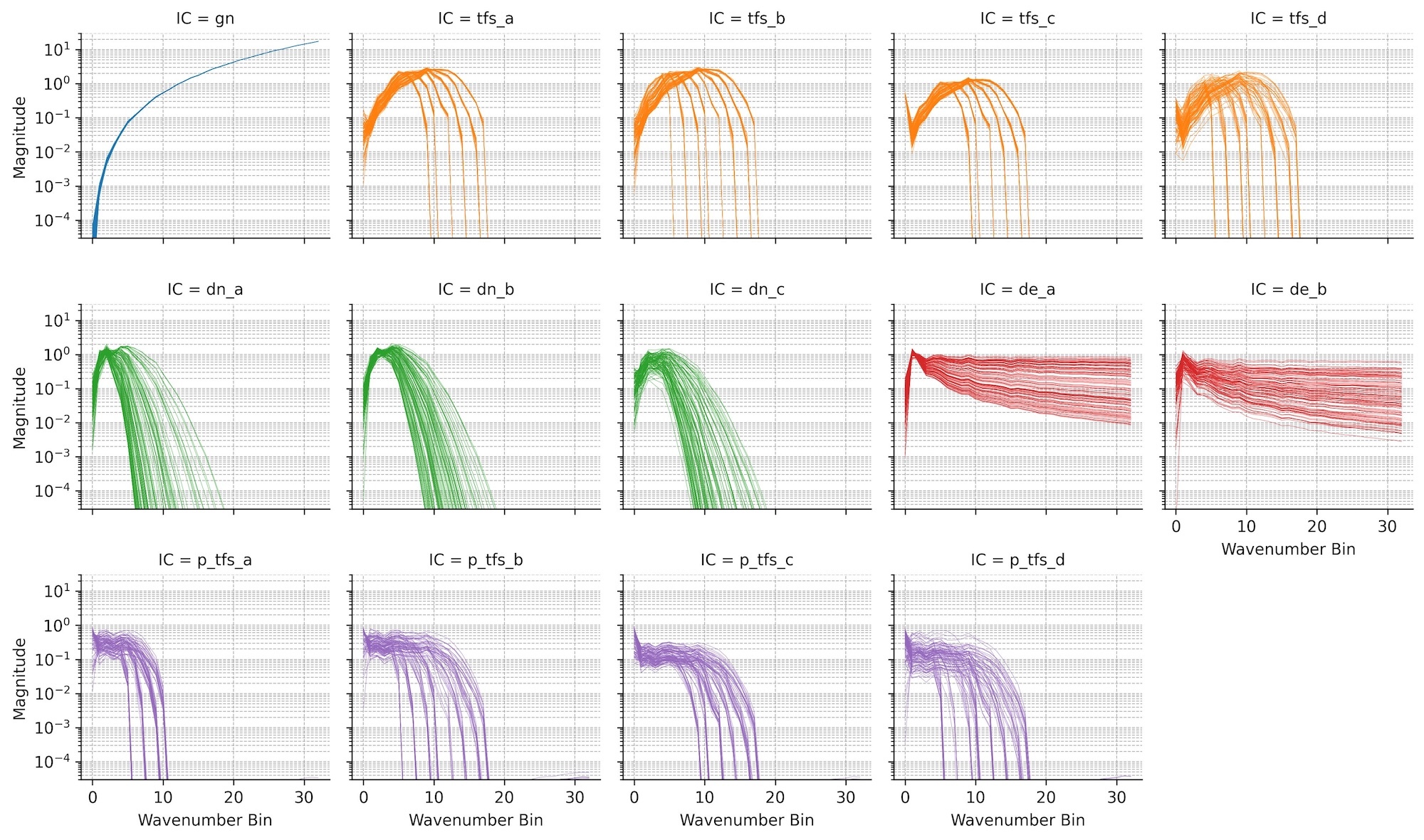}
         \caption{Spectra of different pre-training solutions. Source: \citet{tadpole}.}
         \label{fig:spectral_pdes}
\end{figure}

To achieve complete spectral coverage, the chosen suite of initial conditions (IC) and pre-training equations, e.g., diffusion, advection, wave equations, reaction-diffusion... was curated to cover the full spectrum of spatial wavenumbers. Sampling single-channel spatial crops on-the-fly ensures the representation remains system-agnostic and decoupled from any specific state vector or time-stepping scheme, while preserving power spectral density across high and low frequencies. Figure~\ref{fig:spectral_pdes} shows the spectrum of the solutions to different PDEs with different IC, showing a very varied spectral behavior. Samples are cropped to $32^3$ regions for our experiments.

\subsection{Physics pre-training results}

Because training samples are synthesized dynamically on-the-fly at each iteration, the optimization pipeline operates without a static validation set. Quantitatively, the self-supervised masked reconstruction objective demonstrates high fidelity, reaching a normalized mean squared error (NMSE) on the order of $10^{-4}$. This applies for the pre-training autoencoding objective on single snapshots, without any dynamics. The model is later fine-tuned to learn dynamic evolution between consecutive time steps, naturally a more difficult task.

\subsection{Model size}

\begin{figure}[h!]
    \centering
    \includegraphics[width=0.8\textwidth]{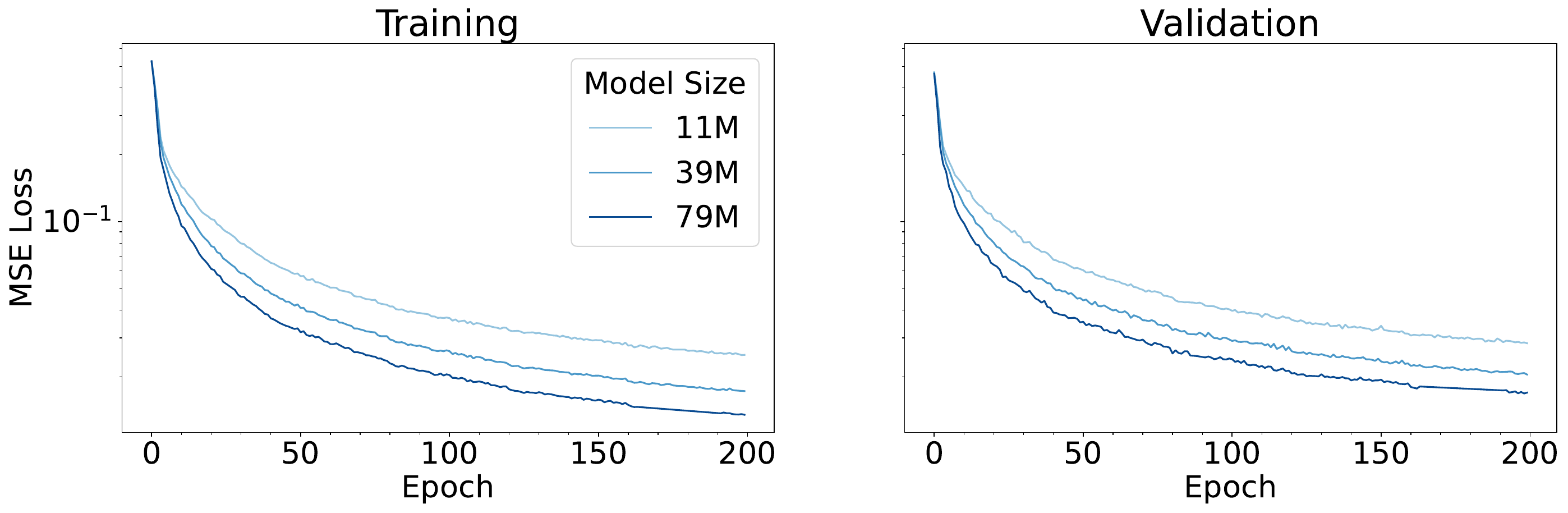}
    \caption{Scaling of the size of SMART.}
    \label{fig:smart_scaling}
\end{figure}

\begin{figure}[h!]
    \centering
    \includegraphics[width=0.8\textwidth]{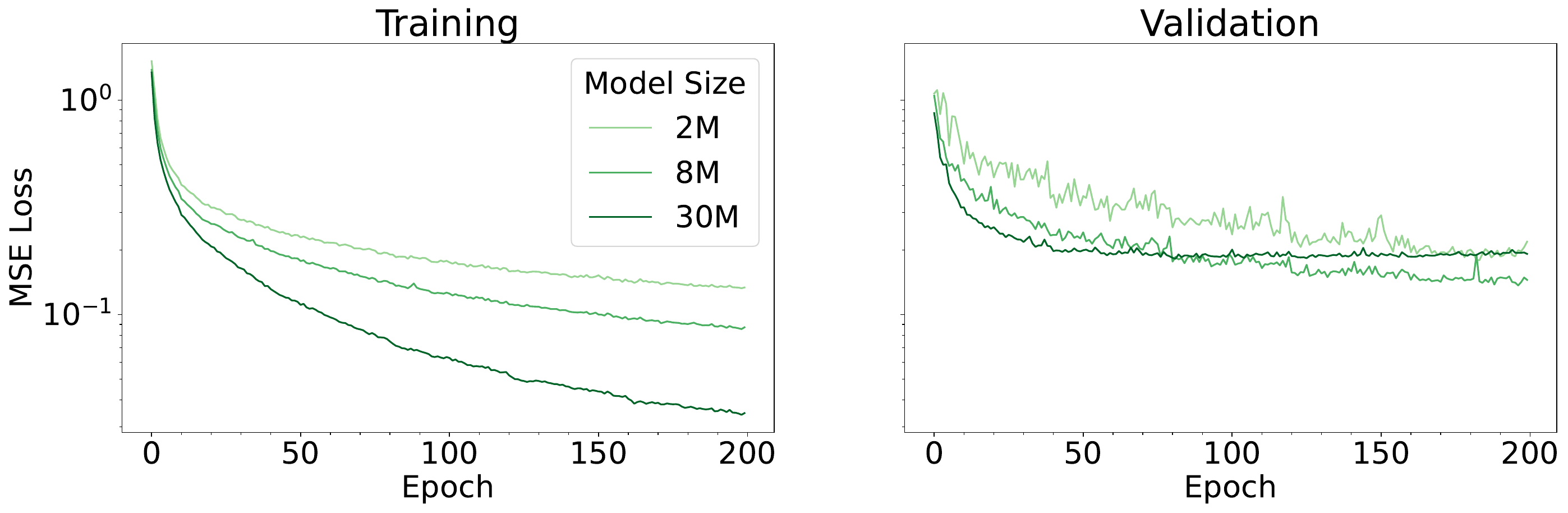}
    \caption{Scaling of the size of Erwin.}
    \label{fig:erwin_scaling}
\end{figure}

The majority of existing unstructured models~\citep{erwin, gaot, abupt, transolver_pp, smart} reported in literature utilize relatively modest parameter budgets, typically ranging between 1--5~M parameters. In contrast, models designed for structured physics data frequently scale to several million~\citep{pdetransformer, p3d} or even billions of parameters~\citep{foundation_walrus}. To investigate the scaling laws within our framework, Figures~\ref{fig:smart_scaling} and~\ref{fig:erwin_scaling} illustrate the performance of two representative architectures, Erwin and SMART, when trained on the DrivAerML dataset. 
For this analysis, we evaluate three variants of SMART containing 11~M (small), 39~M (medium), and 79~M (large) parameters, alongside three versions of Erwin with 2~M (tiny), 8~M (small), and 30~M (medium) parameters. Our observations indicate distinct scaling behaviors: while Erwin appears more challenging to scale—primarily due to a heightened susceptibility to overfitting at higher capacities—SMART demonstrates consistent scaling behavior. However, it should be noted that the performance gains for SMART diminish as capacity increases, yielding only marginal improvements at the highest parameter counts.

\newpage
\subsection{Number of points during fine-tuning}

\begin{figure}[h!]
     \centering
     \begin{subfigure}[c]{0.49\textwidth}
         \centering
         \includegraphics[width=0.9\textwidth]{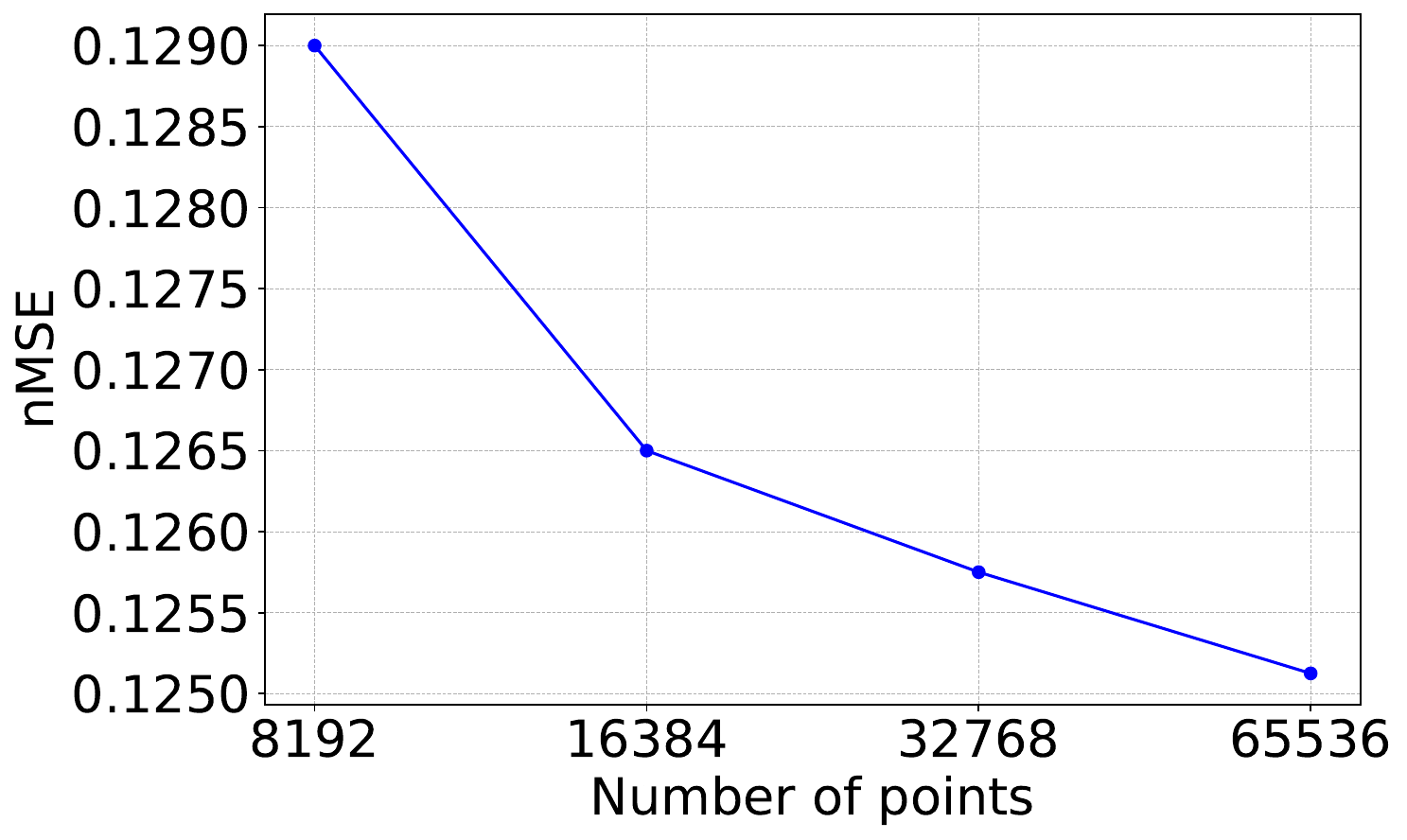}
         \caption{Mean.}
         \label{fig:ablation_num_points_mean}
     \end{subfigure}
     \hfill
     \begin{subfigure}[c]{0.49\textwidth}
         \centering
         \includegraphics[width=0.9\textwidth]{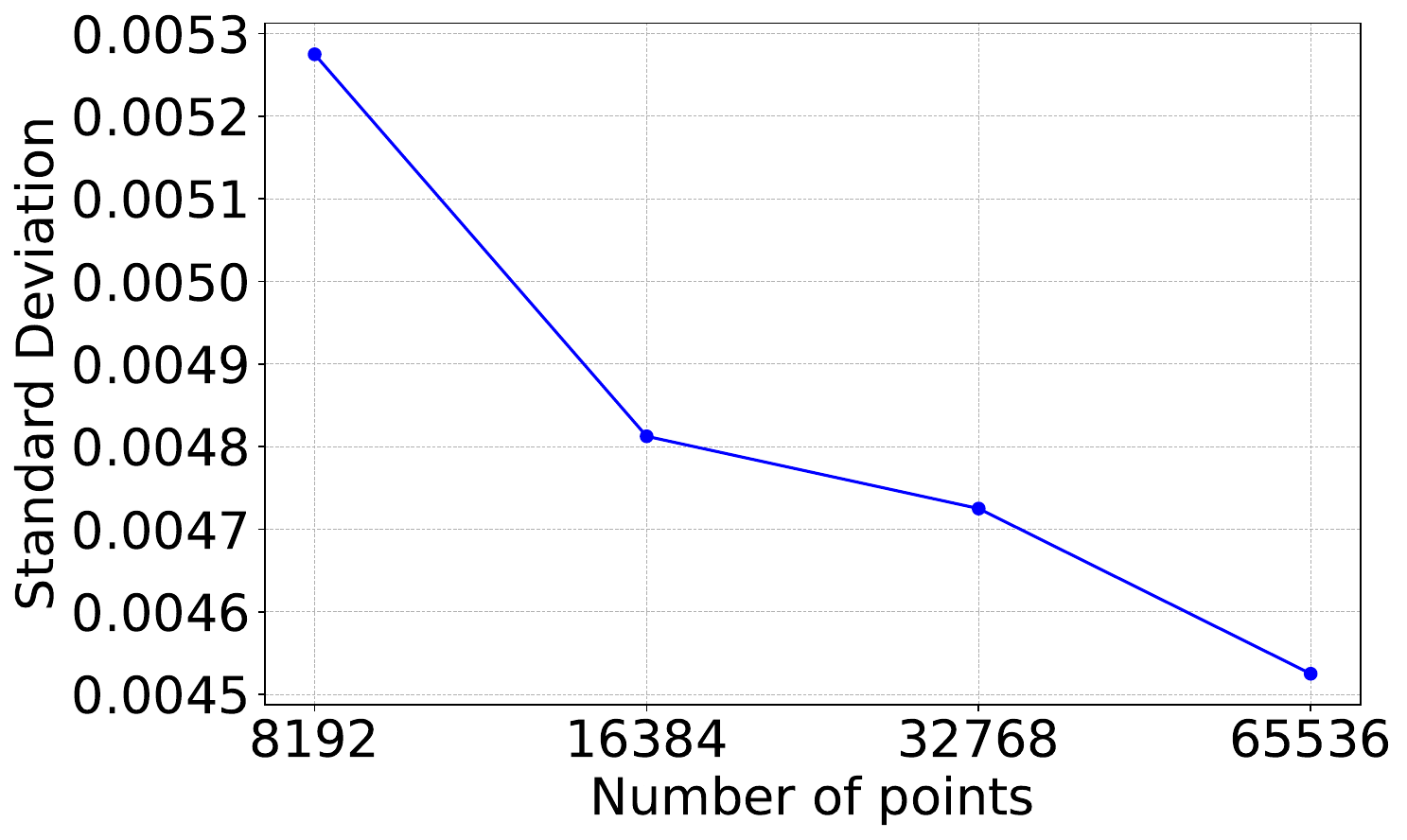}
         \caption{Standard deviation.}
         \label{fig:ablation_num_points_std.pdf}
     \end{subfigure}
     \caption{DrivAerML final test nMSE for different sizes of the point cloud.}
     \label{fig:ablation_num_points}
\end{figure}

In addition, we conduct an ablation study regarding the density of the training point clouds. Current unstructured architectures often encounter out-of-memory (OOM) constraints when processing millions of points simultaneously; consequently, they are typically trained on subsampled representations. To investigate the sensitivity of model performance to this subsampling factor, Figure~\ref{fig:ablation_num_points} illustrates the test nMSE evaluated on the entire samples as a function of the training point count. 
While the results follow the expected downward trend in error, the performance gain for the SMART model is relatively modest, with only a 3\% improvement when increasing the training resolution from $2^{13}$ to $2^{16}$ points. However, a more significant impact is observed in the point-wise error distribution: the standard deviation across points decreases by 14\% over the same range. This suggests that increasing the training point density significantly enhances local stability and reduces the occurrence of high-error outliers.

\FloatBarrier

\section{More on the comparison against other pre-training strategies}\label{app:geopt_comparison}

Although GeoPT \citep{geopt} suggests that static geometric pre-training alone yields negligible downstream gains, our findings demonstrate that static pre-training can substantially improve fine-tuning accuracy without relying on complex synthetic dynamics or high-fidelity CAD models as in \citet{cheap_geometry}. This performance gain is driven by three key components:
\begin{itemize}
    \item \textbf{Enriched geometric descriptors:} Complementing signed distance fields (SDF) or binary occupancy with explicit surface normals and curvature significantly improves representation quality and downstream transfer.
    \item \textbf{Conflict-free optimization via ConFIG:} As demonstrated in our ablation study, scalarized multi-task loss balancing fails to fully exploit the geometric objectives due to conflicting gradient directions. ConFIG resolves this negative transfer across pre-training objectives.
    \item \textbf{Staged fine-tuning via backbone freezing:} Adapting the model to downstream dimensionalities requires training task-specific input and output projection layers from scratch. Implementing an initial warmup phase—where the pre-trained backbone remains frozen—prevents large early gradient updates from corrupting the learned feature space (representational collapse), preserving transfer performance.
\end{itemize}

Table~\ref{tab:steady_final_results} presents a direct comparison against GeoPT as a representative geometric pre-training baseline. To ensure a strictly controlled and equitable evaluation, we executed both the GeoPT pre-training phase and subsequent fine-tuning within our experimental framework, enforcing identical data partitions, backbone architectures, and optimization hyperparameters. Because the official GeoPT repository omits the end-to-end pre-training pipeline required to reproduce their published benchmarks out of the box, we re-implemented their pre-training methodology adhering strictly to the specifications detailed in their manuscript. While minor numerical discrepancies are unavoidable in the absence of an end-to-end public release, this unified setup ensures a faithful, apples-to-apples comparison. For completeness, Table~\ref{tab:geopt_real_results} compiles the original numerical results reported by the GeoPT authors. These results are not directly comparable to ours. Although we both consider three similar downstream scenarios (car aerodynamics, aircraft aerodynamics, and automotive crash tests), only DrivAerML uses exactly the same dataset, while the other two are different. Besides, model sizes, train/test data splits, the particular output fields predicted, and other hyperparameters may also have a major influence on the final results. Despite these differences, the comparison still provides an informative external reference point.

\begin{table}[h!]
    \centering
    \caption{GeoPT comparison of relative $L_2$ error between Transolver (scratch) and GeoPT at 200 epochs, alongside relative improvement across varying sample sizes. Source:~\citet{geopt}.}
    \label{tab:geopt_real_results}
    \setlength{\tabcolsep}{5pt}
    \begin{tabular}{l *{6}{c}} 
    \toprule
    & \multicolumn{2}{c}{DrivAerML} & \multicolumn{2}{c}{NASA CRM} & \multicolumn{2}{c}{Car Crash} \\
    \cmidrule(lr){2-3} \cmidrule(lr){4-5} \cmidrule(lr){6-7}
    Training samples & 32 & 64 & 32 & 64 & 32 & 64 \\ 
    \midrule
    Transolver (scratch) $\downarrow$ & 0.166 & 0.110 & 0.242 & 0.136 & 0.315 & 0.214 \\
    Transolver (GeoPT) $\downarrow$   & 0.126 & 0.090 & 0.203 & 0.113 & 0.235 & 0.192 \\
    \midrule
    Rel. Improvement $\uparrow$ & \cellcolor{Green!24}24\% & \cellcolor{Green!18}18\% & \cellcolor{Green!16}16\% & \cellcolor{Green!17}17\% & \cellcolor{Green!25}25\% & \cellcolor{Green!10}10\% \\
    \bottomrule
    \end{tabular}
\end{table}

\section{Training hyperparameters}\label{app:hyperparameters}

We first detail the physics pre-training hyperparameters in Table~\ref{tab:hyperparams_physics}, where $H'$, $W'$, and $D'$ denote the final crop dimensions of the regular grid data. Since the underlying PDE data is generated at varying resolutions, this cropping procedure ensures a uniform input format across the pipeline. To ensure a broad range of spectral behaviors, we employ the diverse PDE types and initial conditions provided in the Tadpole framework~\citep{tadpole}. These data generation details are particularly salient, as our online generation process represents a significant departure from the conventional disk-based training pipelines described in subsequent tables. For instance, since data is constantly being generated, there is no fixed definition of an epoch, and we simply assign it to 1000 samples.

Additionally, Table~\ref{tab:hyperparams_geo} provides the hyperparameters for geometric pre-training, e.g., the number of points used during pre-training: 1024 on the surface, and 2048 in the volume. These values are much smaller than those in the fine-tuning datasets (DrivAerML and SHIFT-Wing), but pre-training still captures relevant geometric information.

Table~\ref{tab:hyperparams_transient} covers the three transient benchmarks: the 2D Kuramoto-Sivashinsky (KS) equation, 2D flow around an ellipse, and 3D SHIFT-Crash, and Tables~\ref{tab:hyperparams_drivaerml}, \ref{tab:hyperparams_shiftwing} and~\ref{tab:hyperparams_shifcrash} summarize the settings for the steady-state cases. For brevity, model-specific configurations are omitted from these summary tables and are instead detailed in the following section of the Appendix.

\begin{table}[h!]
    \centering
    \caption{Hyperparameters for the physics pre-training.}
    \label{tab:hyperparams_physics}
    \begin{tabular}{ll p{6cm}} 
        \toprule
        Category & Hyperparameter & Value / Description \\
        \midrule
        Data generation & H', W', D'    & 32 \\
                        & PDEs            & Diffusion, Hyper-diffusion, Burgers, Korteweg-de Vries (KdV), Kuramoto-Sivashinsky (KS), Fisher-KPP, Swift-Hohenberg\\
                        & Initial conditions & Gaussian noise, Truncated Fourier series, Diffused noise, Decayed Energy, Poisson\\
                        & Resolution        & 64, 128, 256, 384\\
                        & \# surface points & 1024 \\
                        & \# volume points & 2048 \\
                        & \# data GPUs      & 1 \\
        \midrule
        Optimization & Optimizer & AdamW \\
                     & Learning Rate & 1e-4 \\
                     & LR schedule & cosine \\
                     & Batch Size & 32 \\
                     & Weight Decay & $10^{-5}$ \\
        \midrule
        Training     & Epochs & 200 \\
                     & Samples per epoch & 1000 \\
                     & Training time & 42 hours \\
                     & Precision & Single (FP32) \\
                     & Hardware & 2 x NVIDIA RTX A5000 Ampere (24 GB) \\
        \bottomrule
    \end{tabular}
\end{table}

\begin{table}[h!]
    \centering
    \caption{Hyperparameters for the geometric pre-training.}
    \label{tab:hyperparams_geo}
    \begin{tabular}{ll p{6cm}} 
        \toprule
        Category & Hyperparameter & Value / Description \\
        \midrule
        Data            & \# surface points & 1024 \\
                        & \# volume points & 2048 \\
        \midrule
        Optimization & Optimizer & AdamW \\
                     & Learning Rate & 1e-4 \\
                     & LR schedule & cosine \\
                     & Batch Size & 8 \\
                     & Weight Decay & $10^{-5}$ \\
        \midrule
        Training     & Epochs & 200 \\
                     & Samples per epoch & 1000 \\
                     & Training time & 4 hours \\
                     & Precision & Single (FP32) \\
                     & Hardware & 1 x NVIDIA A100 \\
        \bottomrule
    \end{tabular}
\end{table}

\begin{table}[h!]
    \centering
    \caption{Hyperparameters for autoregressive training.}
    \label{tab:hyperparams_transient}
    \begin{tabular}{ll p{6cm}} 
        \toprule
        Category & Hyperparameter & Value / Description \\
        \midrule
        Data            & Training/Test split & KS: 580/50 trajectories \\
                        &                     & Ellipse: 1000/200 trajectories \\
                        &                     & Crash: 600/100 trajectories \\
        \midrule
        Optimization & Optimizer & AdamW \\
                     & Max learning Rate & 1e-4 \\
                     & Lr. schedule & Cosine \\
                     & Batch Size & 8 \\
        \midrule
        Training     & Epochs & 20 \\
                     & Weights warmup   & 1 epoch \\
                     & Training time & 8 hours (for GMP-Net) \\
                     & Precision & Single (FP32) \\
                     & Hardware & 1 x NVIDIA RTX A5000 Ampere (24 GB) \\
        \bottomrule
    \end{tabular}
\end{table}

\begin{table}[h!]
    \centering
    \caption{Hyperparameters for steady DrivAerML.}
    \label{tab:hyperparams_drivaerml}
    \begin{tabular}{ll p{4cm}}
        \toprule
        Category & Hyperparameter & Value / Description \\
        \midrule
        Data            & Training/Test split & (Varying size from 16 to 128) \\
                        & \# points           & 32768 (S) + 32768 (V)\\
        \midrule
        Optimization & Optimizer & LION \\
                     & Learning Rate & 1e-4 \\
                     & LR schedule & Cosine \\
                     & Batch Size & 1 \\
        \midrule
        Training     & Epochs & 200 \\
                     & Training time & 12 hours (max for GMP-Net) \\
                     & Precision & Single (FP32) \\
                     & Hardware & 1 x NVIDIA L40S \\
        \bottomrule
    \end{tabular}
\end{table}

\begin{table}[h!]
    \centering
    \caption{Hyperparameters for steady SHIFT-Wing.}
    \label{tab:hyperparams_shiftwing}
    \begin{tabular}{ll p{4cm}} 
        \toprule
        Category & Hyperparameter & Value / Description \\
        \midrule
        Data            & Training/Test split & (Varying size from 16 to 128) \\
                        & \# points           & 32768 (S) + 32768 (V) \\
        \midrule
        Optimization & Optimizer & LION \\
                     & Learning Rate & 1e-4 \\
                     & LR schedule & Cosine \\
                     & Batch Size & 1 \\
        \midrule
        Training     & Epochs & 50 \\
                     & Training time & 20 hours (max for GMP-Net) \\
                     & Precision & Single (FP32) \\
                     & Hardware & 1 x NVIDIA A100 \\
        \bottomrule
    \end{tabular}
\end{table}

\begin{table}[h!]
    \centering
    \caption{Hyperparameters for steady SHIFT-Crash.}
    \label{tab:hyperparams_shifcrash}
    \begin{tabular}{ll p{4cm}} 
        \toprule
        Category & Hyperparameter & Value / Description \\
        \midrule
        Data            & Training/Test split & (Varying size from 16 to 128) \\
                        & \# points           & 32768\\
        \midrule
        Optimization & Optimizer & LION \\
                     & Learning Rate & 1e-4 \\
                     & LR schedule & Cosine \\
                     & Batch Size & 1 \\
        \midrule
        Training     & Epochs & 50 \\
                     & Training time & 16 hours (max for GMP-Net) \\
                     & Precision & Single (FP32) \\
                     & Hardware & 1 x NVIDIA H100 \\
        \bottomrule
    \end{tabular}
\end{table}

\clearpage
\section{Baselines hyperparameters}\label{app:models}

This section provides a comprehensive overview of the configurations for all models and baselines evaluated in this work, as detailed in Tables~\ref{tab:hyperparams_smart_steady}, \ref{tab:hyperparams_smart_transient}, \ref{tab:hyperparams_abupt}, and \ref{tab:hyperparams_transolver}. The disclosure of these hyperparameters is intended to ensure full transparency and to provide the necessary technical specifications to facilitate the exact reproduction of our empirical results.

\begin{table}[h!]
    \centering
    \caption{GMP-Net/SMART model configuration for steady problems.}
    \label{tab:hyperparams_smart_steady}
    \begin{tabular}{l p{5cm}} 
        \toprule
        Hyperparameter & Value / Description \\
        \midrule
        spatial dim                 & 3 \\
        Surface channels            & 4 \\
        Volume channels             & 4 \\
        Parameter channels          & 16 (DrivAerML) \\
                  & 1 (SHIFT-Wing) \\
                  & 6 (SHIFT-Crash) \\
        Latent dim                  & 256 \\
        Latent geometry points      & 4096 \\
        Subsampled geometry points  & 16384 \\
        Num encoder/decoder blocks  & 8 \\
        Num heads                   & 8 \\
        pos scale factor            & 1000 \\
        dropout                     & 0.0 \\
        
        \bottomrule
    \end{tabular}
\end{table}

\begin{table}[h!]
    \centering
    \caption{GMP-Net model configuration for transient problems.}
    \label{tab:hyperparams_smart_transient}
    \begin{tabular}{l p{5cm}} 
        \toprule
        Hyperparameter & Value / Description \\
        \midrule
        spatial dim                 & 2 (KS) / 2 (Ellipse) / 3 (Crash) \\
        Input surface channels      & 0 (KS) / 0 (Ellipse) / 1 (Crash) \\
        Input volume channels       & 1 (KS) / 3 (Ellipse) / 0 (Crash) \\
        Output surface channels     & 0 (KS) / 0 (Ellipse) / 1 (Crash) \\
        Output volume channels      & 1 (KS) / 3 (Ellipse) / 0 (Crash) \\
        Parameter channels          & 1 \\
        Latent dim                  & 512 \\
        Latent geometry points      & 4096 \\
        Subsampled geometry points  & 4096 \\
        Num encoder/decoder blocks  & 1 \\
        Num heads                   & 16 \\
        pos scale factor            & 1000 \\
        dropout                     & 0.0 \\
        
        \bottomrule
    \end{tabular}
\end{table}

\begin{table}[h!]
    \centering
    \begin{minipage}{0.45\textwidth}
    \centering
    \caption{AB-UPT model configuration.}
    \label{tab:hyperparams_abupt}
    \begin{tabular}{l p{3cm}} 
        \toprule
        Hyperparameter & Value / Description \\
        \midrule
        ndim                & 3 \\
        input dim           & 3 \\
        output dim surface  & 4 \\
        output dim volume   & 4 \\
        dim                 & 192 \\
        geometry depth      & 3 \\
        num heads           & 8 \\
        blocks              & pscscs \\
        num volume blocks   & 8 \\
        num surface blocks  & 8 \\
        radius              & 0.25 \\
        supernode fraction  & 0.125 \\
        num anchors         & 8192 \\
        
        \bottomrule
    \end{tabular}
    \end{minipage}
    \hfill
    \begin{minipage}{0.5\textwidth}
        \centering
    \caption{Transolver++ model configuration.}
    \label{tab:hyperparams_transolver}
    \begin{tabular}{l p{3cm}} 
        \toprule
        Hyperparameter & Value / Description \\
        \midrule
        space dim           & 3 \\
        n layers            & 6 \\
        n hidden            & 512 \\
        n heads             & 16 \\
        dropout             & 0.0 \\
        activation          & gelu \\
        mlp ratio           & 1.0 \\
        fun dim             & 0 \\
        out dim             & 4 \\
        slice num           & 64 \\
        condition size          & 16 (DrivAerML) \\
                  & 1 (SHIFT-Wing) \\
                  & 6 (SHIFT-Crash) \\
        \bottomrule
    \end{tabular}
    \end{minipage}
\end{table}

\FloatBarrier

\section{Base architecture selection}\label{app:model_selection}

Although our physics and geometric pre-training setups could be applied independently to any backbone architecture, it is essential to evaluate these strategies using the most competitive architectures available. Establishing robust pre-training frameworks for high-performing architectures represents a more rigorous and relevant challenge than improving lower-performing baselines that naturally possess more headroom for improvement.
Below, we compare several recent models \citep{abupt, smart, erwin, transolver_pp} to establish an architectural basis. To ensure a fair comparison, all models are configured with approximately 10M parameters. One key distinction must be noted: while SMART~\citep{smart} and AB-UPT~\citep{abupt} are designed to process surface and volume data simultaneously, other models do not inherently distinguish between the two. Consequently, we train those models separately for surface and volume tasks with a halved parameter count to maintain a consistent total.
To evaluate these models, we utilize some of the most established high-fidelity datasets in the community: DrivAerML~\citep{drivaerml}, SHIFT-Wing~\citep{shift_wing} and SHIFT-Crash ~\citep{shift_crash}. 
As our primary metric, we use the relative $L_2$ error.
All models are trained by randomly subsampling the full point cloud to $2^{15} = 32,768$ points for surface and volume.
As SMART consistently demonstrates superior accuracy across both datasets for both surface and volume data, we focus the analysis on this architecture.
Table \ref{tab:unified_benchmarks_l2} shows the exact averaged results. Finally, Tables \ref{tab:app_ks}, \ref{tab:app_ellipse} and \ref{tab:app_crash} show the autoregressive results from the main manuscript with the corresponding standard deviations.

\begin{table}[htbp]
    \caption{Test relative $L_2$ error (\%) averaged across surface and volume fields for DrivAerML, SHIFT-Wing, and SHIFT-Crash benchmarks.}
    \label{tab:unified_benchmarks_l2}
    \centering
    \small
    \setlength{\tabcolsep}{8pt}
    \begin{tabular}{lccccc}
        \toprule
        \multirow{2}{*}{Model} & \multicolumn{2}{c}{DrivAerML} & \multicolumn{2}{c}{SHIFT-Wing} & SHIFT-Crash$^*$ \\ 
        \cmidrule(lr){2-3} \cmidrule(lr){4-5} \cmidrule(lr){6-6}
         & Surface & Volume & Surface & Volume & Surface \\ 
        \midrule
        PTV3          & 31.33 & 44.31 & 20.87 & 46.76 & 35.25 \\ 
        Erwin         & 28.52 & 26.67 & 11.52 & 38.67 & 42.15 \\ 
        Transolver++  & 11.32 & 13.04 &  7.67 & 24.04 & 23.97 \\ 
        GeoTransolver & 10.33 & 10.90 &  7.50 & 24.90 & 23.39 \\ 
        AB-UPT        & 10.76 &  8.83 & 10.07 & 25.11 & 39.65 \\ 
        SMART         &  7.93 &  7.01 &  7.95 &  9.36 & 20.56 \\ 
        \bottomrule
    \end{tabular}
\end{table}

\begin{table}[h!]
    \caption{2D KS equation. Mean RMSE for timesteps 1, 5, and 10.}
    \label{tab:app_ks}
    \centering
    \begin{tabular}{lccc}
        \toprule
        Model & 1 & 5 & 10 \\ 
        \midrule
        Erwin           & 0.143 $\pm$ 0.011 & 0.713 $\pm$ 0.043 & 1.281 $\pm$ 0.073 \\
        Transolver++    & 0.203 $\pm$ 0.014 & 0.779 $\pm$ 0.052 & 1.028 $\pm$ 0.035 \\ 
        \midrule
        GMP-Net (scratch)           & 0.030 $\pm$ 0.003 & 0.196 $\pm$ 0.039 & 0.615 $\pm$ 0.060 \\
        GMP-Net (PT)          & 0.024 $\pm$ 0.002 & 0.148 $\pm$ 0.030 & 0.479 $\pm$ 0.073 \\ 
        \bottomrule
    \end{tabular}
\end{table}

\begin{table}[h!]
    \caption{2D flow around an ellipse. Mean RMSE for timesteps 1, 5, and 10.}
    \label{tab:app_ellipse}
    \centering
    \begin{tabular}{lccc}
        \toprule
        Model & 1 & 5 & 10 \\ 
        \midrule
        Erwin           & 0.016 $\pm$ 0.002 & 0.083 $\pm$ 0.034 & 0.438 $\pm$ 0.244 \\
        Transolver++    & 0.006 $\pm$ 0.001 & 0.021 $\pm$ 0.003 & 0.042 $\pm$ 0.005 \\
        \midrule
        GMP-Net (scratch)          & 0.008 $\pm$ 0.001 & 0.025 $\pm$ 0.043 & 0.044 $\pm$ 0.009 \\
        GMP-Net (PT)          & 0.006 $\pm$ 0.001 & 0.020 $\pm$ 0.004 & 0.033 $\pm$ 0.011 \\ 
        \bottomrule
    \end{tabular}
\end{table}

\begin{table}[h!]
    \caption{Deformation and von Mises stress average of SHIFT-Crash. Mean RMSE for timesteps 1, 5, and 10.}
    \label{tab:app_crash}
    \centering
    \begin{tabular}{lccc}
        \toprule
        Model & 1 & 5 & 10 \\ 
        \midrule
        Erwin           & 0.039 $\pm$ 0.001 & 0.131 $\pm$ 0.034 & 0.245 $\pm$ 0.021\\
        Transolver++    & 0.020 $\pm$ 0.016 & 0.072 $\pm$ 0.039 & 0.072 $\pm$ 0.021 \\ 
        \midrule
        GMP-Net (scratch)           & 0.009 $\pm$ 0.008 & 0.030 $\pm$ 0.017 & 0.040 $\pm$ 0.024 \\
        GMP-Net (PT)          & 0.008 $\pm$ 0.007 & 0.026 $\pm$ 0.018 & 0.034 $\pm$ 0.013 \\  
        \bottomrule
    \end{tabular}
\end{table}

To continue the arguments on the selection of the backbone architecture, a summary of computational efficiency is provided in Table~\ref{tab:model_efficiency}. This is tested with subsampled DrivAerML samples on an NVIDIA L40S GPU. While Erwin and Transolver++ exhibit significantly better efficiency, the superior accuracy demonstrated by SMART makes it the most compelling candidate for our pre-training framework.

\begin{table}[h!]
    \centering
    \caption{Model complexity and computational efficiency comparison (batch size 1).}
    \label{tab:model_efficiency}
    \begin{tabular}{@{}l ccc@{}}
        \toprule
        Model & Size (M) $\downarrow$ & VRAM (GB) $\downarrow$ & Throughput (it/s) $\uparrow$ \\ 
        \midrule
        Erwin \citep{erwin}                  & \best{6.8} & \best{2.15} & \textbf{2.45} \\
        Transolver++ \citep{transolver_pp}   & 6.9 & \textbf{3.50} & \best{3.27} \\
        AB-UPT \citep{abupt}                 & 11.4 & 12.79 & 0.47 \\
        SMART \citep{smart}                  & 11.4 & 9.58 & 0.70 \\ 
        \bottomrule
    \end{tabular}
\end{table}

\section{Aerodynamic forces calculation}\label{app:aero_forces}

For real-world applications, the calculation of the aerodynamic forces acting on an object is the ultimate goal. The total aerodynamic force $\textbf{F}$ is driven by pressure differences and viscous stress. Together, the total force is computed via a surface integral that is numerically solved with a simple quadrature rule
\begin{equation}
   \textbf{F} = \int_S (-p \textbf{n} + \tau)dS \approx \sum_i (-p_i \textbf{n}_i + \tau_i)S_i
\end{equation}
where we sum over all the surface cells, $p_i$ and $\tau_i$ are the pressure and wall shear stress at the center of the $i$-th cell, $\textbf{n}_i$ its normal and $S_i$ its area. Then, the drag is simply the horizontal force along the moving direction, and lift the vertical component. For aerospace applications, we must additionally account for the angle of attack.

\begin{figure}[h!]
     \centering
     \begin{subfigure}[c]{0.49\textwidth}
         \centering
         \includegraphics[width=0.8\textwidth]{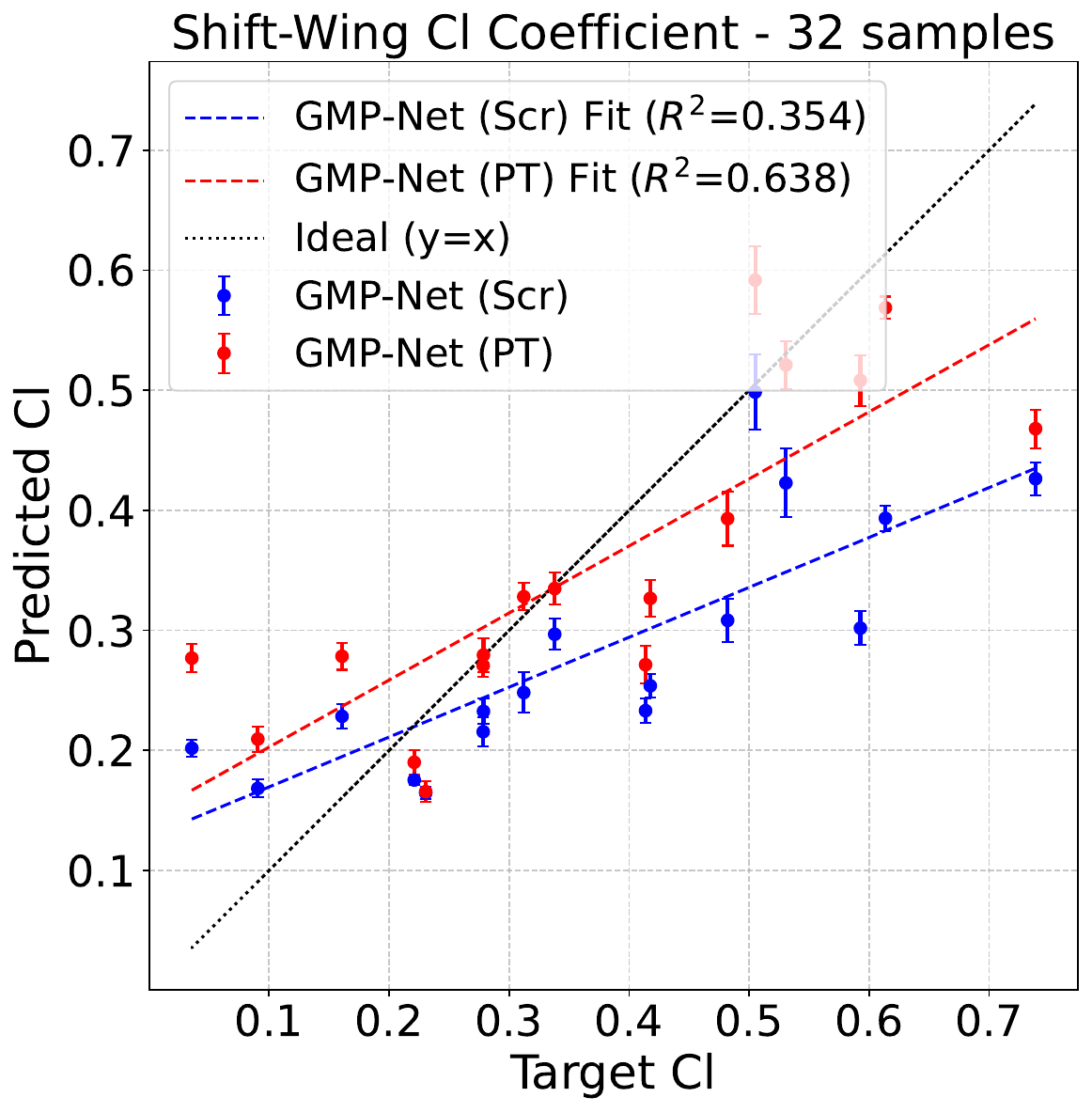}
         \caption{$C_l$ for 32 samples.}
     \end{subfigure}
     \hfill
     \begin{subfigure}[c]{0.49\textwidth}
         \centering
         \includegraphics[width=0.8\textwidth]{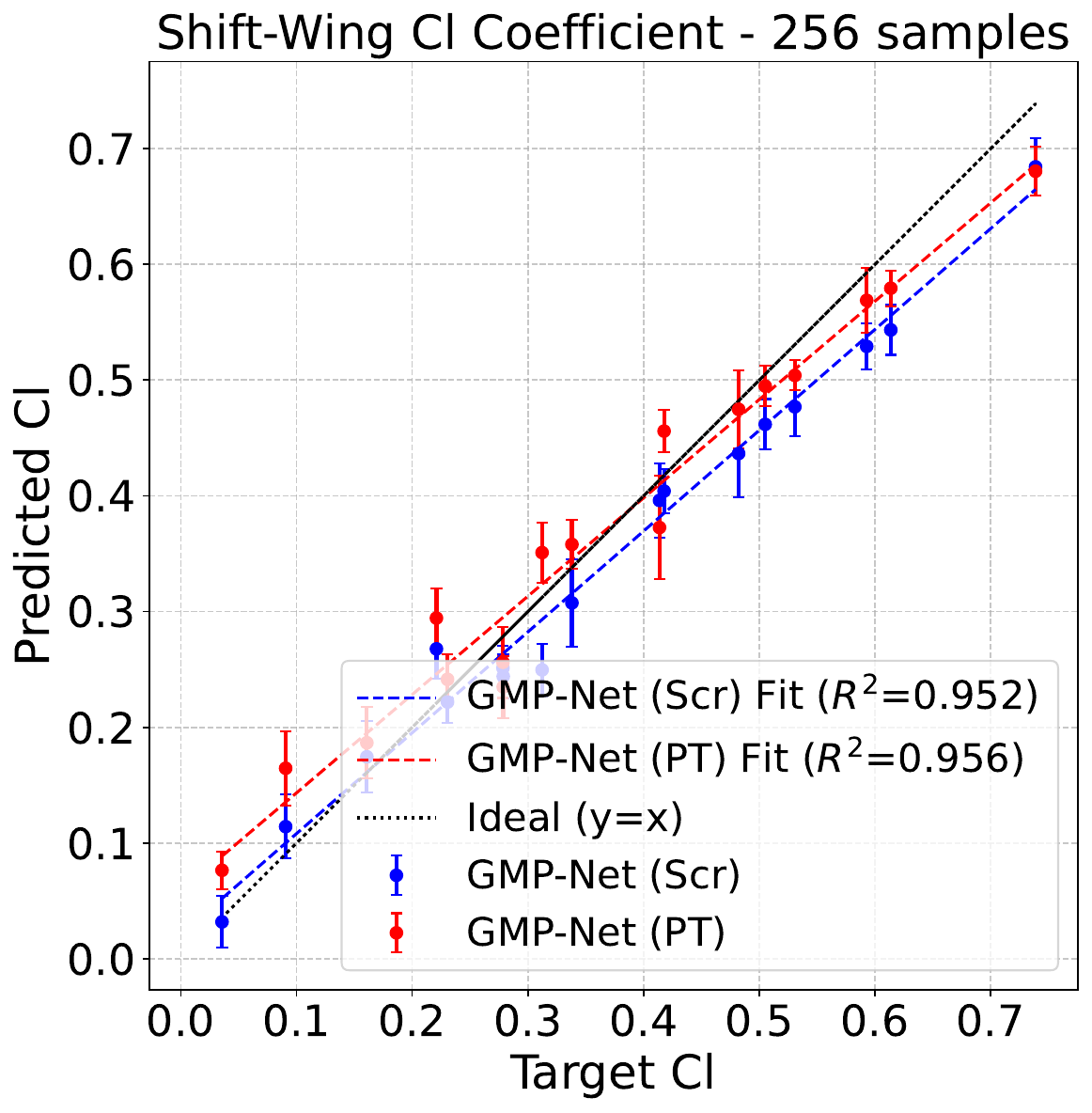}
         \caption{$C_l$ for 256 samples.}
     \end{subfigure}
     \caption{Lift coefficient computation for SHIFT-Wing test samples when using 32 and 256 training samples. While the 32 case seems to be too sparse to accurately predict the aerodynamic forces of test samples, both cases clearly show increased accuracy of our GMP pre-training.}
     \label{fig:aero_forces_visual}
\end{figure}

Figure \ref{fig:aero_forces_visual} shows the lift coefficient for SHIFT-Wing test samples when using 32 and 256 training samples. It provides a comparison between the trained from scratch GMP-Net, and our pre-trained results. While training with only 32 samples is too few to accurately predict the aerodynamic forces of test samples, both cases clearly show increased accuracy of our GMP pre-training.

\FloatBarrier

\section{Intersection-free meshes via Gaussian random fields} \label{app:gaussianDefo}

As analytic mesh generation component if $\mathcal{G}$ in Section~\ref{geometric_pretraining} we employ intersection-free deformations generated by frequency controlled Gaussian random fields. The following section details this step in two dimensions for brevity, it naturally extends to three spatial dimensions.

The construction follows three key steps: (1) Generate smooth Gaussian random fields $(u, v)$. (2) Normalize them to satisfy a global Lipschitz bound ($|D\mathbf{d}| < 1$). (3) Apply the deformation to the grid nodes.
This yields a random but globally valid deformation, combining smoothness (from the Gaussian field), randomness (from spectral sampling), and strict geometric guarantees (from the Lipschitz constraint).

We consider a regular Cartesian grid over a rectangular domain
$
\Omega = [0, L_x] \times [0, L_y],
$
with grid points
$
(x_i, y_j), \quad i = 1,\dots,N,; j = 1,\dots,M
$.
Our goal is to construct a random, smooth deformation
\begin{equation}
\Phi : \Omega \rightarrow \mathbb{R}^2, \quad \Phi(x,y) = (x,y) + \mathbf{d}(x,y),
\end{equation}
such that the resulting mesh is non-overlapping and orientation-preserving, i.e.
\begin{equation}
\det D\Phi(x,y) > 0 \quad \forall (x,y) \in \Omega.
\end{equation}

We define the displacement field
$
\mathbf{d}(x,y) = \begin{pmatrix} u(x,y) \ v(x,y) \end{pmatrix},
$
where $u$ and $v$ are independent realizations of smooth Gaussian random fields.
They are constructed via a spectral representation:
\begin{equation}
u(x,y) = \mathcal{F}^{-1}\left( \hat{\eta}(k_x,k_y), \sqrt{S(k_x,k_y)} \right),
\end{equation}
where:
$\hat{\eta}$ is complex white noise in Fourier space, and
$S(k_x,k_y)$ is a power spectrum of the form
  $
  S(k_x,k_y) = \exp\big(-\ell^2 (k_x^2 + k_y^2)\big),
  $
  which enforces smoothness with correlation length $\ell$.
The same construction is used for $v(x,y)$.
The deformation gradient (Jacobian) is given by
\begin{equation}
D\Phi(x,y) = I + D\mathbf{d}(x,y)
\begin{pmatrix}
1 + \partial_x u & \partial_y u \\
\partial_x v & 1 + \partial_y v
\end{pmatrix}.
\end{equation}
with 
the determinant 
$\det D\Phi = (1 + \partial_x u)(1 + \partial_y v) - (\partial_y u)(\partial_x v).$
Then a necessary and sufficient condition for no local inversion is 
$\det D\Phi(x,y) > 0 \quad \forall (x,y).$
Instead of directly enforcing positivity of the determinant, we impose a stronger but simpler condition:

\begin{equation}
| D\mathbf{d}(x,y) |_2 \le L < 1 \quad \forall (x,y).
\end{equation}

This implies that all eigenvalues of $D\mathbf{d}$ satisfy $|\lambda_i| < 1,$
$
\det D\Phi > 0$ , and thus $\Phi$
is locally invertible everywhere.
This condition ensures global injectivity on $\Omega$, i.e. no self-intersections.

Given a sampled displacement field, we estimate
the maximum $L$ via a Frobenius norm and then 
rescale the field:
\begin{equation}
\tilde{\mathbf{d}}(x,y) = \frac{L_{\text{target}}}{L_{\max} + \varepsilon}  \mathbf{d}(x,y).
\end{equation}
This ensures that the final deformation
$\Phi(x,y, \alpha) = (x,y) + \alpha \tilde{\mathbf{d}}(x,y)$
is intersection free for $\alpha\leq 1$. 
Each quadrilateral cell is defined by four neighboring vertices. 
To verify validity, we compute the signed area via a triangle decomposition and ensure that $A>0$ everywhere.
Thus, we realize our shape family $\mathcal S$ from Section~\ref{geometric_pretraining} via $\Phi(x,y, \alpha)$ with $\alpha \sim [0.5, 1.0]$. We did not find it necessary to vary the spectral content of the deformation fields, and use a constant $\ell=0.1$. We experimented with various correlation lengths and amplitudes, observing minimal performance variation, and selected a heuristic range bounding point shifts within one cell width to prevent extreme distortions. 

Furthermore, our fine-tuning dataset for the 2D KS equation contains out-of-distribution samples with $\alpha=1.4$. Thus, evaluation on the KS test set (which includes larger deformations) confirms that model performance generalizes robustly without degradation, justifying that more ablation of these parameters are not necessary.

\begin{figure}[h!]
    \centering
    \begin{subfigure}[b]{0.32\textwidth}
        \centering
        \includegraphics[width=\textwidth]{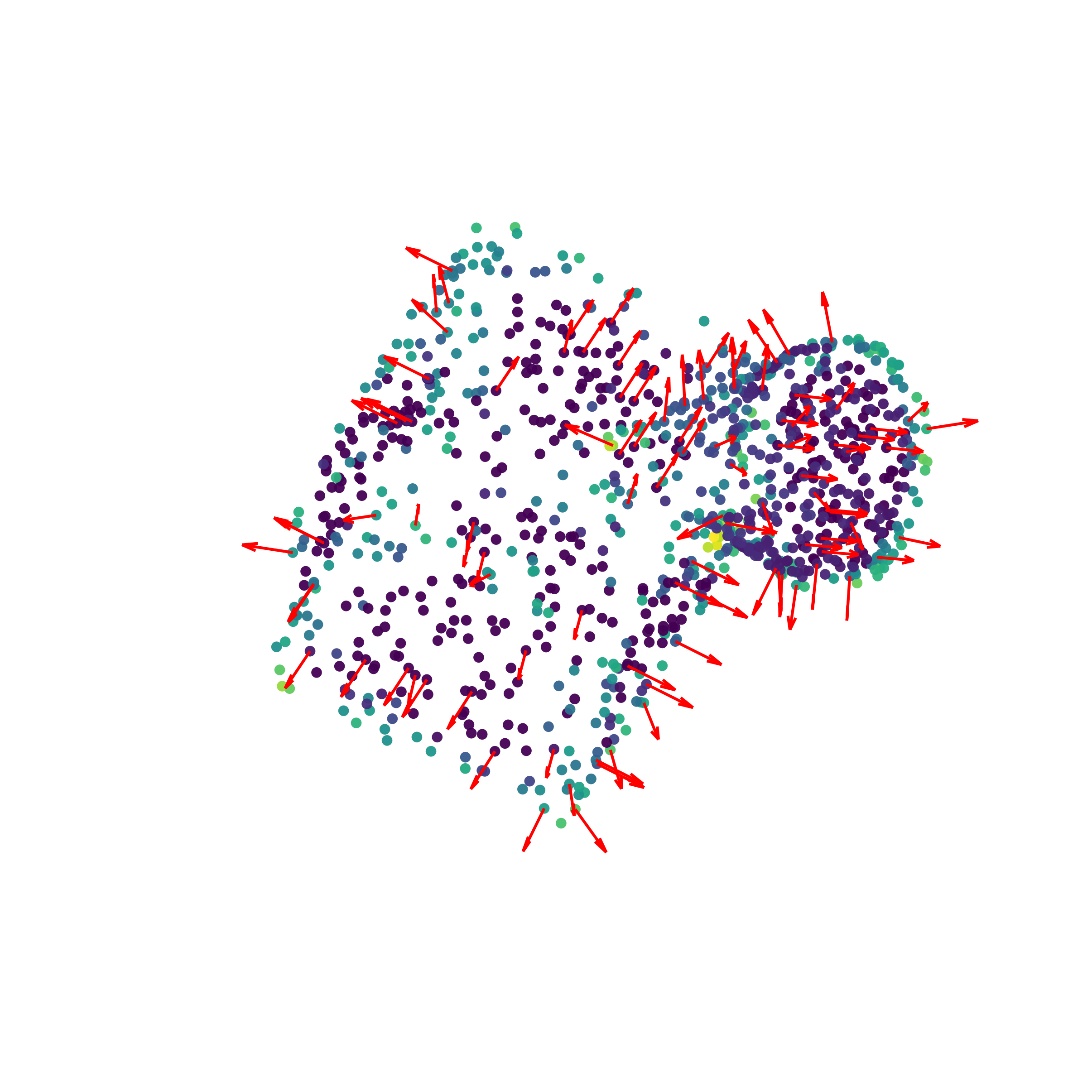}
        \caption{Sample 1}
    \end{subfigure}
    \hfill
    \begin{subfigure}[b]{0.32\textwidth}
        \centering
        \includegraphics[width=\textwidth]{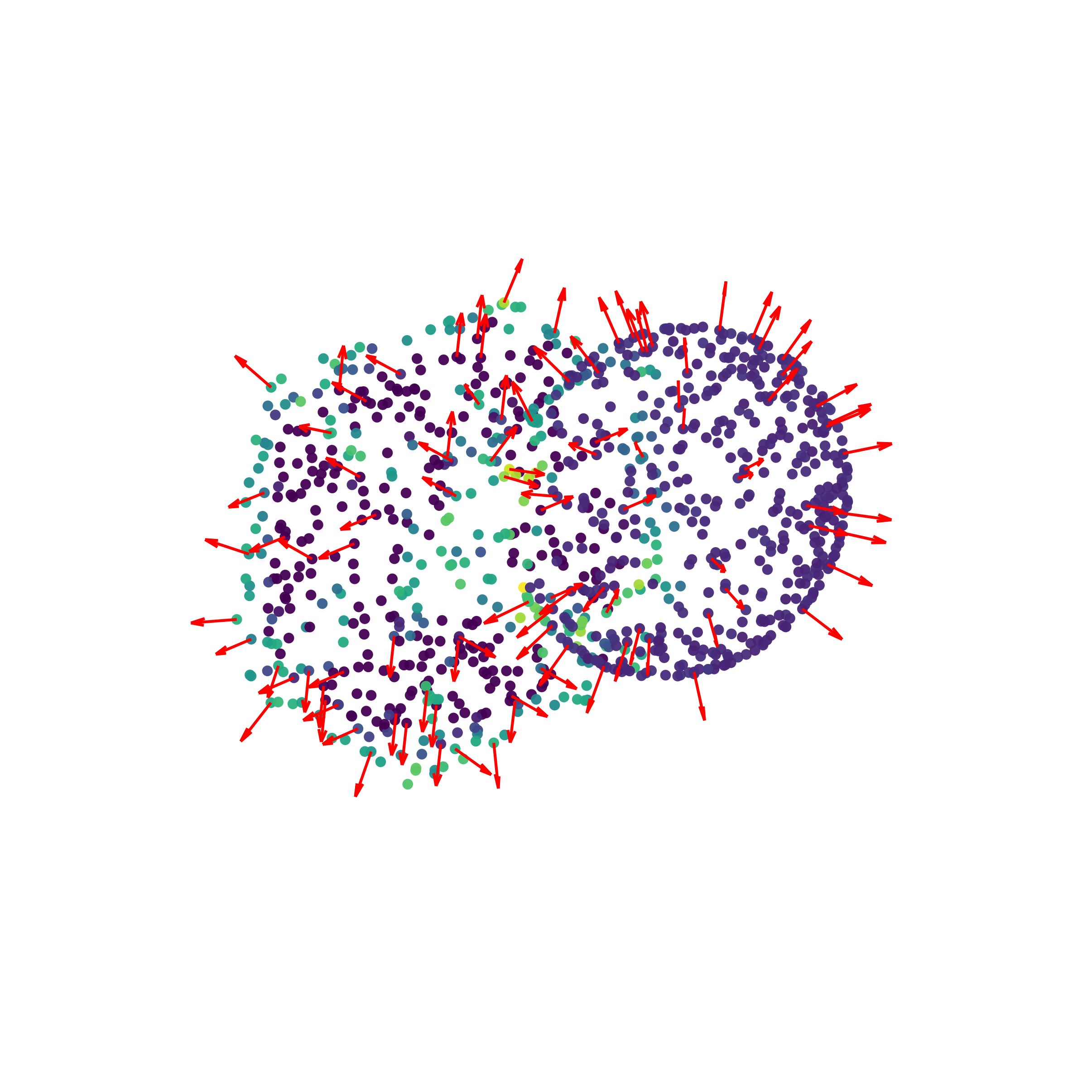}
        \caption{Sample 2}
    \end{subfigure}
    \hfill
    \begin{subfigure}[b]{0.32\textwidth}
        \centering
        \includegraphics[width=\textwidth]{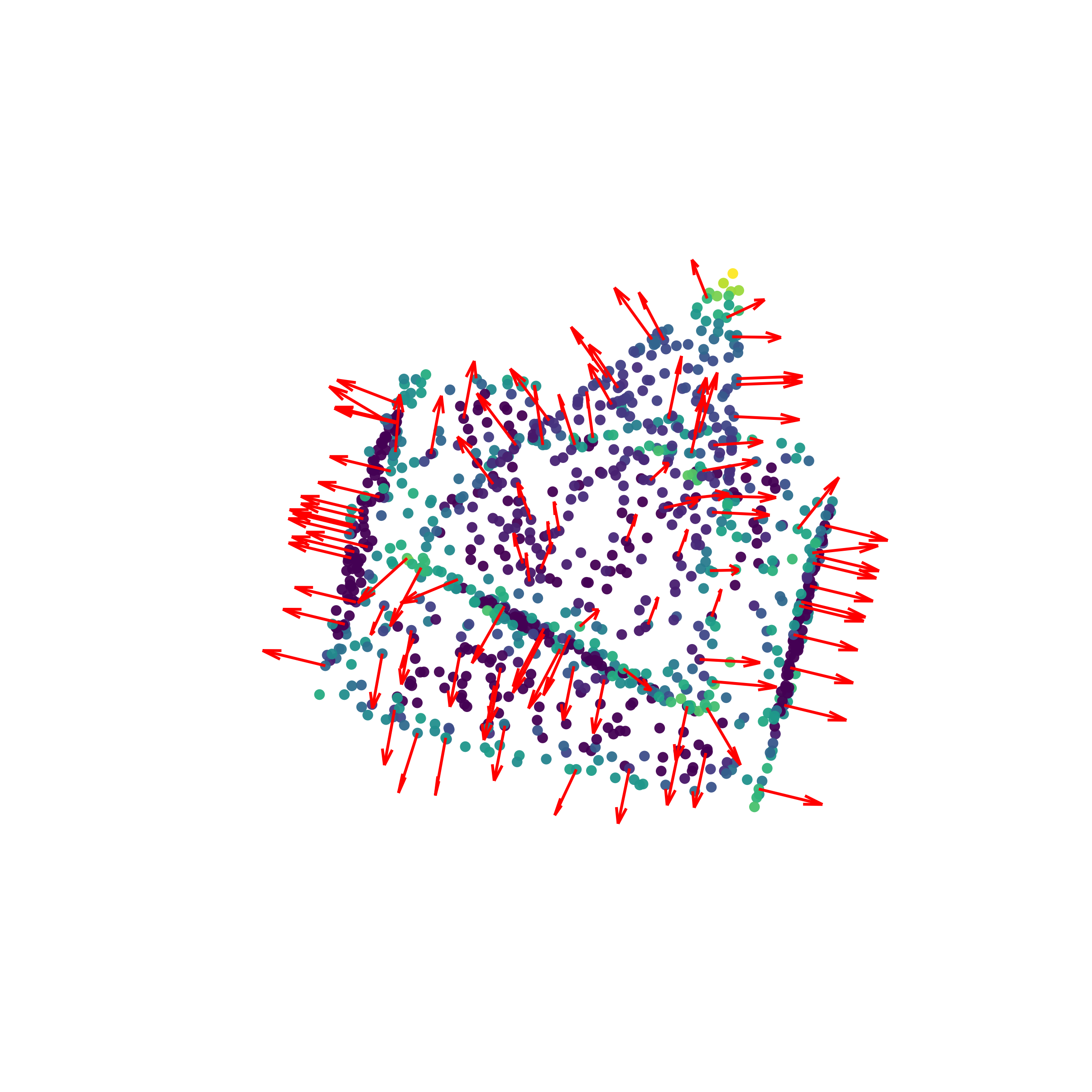}
        \caption{Sample 3}
    \end{subfigure}

    \caption{Example of random pre-training samples built as assemblies of primitive shapes (cubes, spheres, tetrahedra). The plots show the surface curvature and the normals.}
    \label{fig:random_geo_samples}
\end{figure}

Finally, Figure~\ref{fig:random_geo_samples} shows examples of our pre-training shapes, generated as assemblies of geometry primitives such as polyhedra, spheres, and tetrahedra.

\section{Additional results}\label{app:additional_results}

\paragraph{Fine-tuning in complete datasets} Prioritizing performance in low-data regimes is central to the core mission of foundation models in scientific machine learning. In real-world engineering workflows, generating massive 3D CFD/FEM datasets is computationally prohibitive—for instance, generating a single sample in DrivAerML required ~61,000 CPU hours. Practical engineering teams operate predominantly in data-constrained settings. Prior pre-training frameworks exhibit the exact same trend; for instance, GeoPT achieves its highest relative gains under lower sample counts. 
Nonetheless, the transient results in the main text already show that pre-training can enhance accuracy even when working with complete big datasets.
Besides, for steady-state problems, in Table \ref{tab:complete_dataset} we now show that pre-training still provides slight improvements in accuracy when using complete datasets. However, these improvements are marginal for some models. Still, even in data regimes where accuracy gains plateau, pre-training significantly accelerates training convergence speed, offering substantial practical savings in training time and computational resources.

\begin{table}[h!]
    \centering
    \caption{Relative $L_2$ error and pre-training improvement across models for the complete DrivAerML dataset.}
    \label{tab:complete_dataset}
    \small
    \setlength{\tabcolsep}{6pt}
    \begin{tabular}{l c} 
    \toprule
    (Entire dataset) & DrivAerML \\
    \midrule

    Transolver++ (scratch) $\downarrow$    & 0.120 \\
    Transolver++ (PT) $\downarrow$         & 0.104 \\
    Rel. \hspace{0.07cm}Improvement $\uparrow$ & \cellcolor{Green!13}13\% \\

    \midrule

    AB-UPT (scratch) $\downarrow$          & 0.098 \\
    AB-UPT (PT) $\downarrow$               & 0.093 \\
    Rel. \hspace{0.07cm}Improvement $\uparrow$ & \cellcolor{Green!5}5\% \\

    \midrule

    GMP-Net (scratch) $\downarrow$           & 0.075 \\
    GMP-Net (PT) $\downarrow$                & 0.074 \\
    Rel. \hspace{0.07cm}Improvement $\uparrow$ & \cellcolor{Green!1}1\% \\

    \bottomrule
    \end{tabular}
\end{table}

\paragraph{Regional breakdown and theoretical insights} 
Figure~\ref{fig:initial_plot} demonstrates that point-wise prediction errors are substantially reduced across aerodynamically critical regions, such as the side mirrors and front bumper of automotive geometries. A detailed regional breakdown quantifying these performance improvements across distinct geometric components is provided in Table~\ref{tab:aerodynamic_breakdown}.

\begin{table}[h!]
  \caption{Aerodynamic regional breakdown for automotive and aerospace benchmarks. 64 training samples with GMP-Net.}
  \label{tab:aerodynamic_breakdown}
  \small
  \centering
  \begin{tabular}{llr}
    \toprule
    \textbf{Benchmark} & \textbf{Region} & \textbf{Improvement (\%)} \\
    \midrule
    DrivAerML
      & Front      & 23.5 \\
      & Mid car     & 10.0 \\
      & Underbody         & 17.4 \\
      & Tail / Diffuser      & 19.8 \\
    \midrule
    SHIFT-Wing
      & Leading edge              & 21.6 \\
      & Upper surface             & 15.3 \\
      & Lower surface             & 22.1 \\
    \bottomrule
  \end{tabular}
\end{table}

\section{Extra plots}\label{app:extra_plots}

In the final section of the Appendix we provide additional plots and figures with qualitative results.

\begin{figure}[h]
    \centering
    \includegraphics[width=\textwidth]{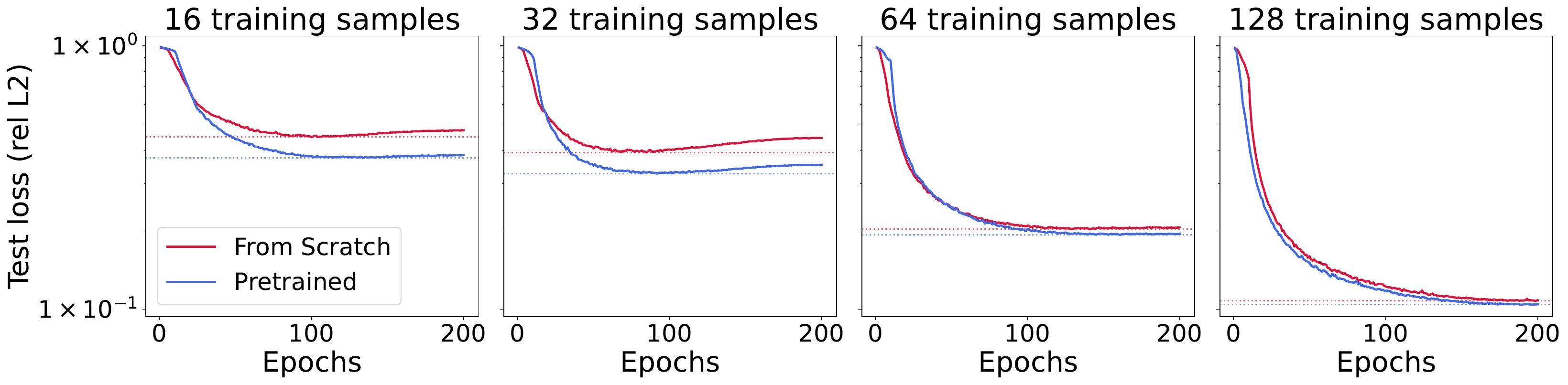}
    \caption{Test loss of GMP-Net trained from scratch and with pre-trained weights. Horizontal lines indicate the minimum achieved error. Pre-trained models achieve lower error and faster convergence on DrivAerML.}
    \label{fig:app_drivaerml_smart}
\end{figure}

\begin{figure}[h]
    \centering
    \includegraphics[width=\textwidth]{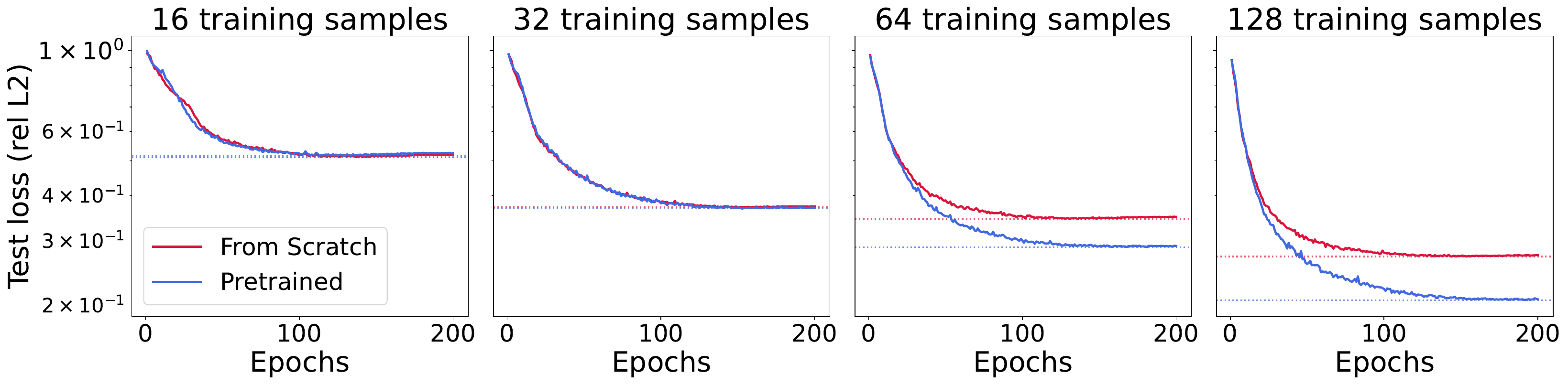}
    \caption{Test loss of AB-UPT trained from scratch and with pre-trained weights. Horizontal lines indicate the minimum achieved error. Pre-trained models achieve lower error and faster convergence on DrivAerML.}
    \label{fig:app_drivaerml_abupt}
\end{figure}

\begin{figure}[h]
    \centering
    \includegraphics[width=\textwidth]{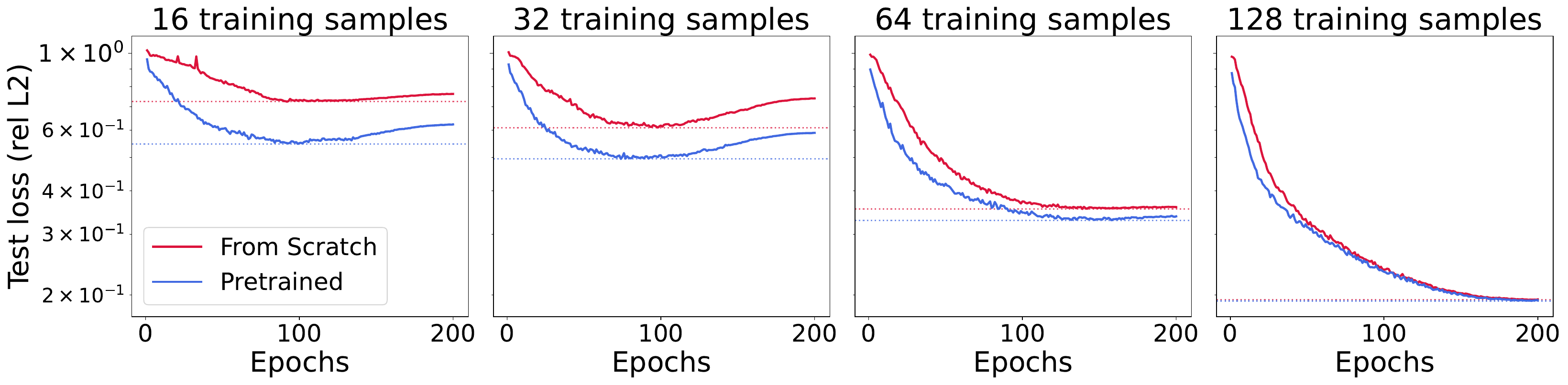}
    \caption{Test loss of Transolver++ trained from scratch and with pre-trained weights. Horizontal lines indicate the minimum achieved error. Pre-trained models achieve lower error and faster convergence on DrivAerML.}
    \label{fig:app_drivaerml_transolver}
\end{figure}

\begin{figure}[h]
    \centering
    \includegraphics[width=\textwidth]{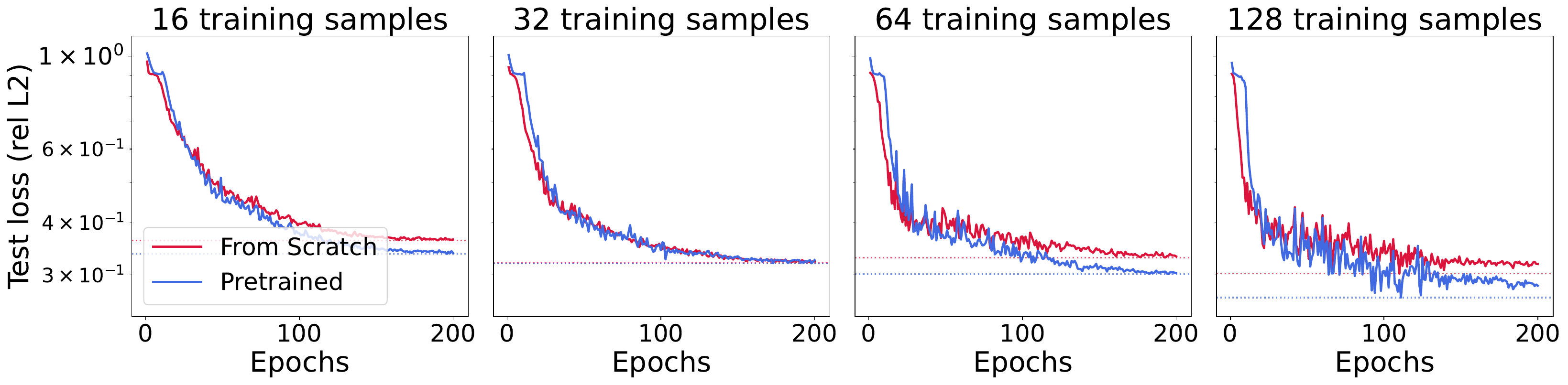}
    \caption{Test loss of GMP-Net trained from scratch and with pre-trained weights. Horizontal lines indicate the minimum achieved error. Pre-trained models achieve lower error and faster convergence on SHIFT-Wing.}
    \label{fig:app_shiftwing_smart}
\end{figure}

\begin{figure}[h]
    \centering
    \includegraphics[width=\textwidth]{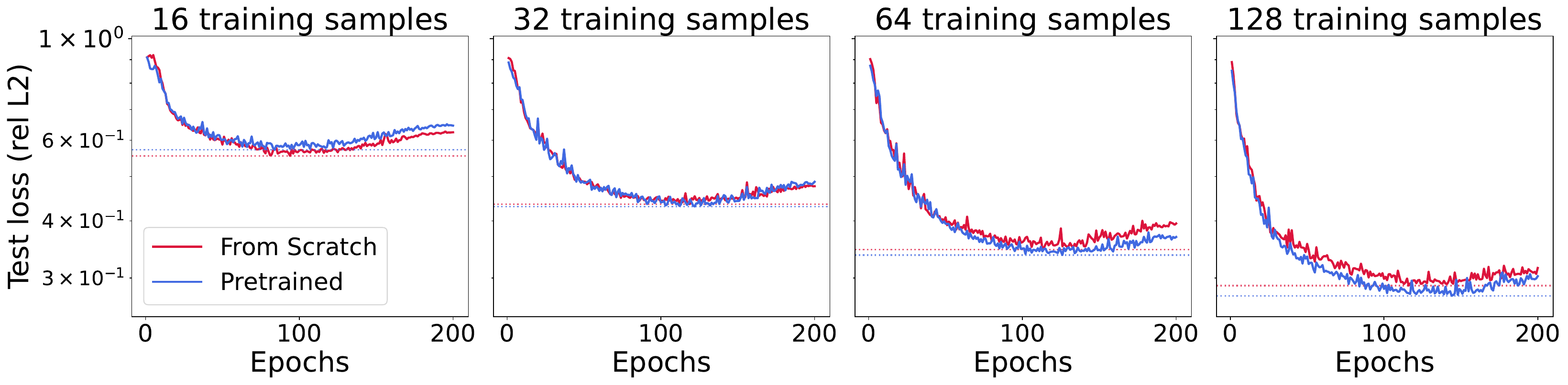}
    \caption{Test loss of AB-UPT trained from scratch and with pre-trained weights. Horizontal lines indicate the minimum achieved error. Pre-trained models achieve lower error and faster convergence on SHIFT-Wing.}
    \label{fig:app_shiftwing_abupt}
\end{figure}

\begin{figure}[h]
    \centering
    \includegraphics[width=\textwidth]{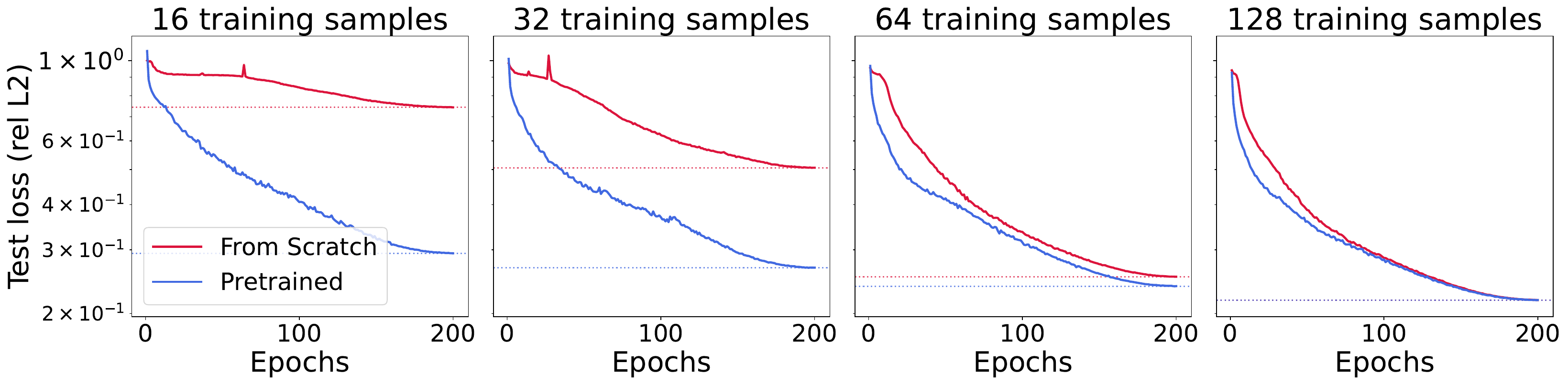}
    \caption{Test loss of Transolver++ trained from scratch and with pre-trained weights. Horizontal lines indicate the minimum achieved error. Pre-trained models achieve lower error and faster convergence on SHIFT-Wing.}
    \label{fig:app_shiftwing_transolver}
\end{figure}

\begin{figure}[h]
    \centering
    \includegraphics[width=\textwidth]{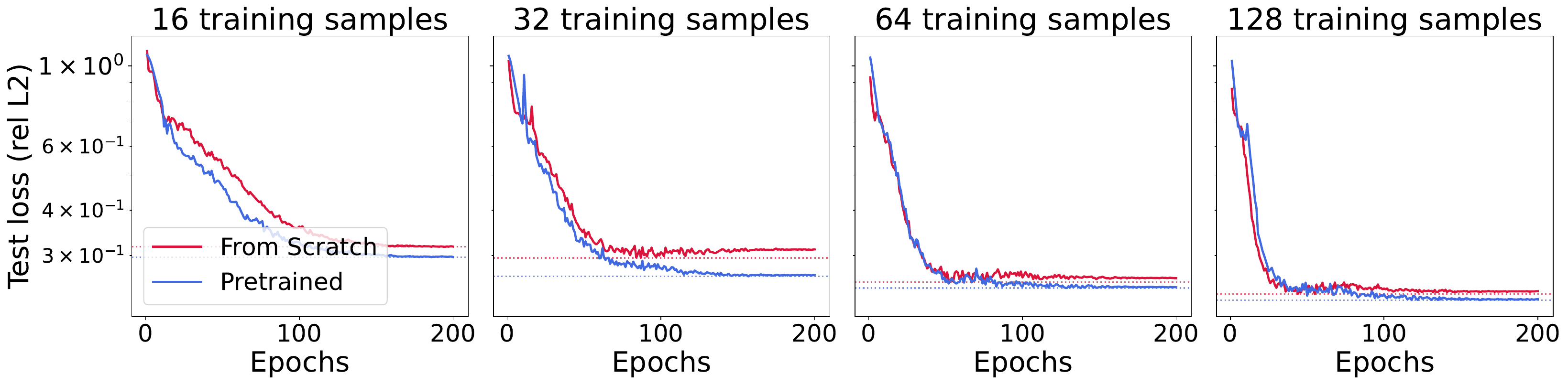}
    \caption{Test loss of GMP-Net trained from scratch and with pre-trained weights. Horizontal lines indicate the minimum achieved error. Pre-trained models achieve lower error and faster convergence on SHIFT-Crash.}
    \label{fig:app_shiftcrash_smart}
\end{figure}

\begin{figure}[h]
    \centering
    \includegraphics[width=\textwidth]{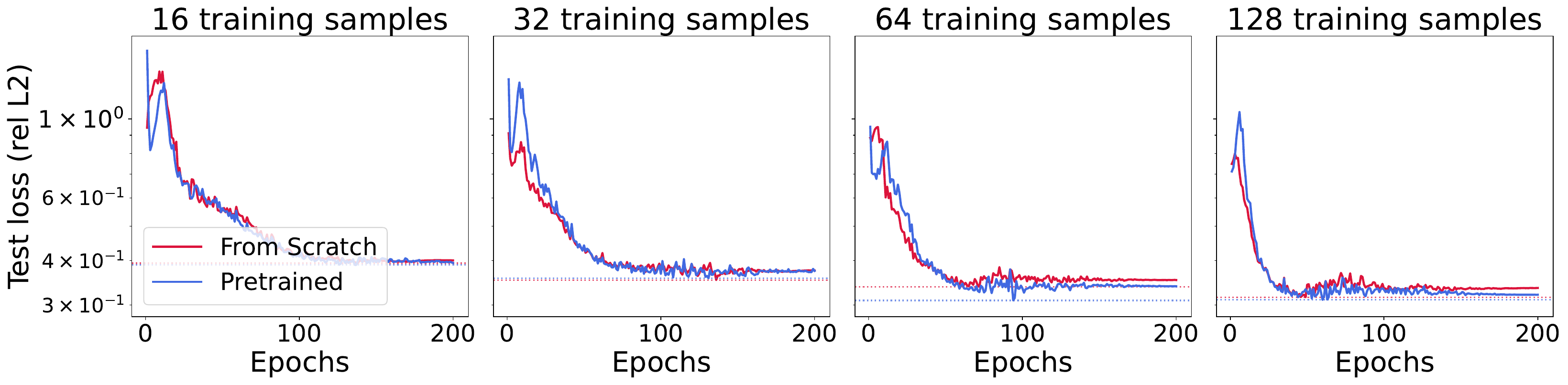}
    \caption{Test loss of AB-UPT trained from scratch and with pre-trained weights. Horizontal lines indicate the minimum achieved error. Pre-trained models achieve lower error and faster convergence on SHIFT-Crash.}
    \label{fig:app_shiftcrash_abupt}
\end{figure}

\begin{figure}[h]
    \centering
    \includegraphics[width=\textwidth]{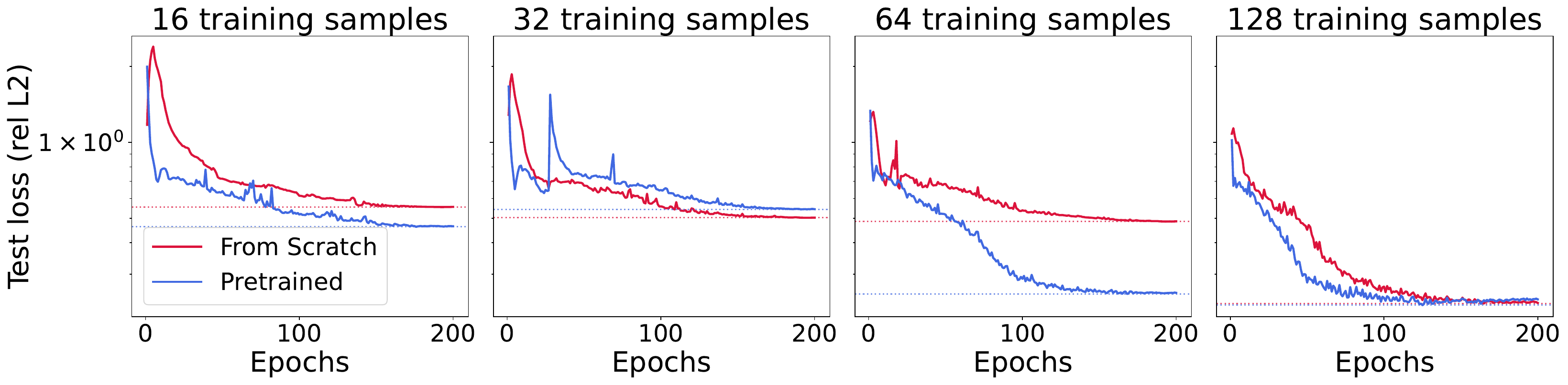}
    \caption{Test loss of Transolver++ trained from scratch and with pre-trained weights. Horizontal lines indicate the minimum achieved error. Pre-trained models achieve lower error and faster convergence on SHIFT-Crash.}
    \label{fig:app_shiftcrash_transolver}
\end{figure}

\begin{figure}[h]
    \centering
    \begin{subfigure}[c]{0.32\textwidth}
        \includegraphics[width=\textwidth]{figures/drivaerml_convergence_improvement.pdf}
        \caption{DrivAerML.}
    \end{subfigure}
    \hfill
    \begin{subfigure}[c]{0.32\textwidth}
        \centering
        \includegraphics[width=\textwidth]{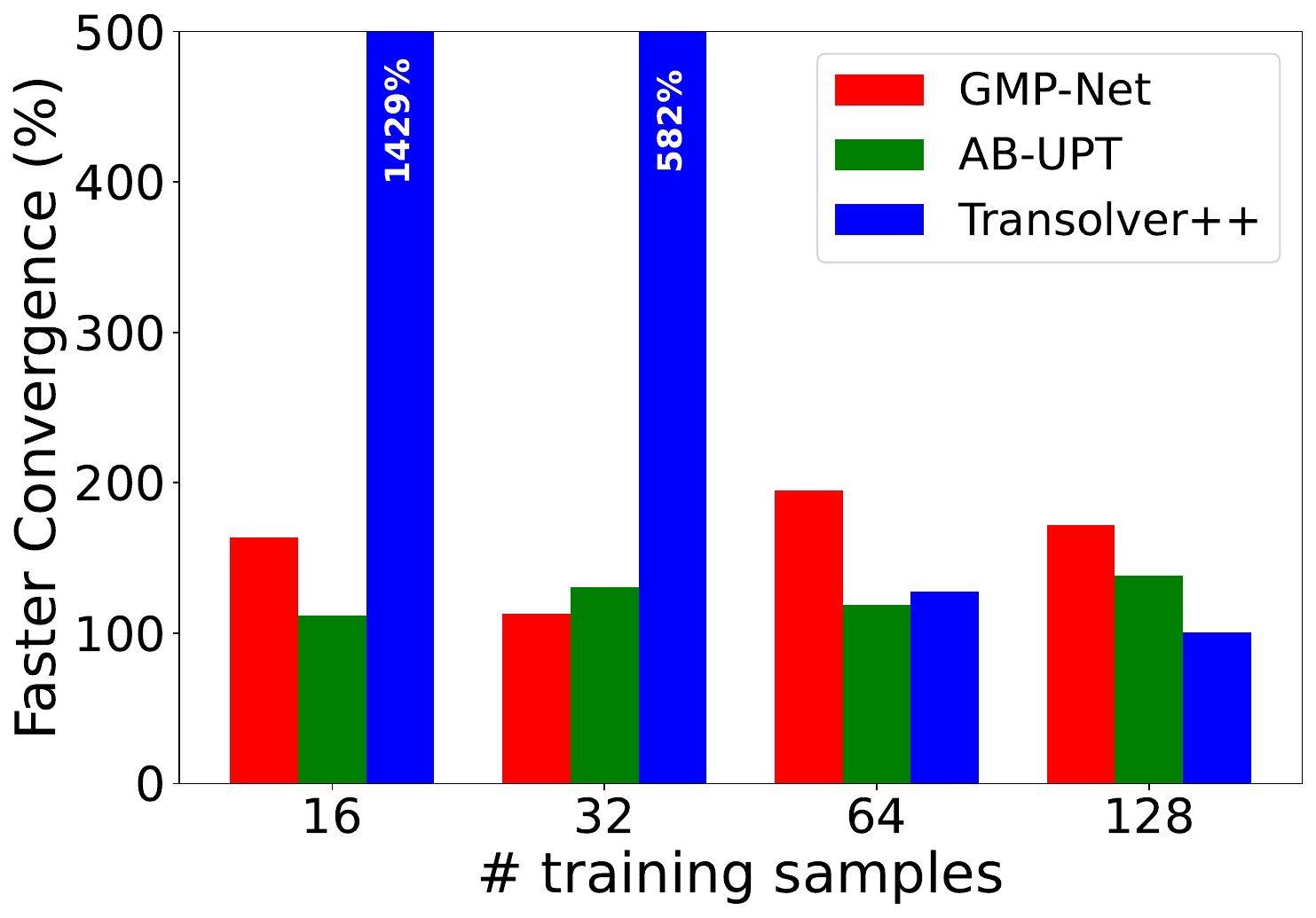}
        \caption{SHIFT-Wing.}
    \end{subfigure}
    \hfill
    \begin{subfigure}[c]{0.32\textwidth}
        \centering
        \includegraphics[width=\textwidth]{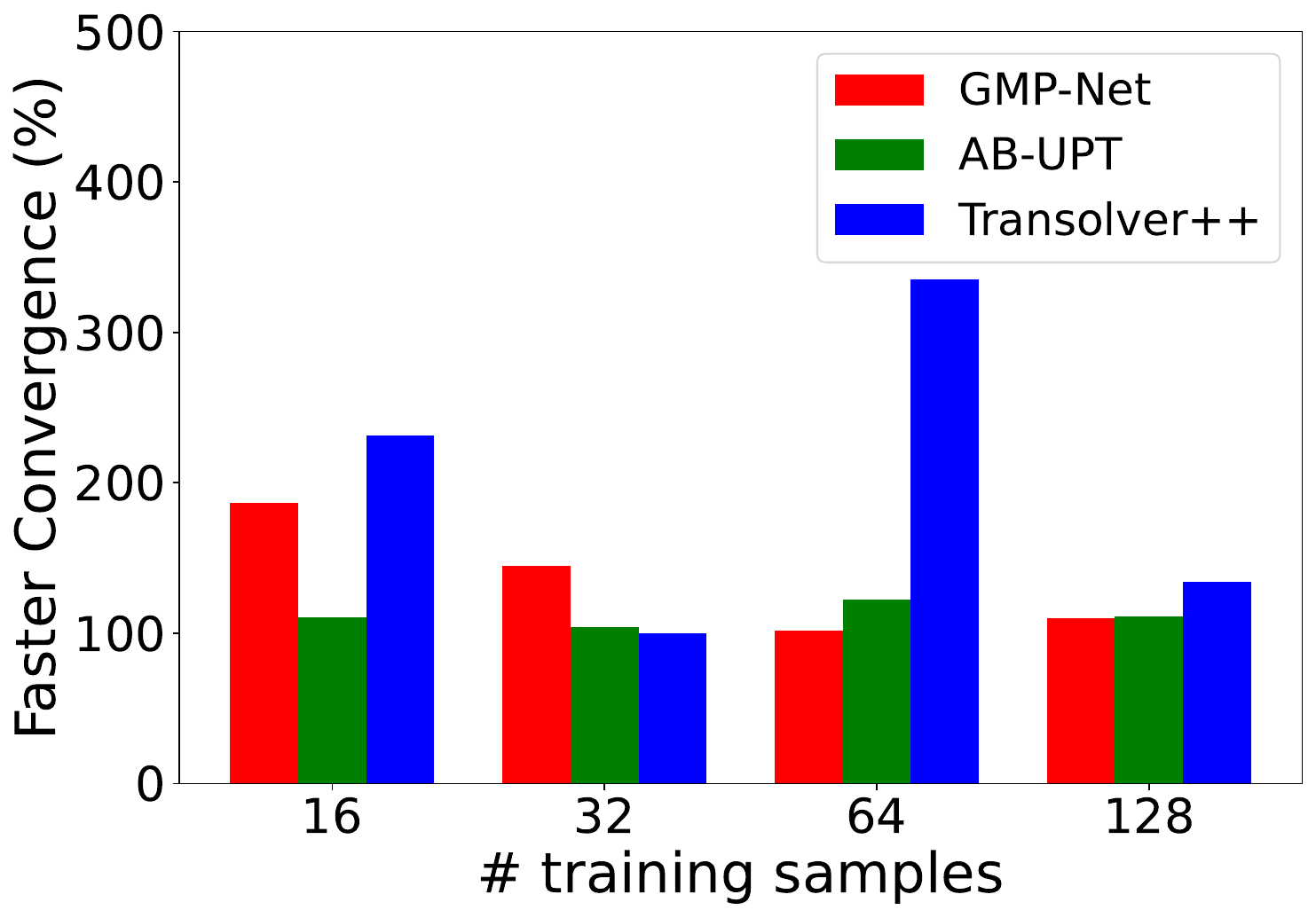}
        \caption{SHIFT-Crash.}
    \end{subfigure}
    \hfill
    \caption{Faster convergence improvement of GMP pre-training.}
    \label{fig:app_convergence}
\end{figure}

\begin{figure}[h]
    \centering
    \begin{subfigure}[c]{0.32\textwidth}
        \includegraphics[width=\textwidth]{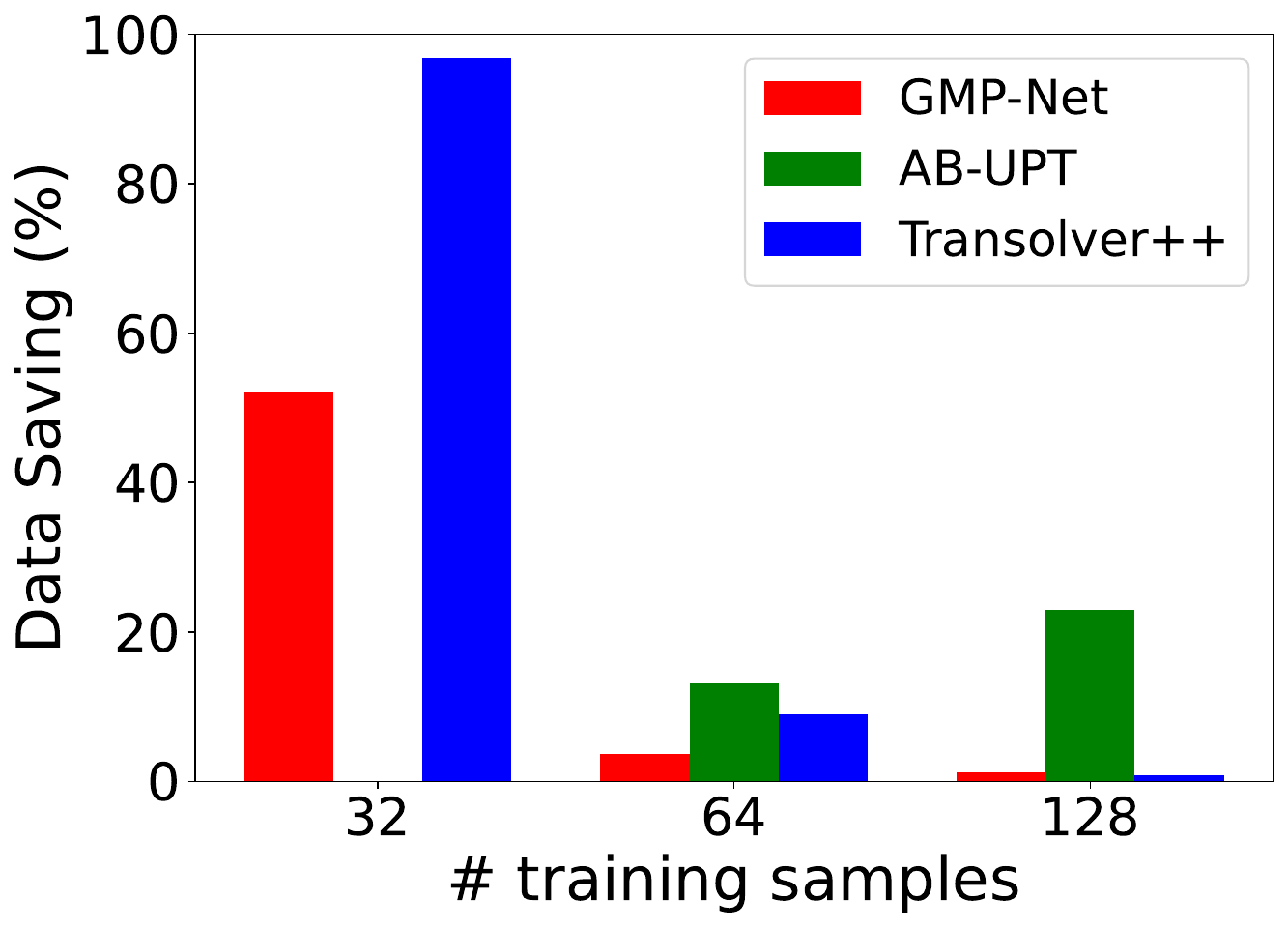}
        \caption{DrivAerML.}
    \end{subfigure}
    \hfill
    \begin{subfigure}[c]{0.32\textwidth}
        \centering
        \includegraphics[width=\textwidth]{figures/shiftwing_data_saving.pdf}
        \caption{SHIFT-Wing.}
    \end{subfigure}
    \hfill
    \begin{subfigure}[c]{0.32\textwidth}
        \centering
        \includegraphics[width=\textwidth]{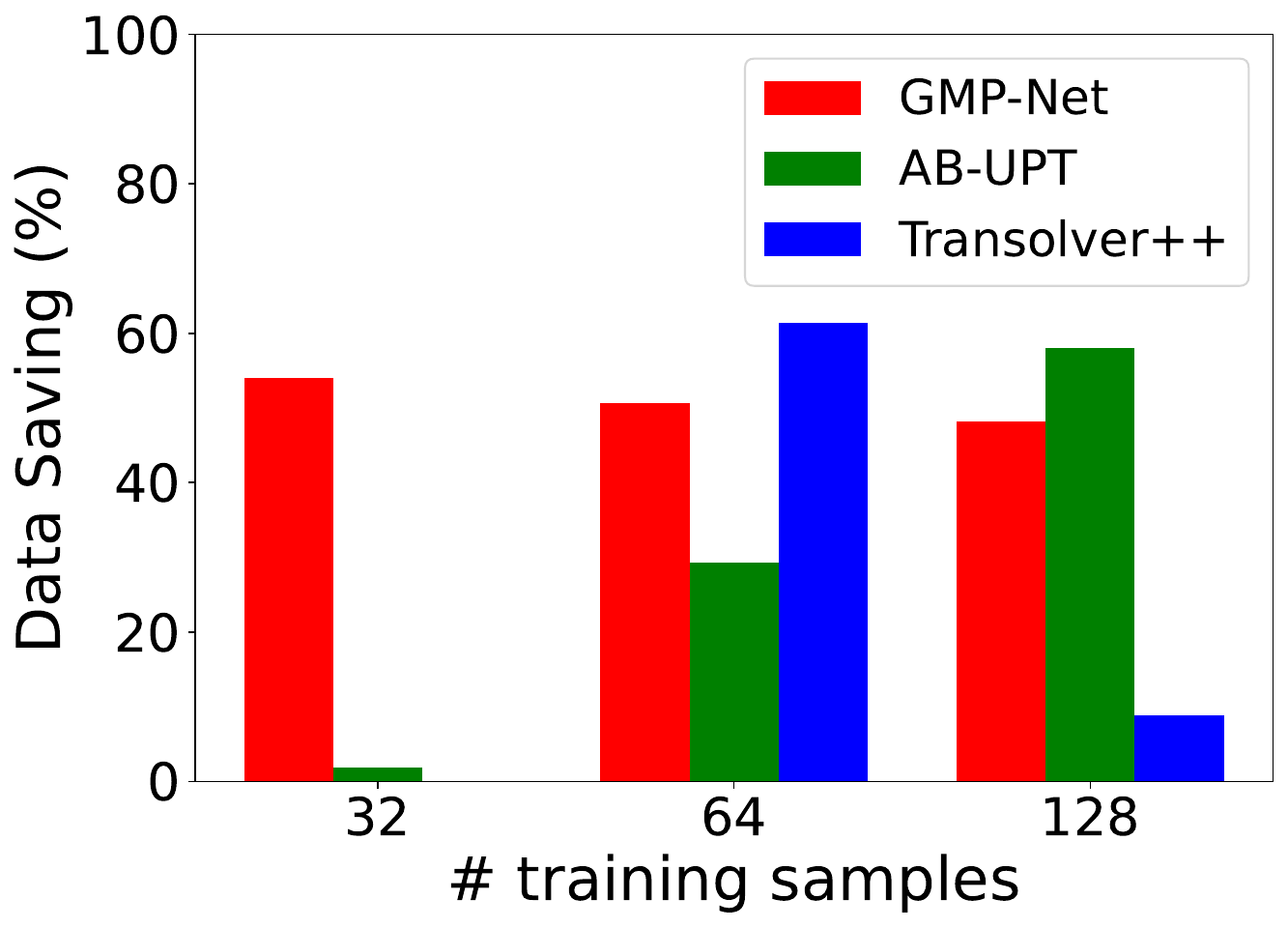}
        \caption{SHIFT-Crash.}
    \end{subfigure}
    \hfill
    \caption{Data saving due to GMP pre-training.}
    \label{fig:app_datasaving}
\end{figure}

\begin{figure}[h!]
    \centering
    \includegraphics[width=\textwidth]{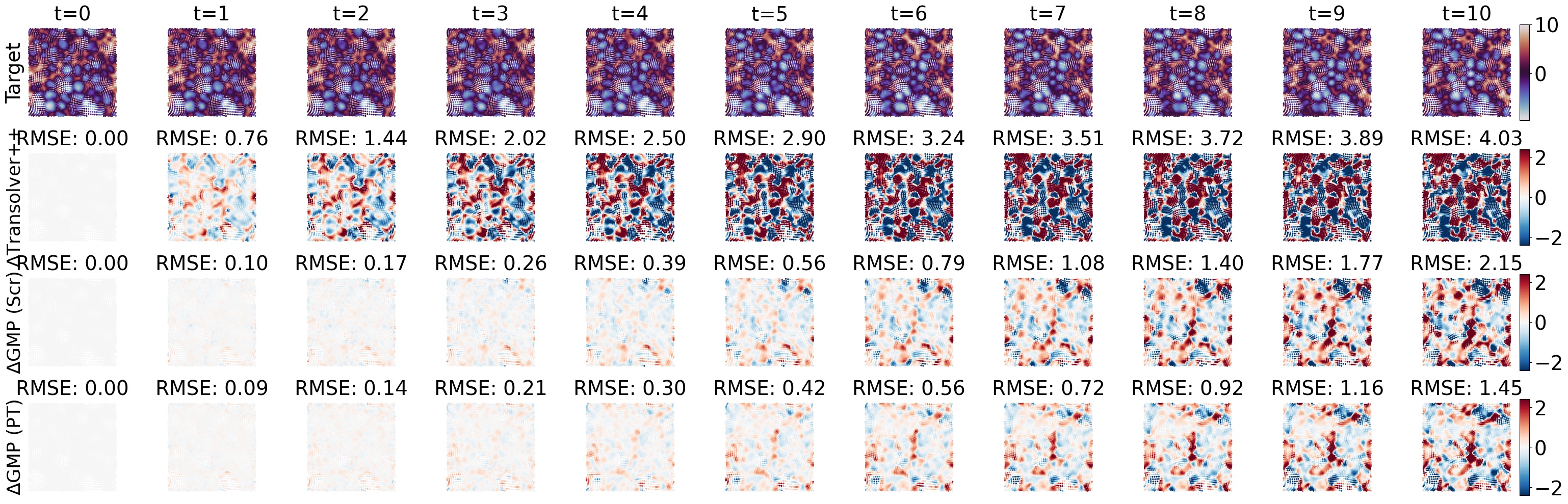} 
    \caption{2D KS equation rollout error comparison (sample 1). GMP successfully improves the accuracy over the from-scratch variant for all timesteps. GMP-Net also achieves lower errors than Transolver++.}
    \label{fig:rollout_ks_error_app}
\end{figure}

\begin{figure}[h!]
    \centering
    \includegraphics[width=\textwidth]{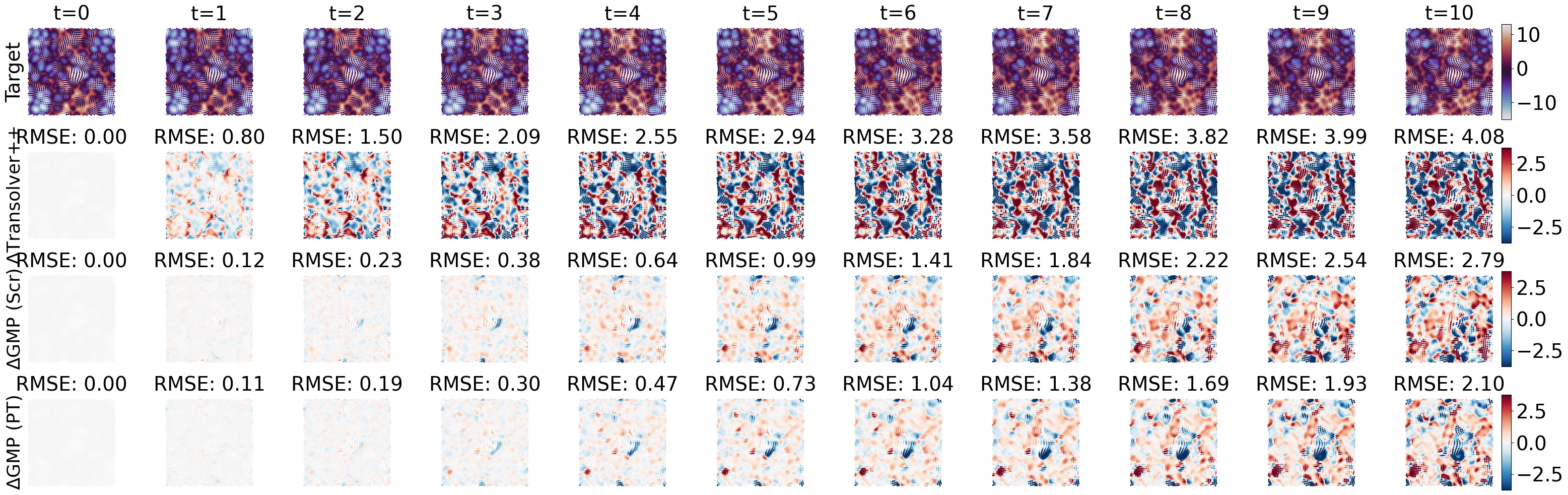} 
    \caption{2D KS equation rollout error comparison (sample 2). GMP successfully improves the accuracy over the from-scratch variant for all timesteps. GMP-Net also achieves lower errors than Transolver++.}
    \label{fig:rollout_ks_error_app_2}
\end{figure}

\begin{figure}[h!]
    \centering
    \includegraphics[width=\textwidth]{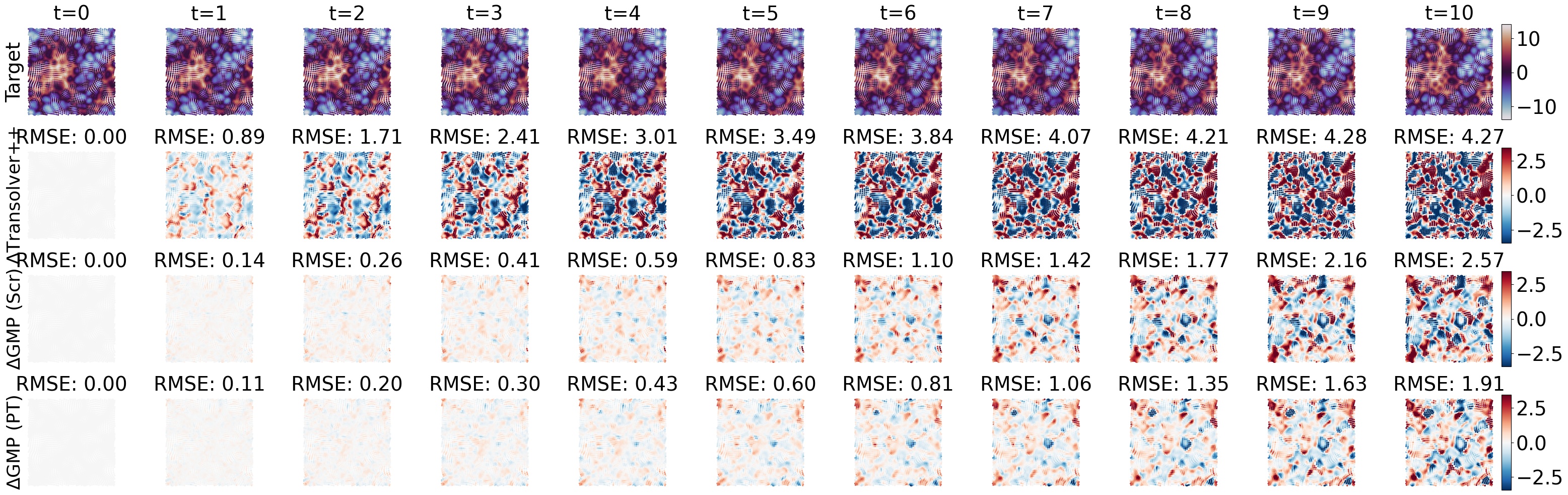} 
    \caption{2D KS equation rollout error comparison (sample 3). GMP successfully improves the accuracy over the from-scratch variant for all timesteps. GMP-Net also achieves lower errors than Transolver++.}
    \label{fig:rollout_ks_error_app_3}
\end{figure}

\begin{figure}[h!]
    \centering
    \includegraphics[width=\textwidth]{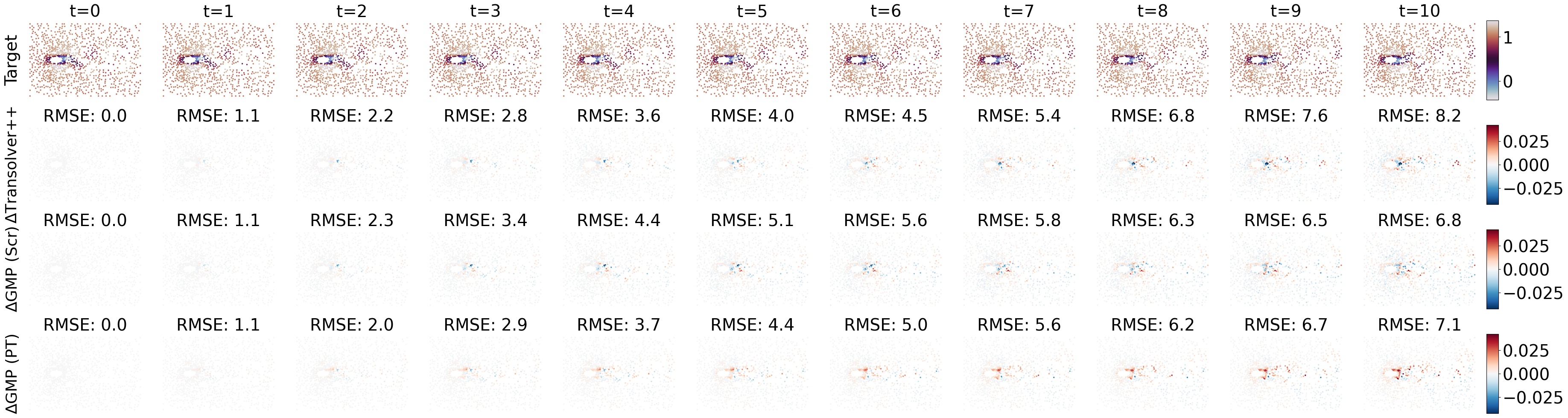} 
    \caption{2D flow around an ellipse rollout error comparison (sample 1). GMP successfully improves the accuracy over the from-scratch variant for all timesteps. Compared to Transolver++, GMP-Net achieves lower errors in longer rollouts, but higher in the intermediate timesteps.}
    \label{fig:rollout_ellipse_error}
\end{figure}

\begin{figure}[h!]
    \centering
    \includegraphics[width=\textwidth]{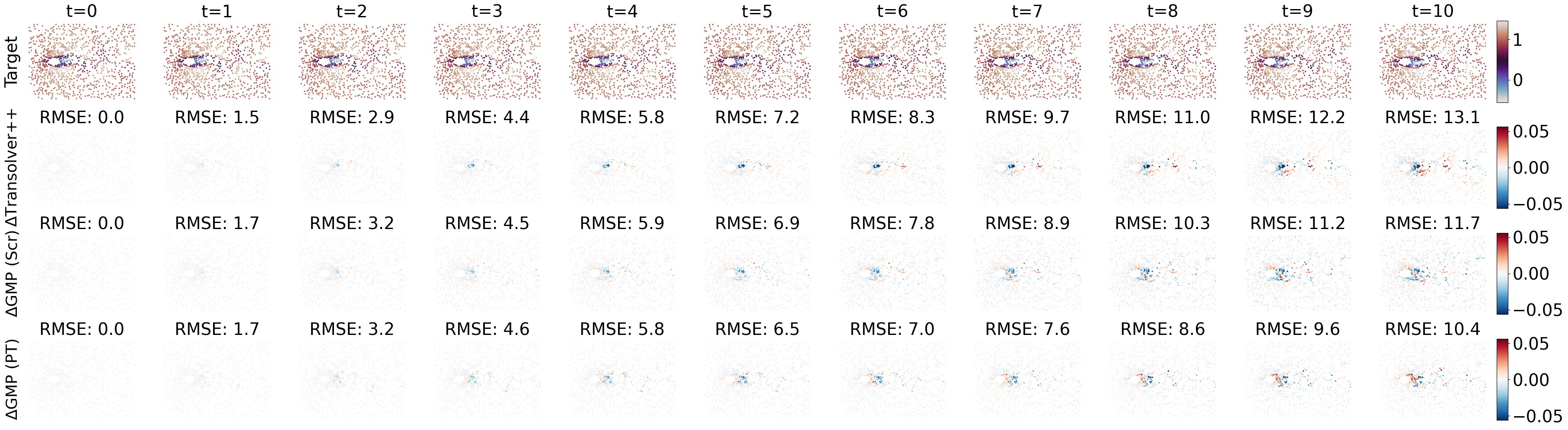} 
    \caption{2D flow around an ellipse rollout error comparison (sample 2). GMP successfully improves the accuracy over the from-scratch variant for all timesteps. Compared to Transolver++, GMP-Net achieves lower errors in longer rollouts, but higher in the intermediate timesteps.}
    \label{fig:rollout_ellipse_error_2}
\end{figure}

\begin{figure}[h!]
    \centering
    \includegraphics[width=\textwidth]{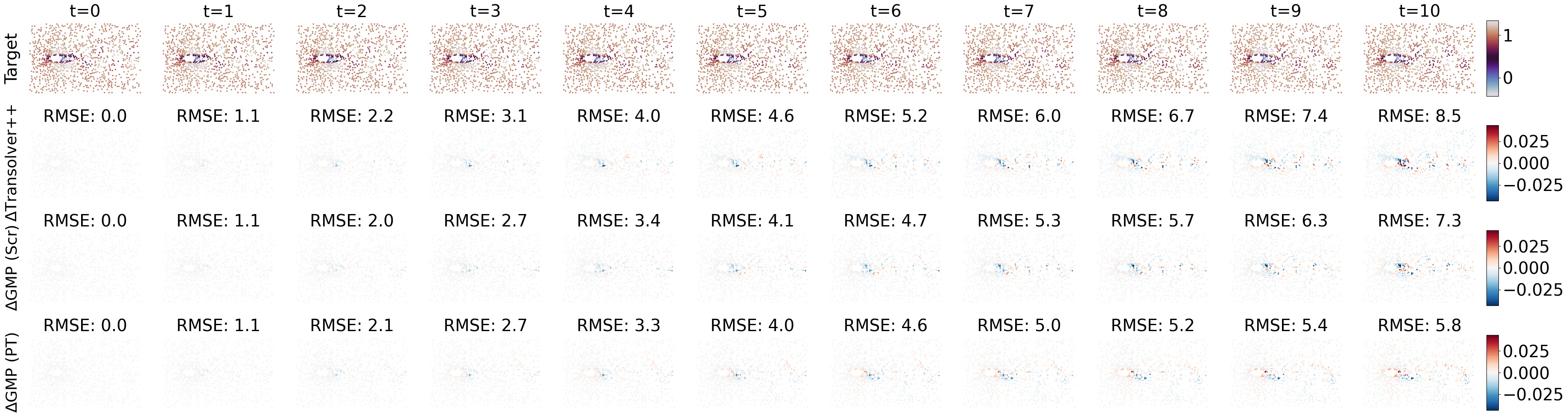} 
    \caption{2D flow around an ellipse rollout error comparison (sample 3). GMP successfully improves the accuracy over the from-scratch variant for all timesteps. Compared to Transolver++, GMP-Net achieves slightly lower errors overall.}
    \label{fig:rollout_ellipse_error_3}
\end{figure}

\begin{figure}[h!]
    \centering
    \includegraphics[width=\textwidth]{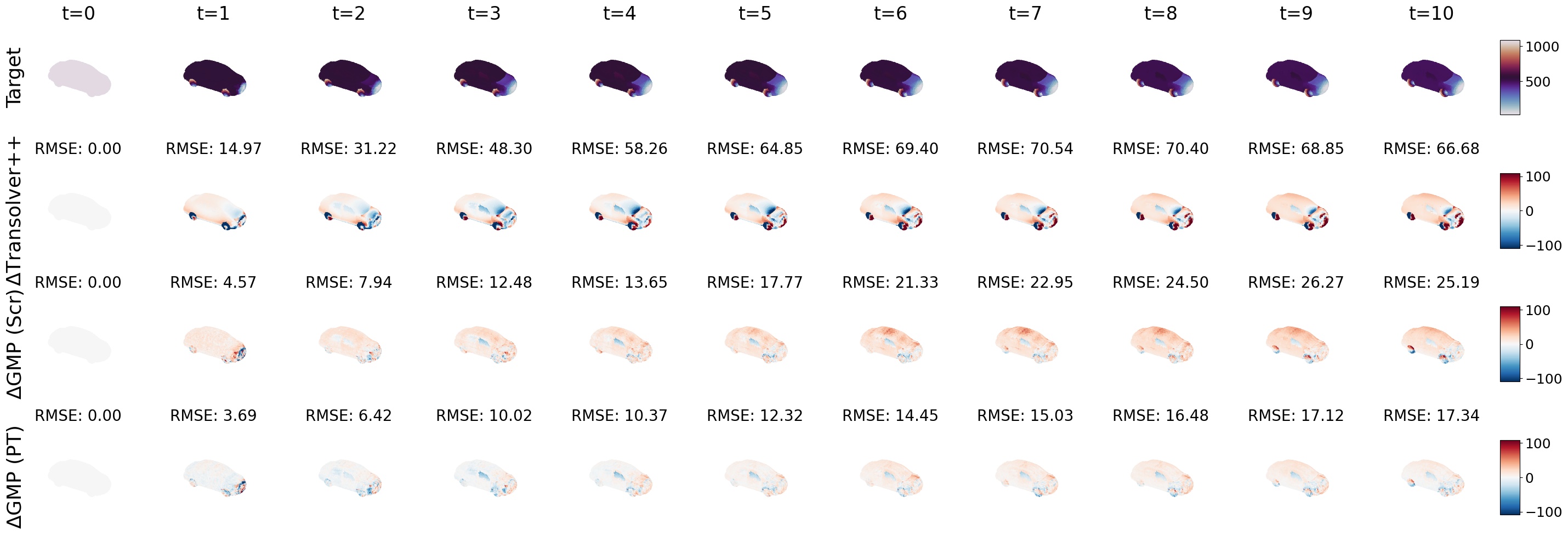} 
    \caption{Crash deformation rollout error comparison (sample 1). GMP successfully improves the accuracy over the from-scratch variant. Compared to Transolver++, GMP-Net achieves lower errors.}
    \label{fig:rollout_crash_1}
\end{figure}

\begin{figure}[h!]
    \centering
    \includegraphics[width=\textwidth]{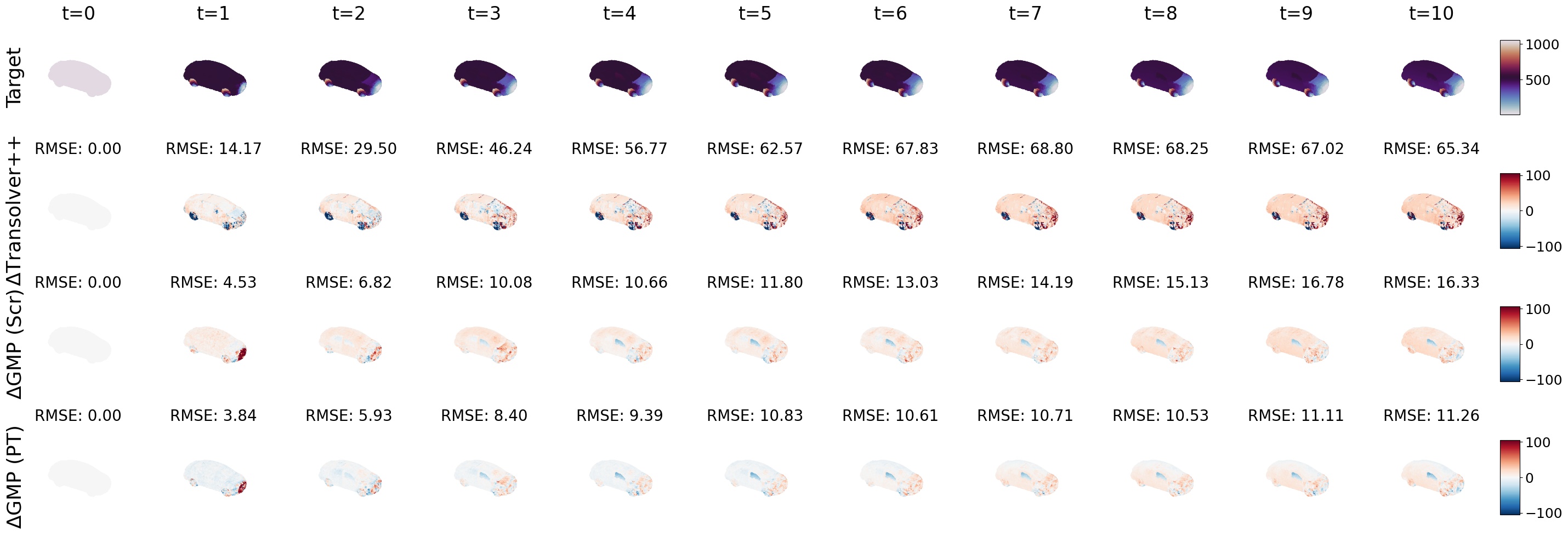} 
    \caption{Crash deformation rollout error comparison (sample 2). GMP successfully improves the accuracy over the from-scratch variant. Compared to Transolver++, GMP-Net achieves lower errors.}
    \label{fig:rollout_crash_2}
\end{figure}

\begin{figure}[h!]
    \centering
    \includegraphics[width=0.8\textwidth]{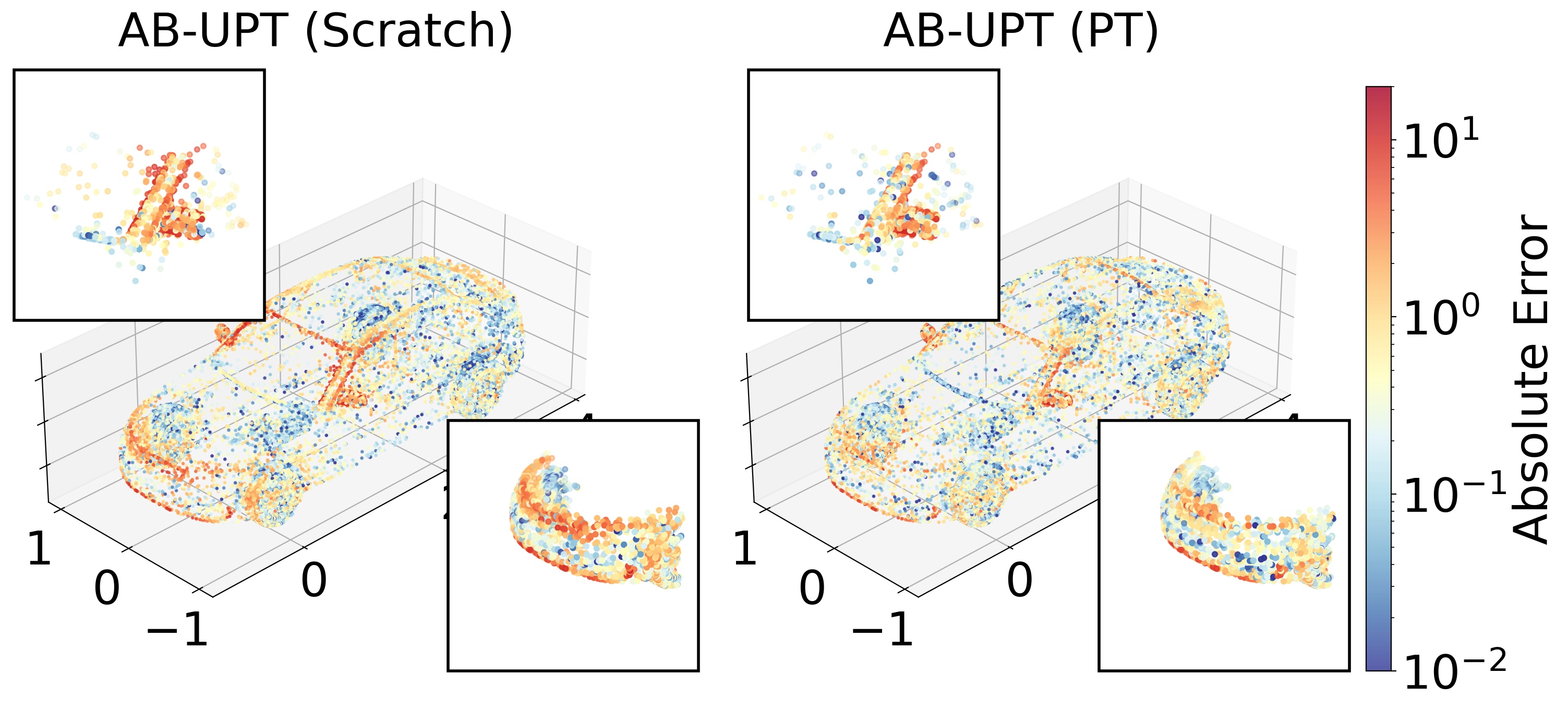} 
    \caption{DrivAerML surface wall shear stress absolute error comparison. 32 training samples for AB-UPT.}
    \label{fig:app_car_1}
\end{figure}

\begin{figure}[h!]
    \centering
    \includegraphics[width=0.8\textwidth]{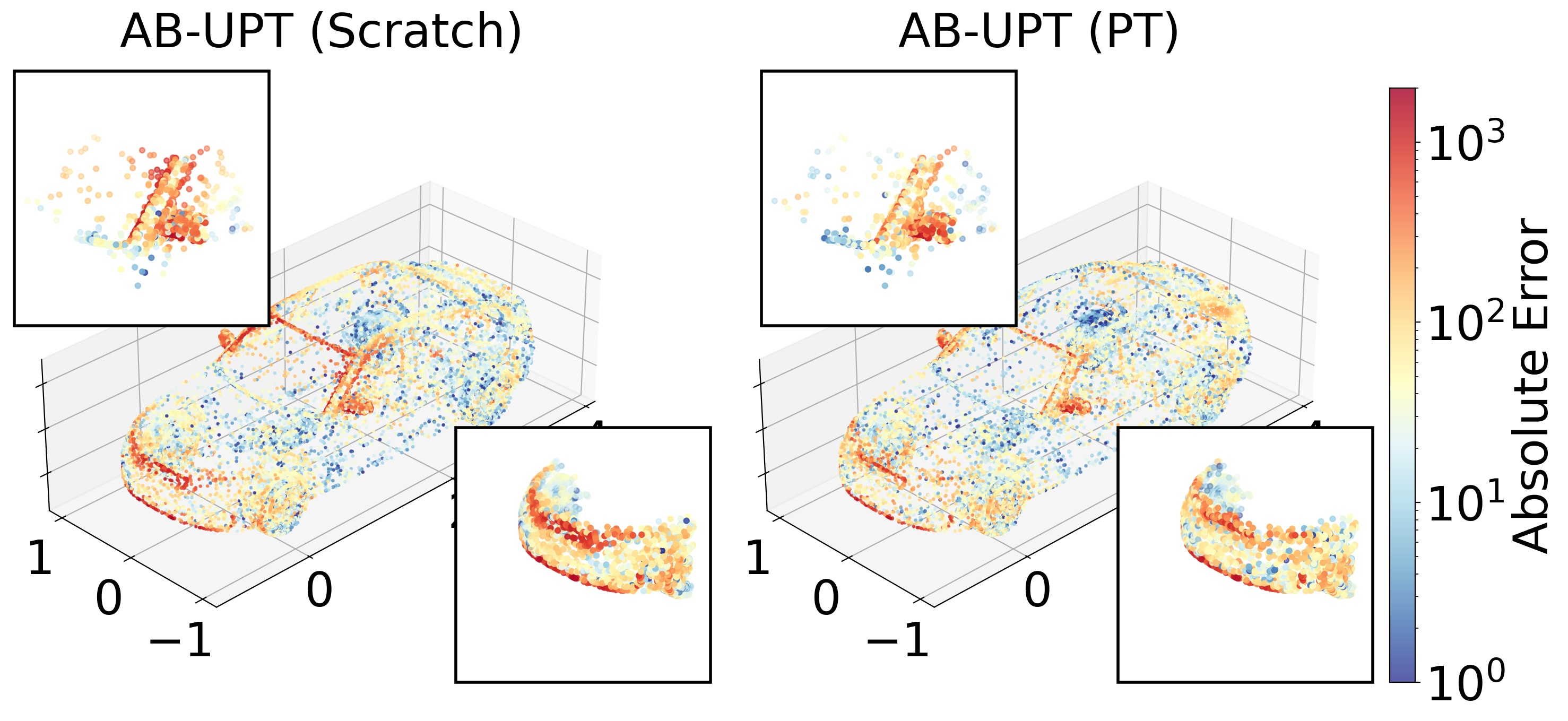} 
    \caption{DrivAerML surface pressure absolute error comparison. 32 training samples for AB-UPT.}
    \label{fig:app_car_2}
\end{figure}

\begin{figure}[h!]
    \centering
    \includegraphics[width=0.8\textwidth]{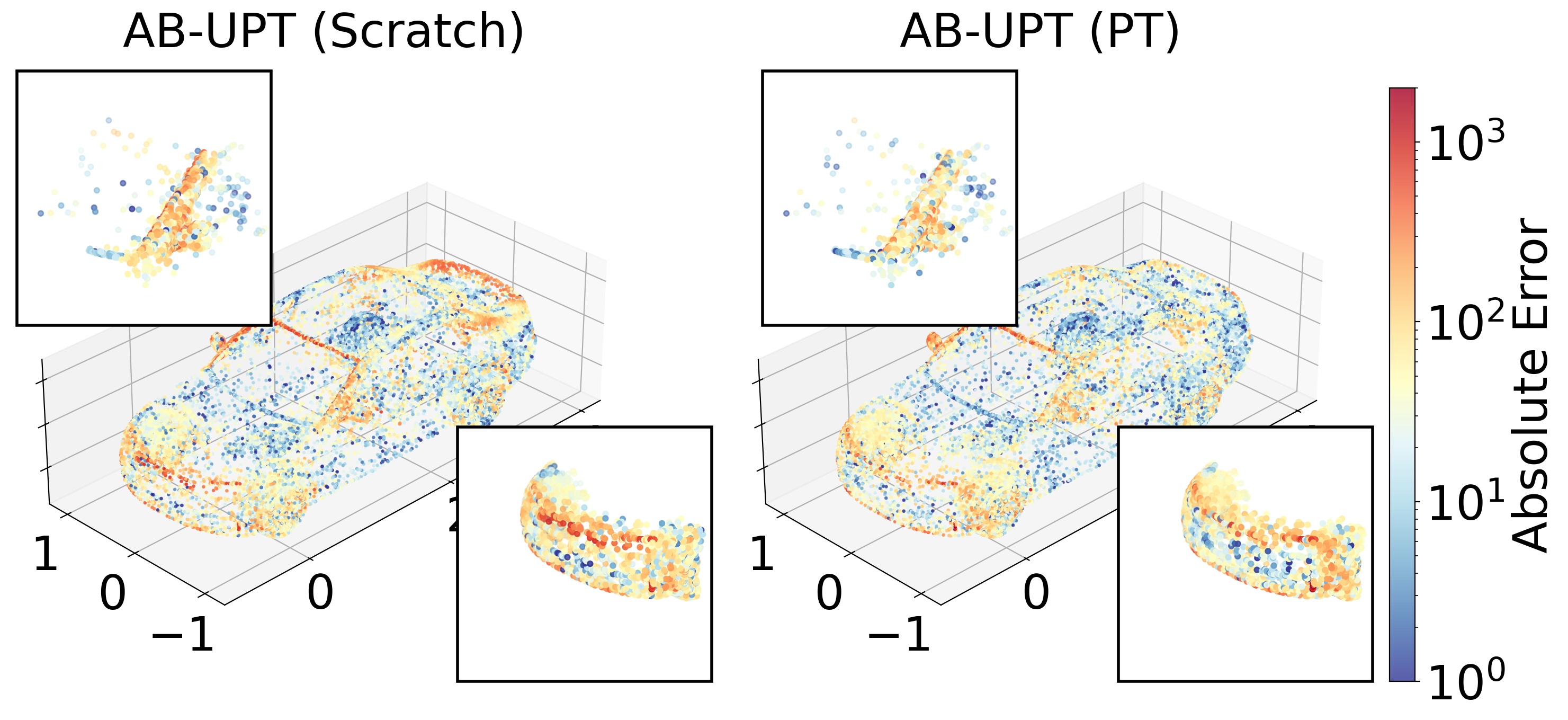} 
    \caption{DrivAerML surface pressure absolute error comparison. 32 training samples for AB-UPT.}
    \label{fig:app_car_3}
\end{figure}

\begin{figure}[h!]
    \centering
    \includegraphics[width=0.8\textwidth]{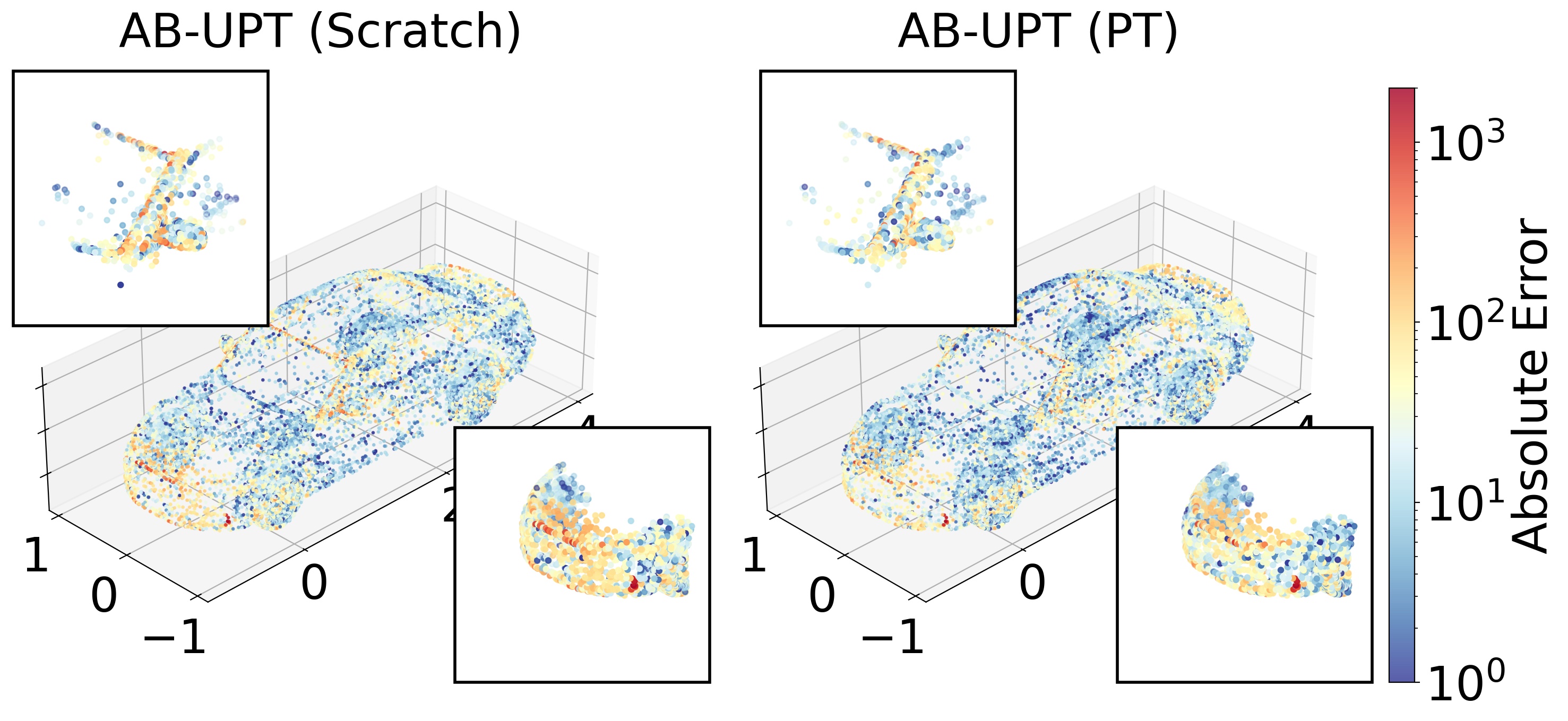} 
    \caption{DrivAerML surface pressure absolute error comparison. 64 training samples for AB-UPT.}
    \label{fig:app_car_4}
\end{figure}

\begin{figure}[h!]
    \centering
    \includegraphics[width=0.8\textwidth]{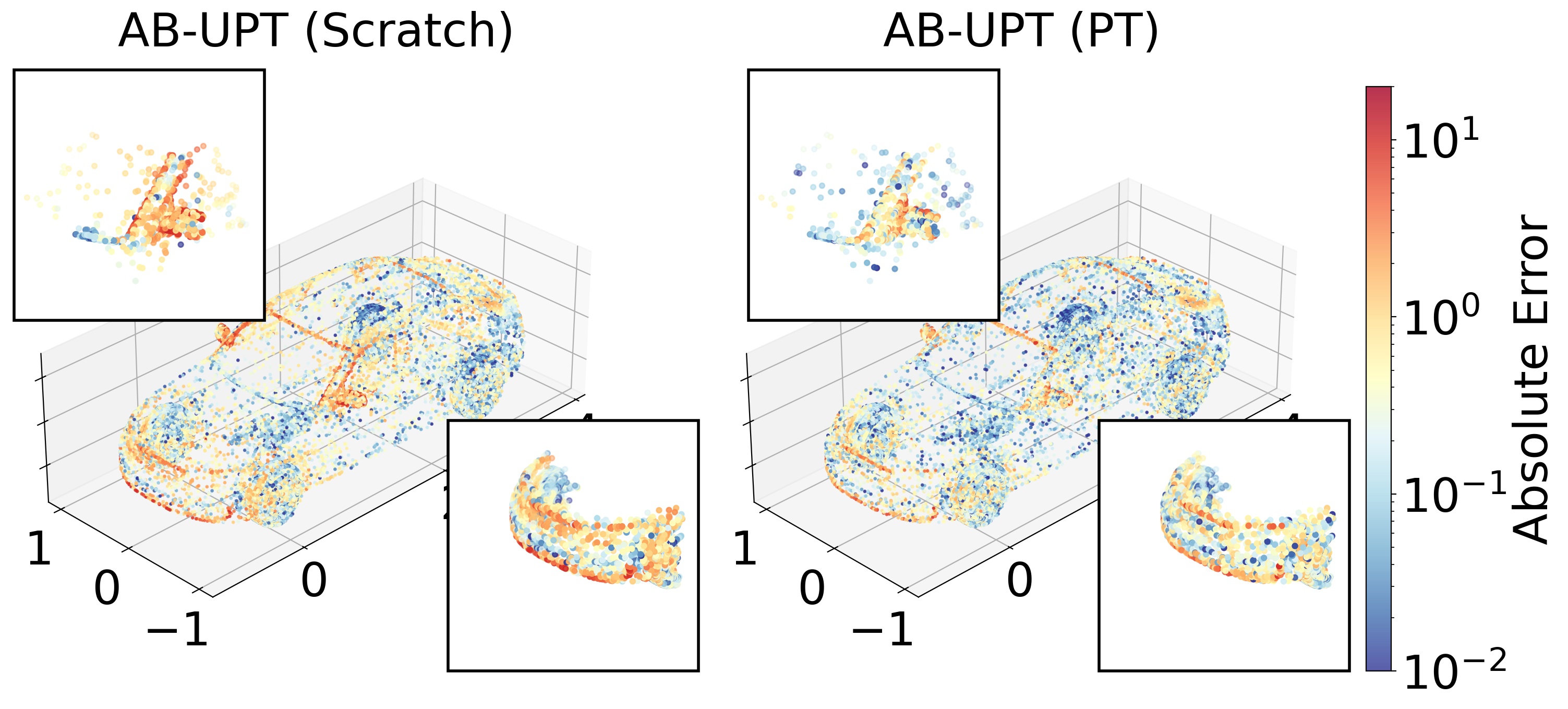} 
    \caption{DrivAerML surface wall shear stress absolute error comparison. 64 training samples for AB-UPT.}
    \label{fig:app_car_5}
\end{figure}

\begin{figure}[h!]
    \centering
    \includegraphics[width=0.8\textwidth]{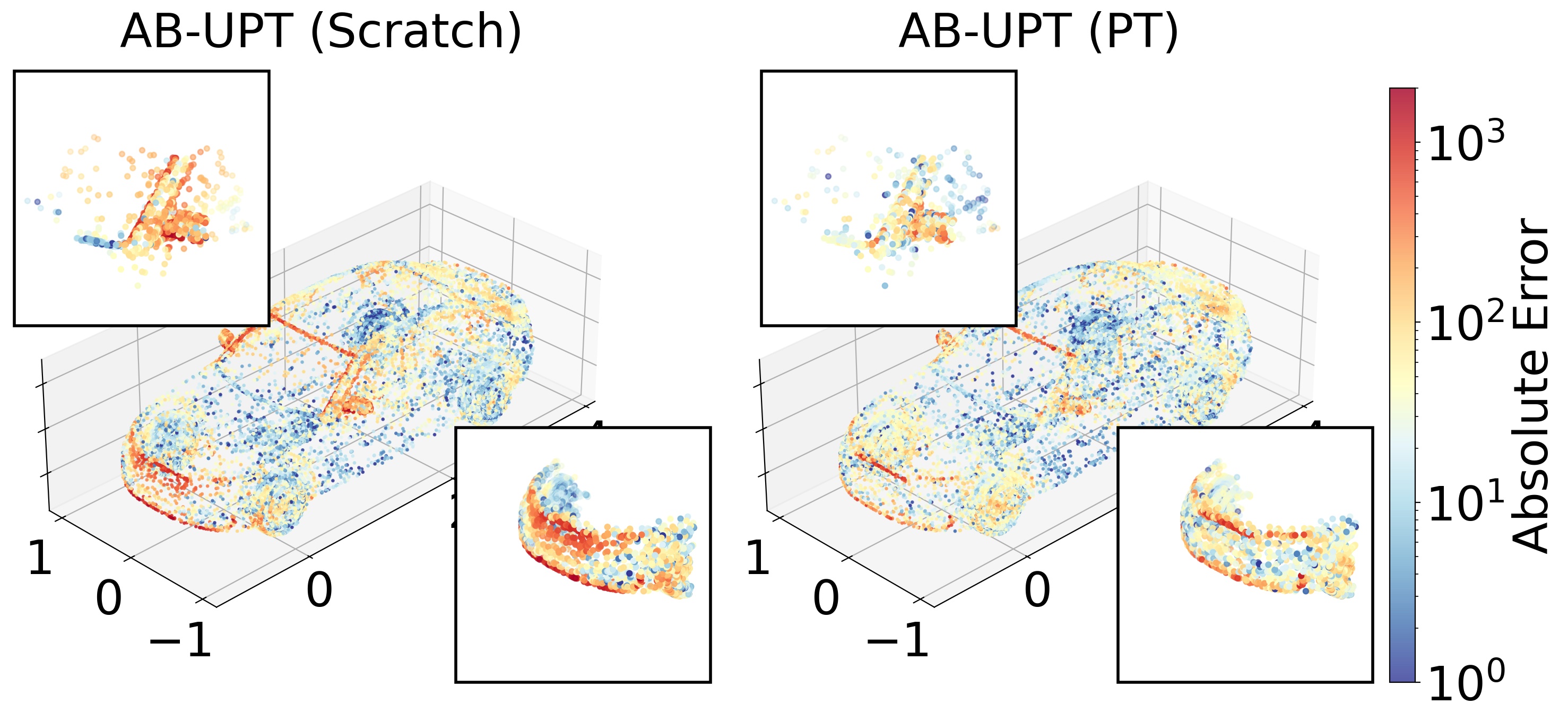} 
    \caption{DrivAerML surface pressure absolute error comparison. 64 training samples for AB-UPT.}
    \label{fig:app_car_6}
\end{figure}

\begin{figure}[h!]
    \centering
    \includegraphics[width=0.8\textwidth]{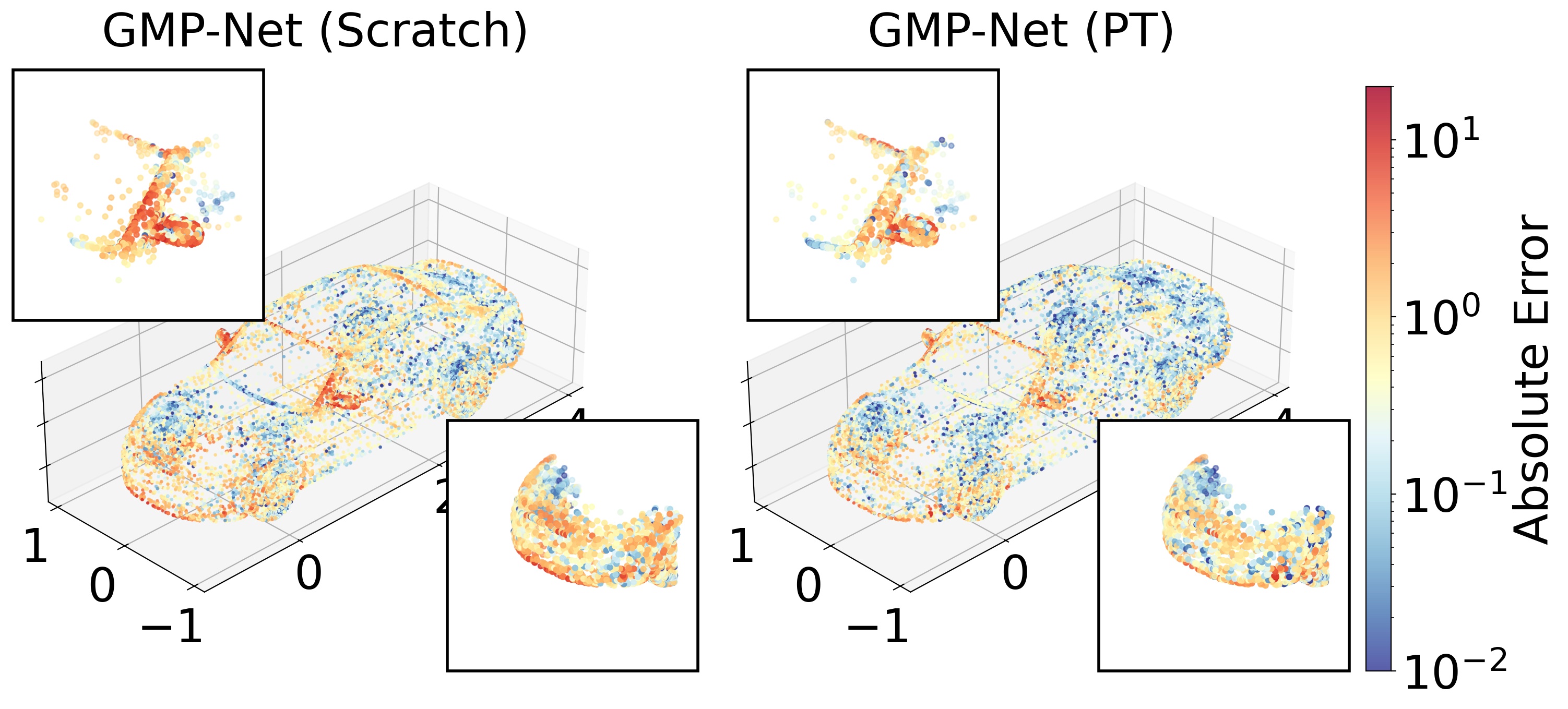} 
    \caption{DrivAerML surface wall shear stress absolute error comparison. 16 training samples for GMP-Net.}
    \label{fig:app_car_7}
\end{figure}

\begin{figure}[h!]
    \centering
    \includegraphics[width=0.8\textwidth]{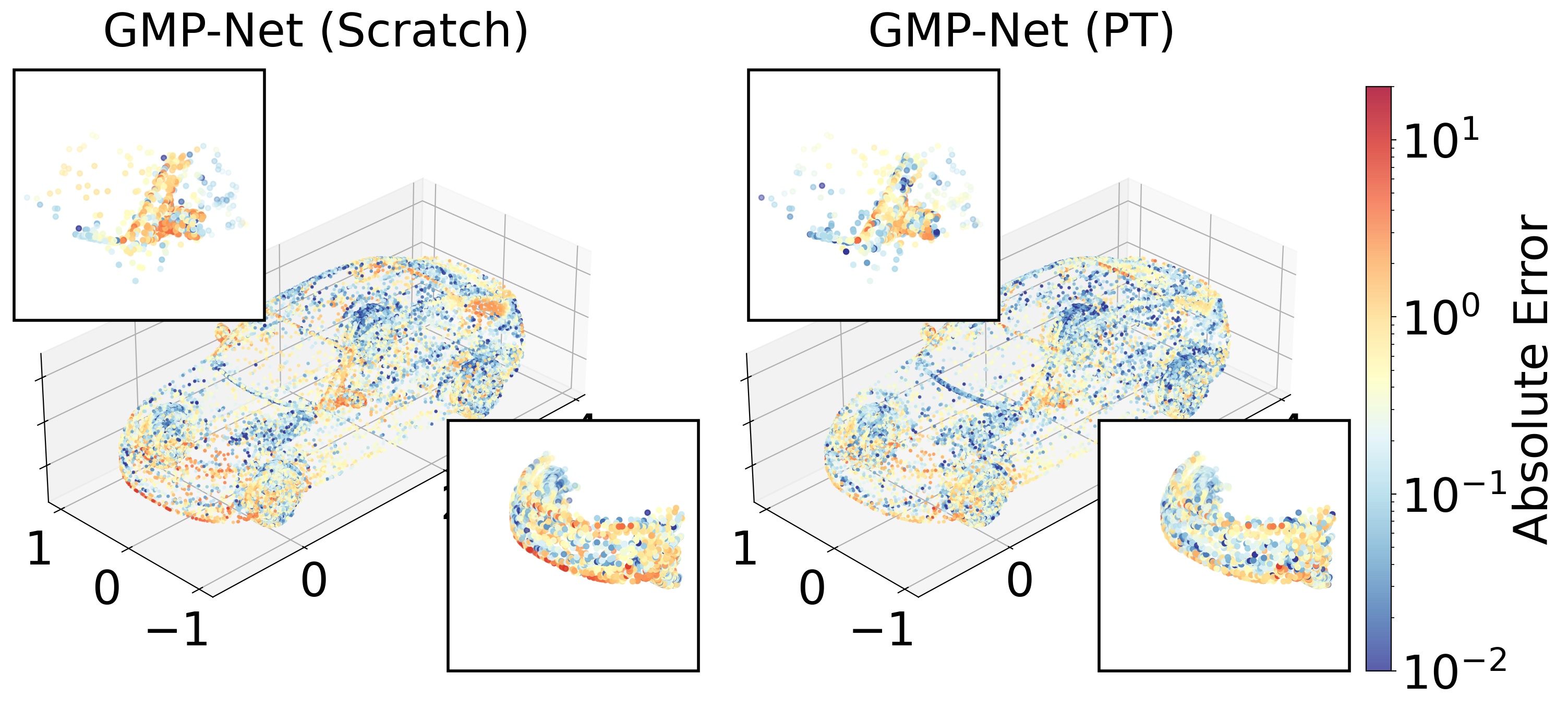} 
    \caption{DrivAerML surface wall shear stress absolute error comparison. 32 training samples for GMP-Net.}
    \label{fig:app_car_8}
\end{figure}

\begin{figure}[h!]
    \centering
    \includegraphics[width=0.8\textwidth]{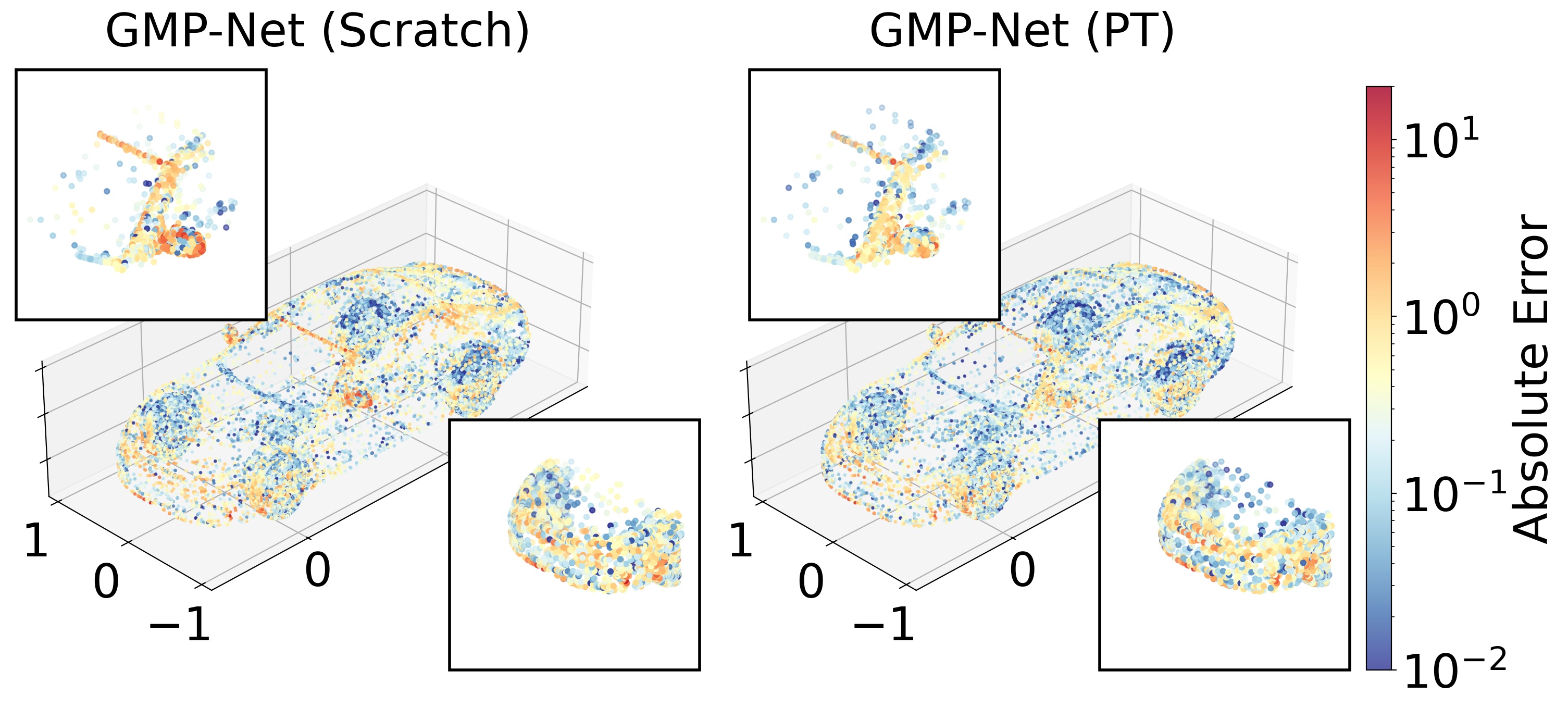} 
    \caption{DrivAerML surface wall shear stress absolute error comparison. 32 training samples for GMP-Net.}
    \label{fig:app_car_9}
\end{figure}

\begin{figure}[h!]
    \centering
    \includegraphics[width=0.8\textwidth]{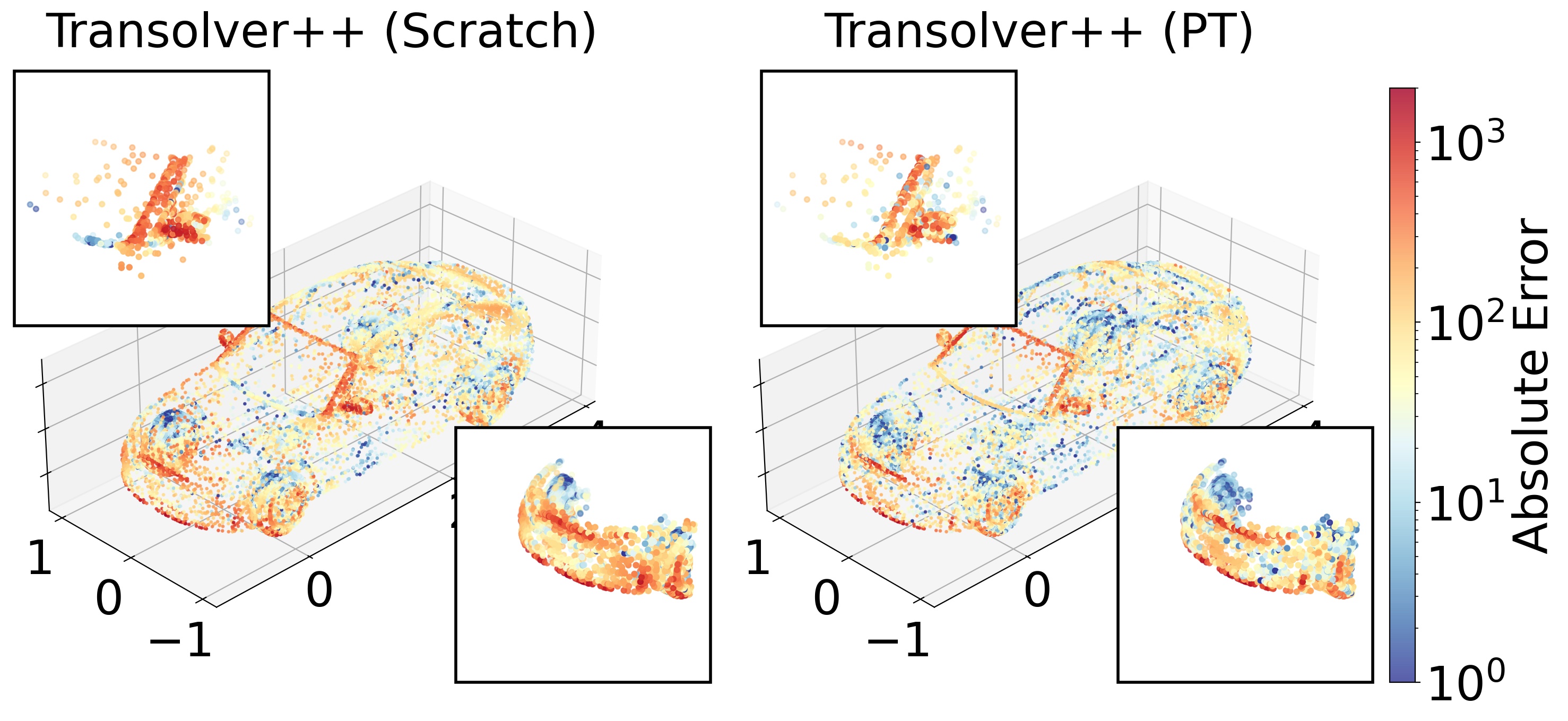} 
    \caption{DrivAerML surface pressure absolute error comparison. 16 training samples for Transolver++.}
    \label{fig:app_car_10}
\end{figure}

\begin{figure}[h!]
    \centering
    \includegraphics[width=0.8\textwidth]{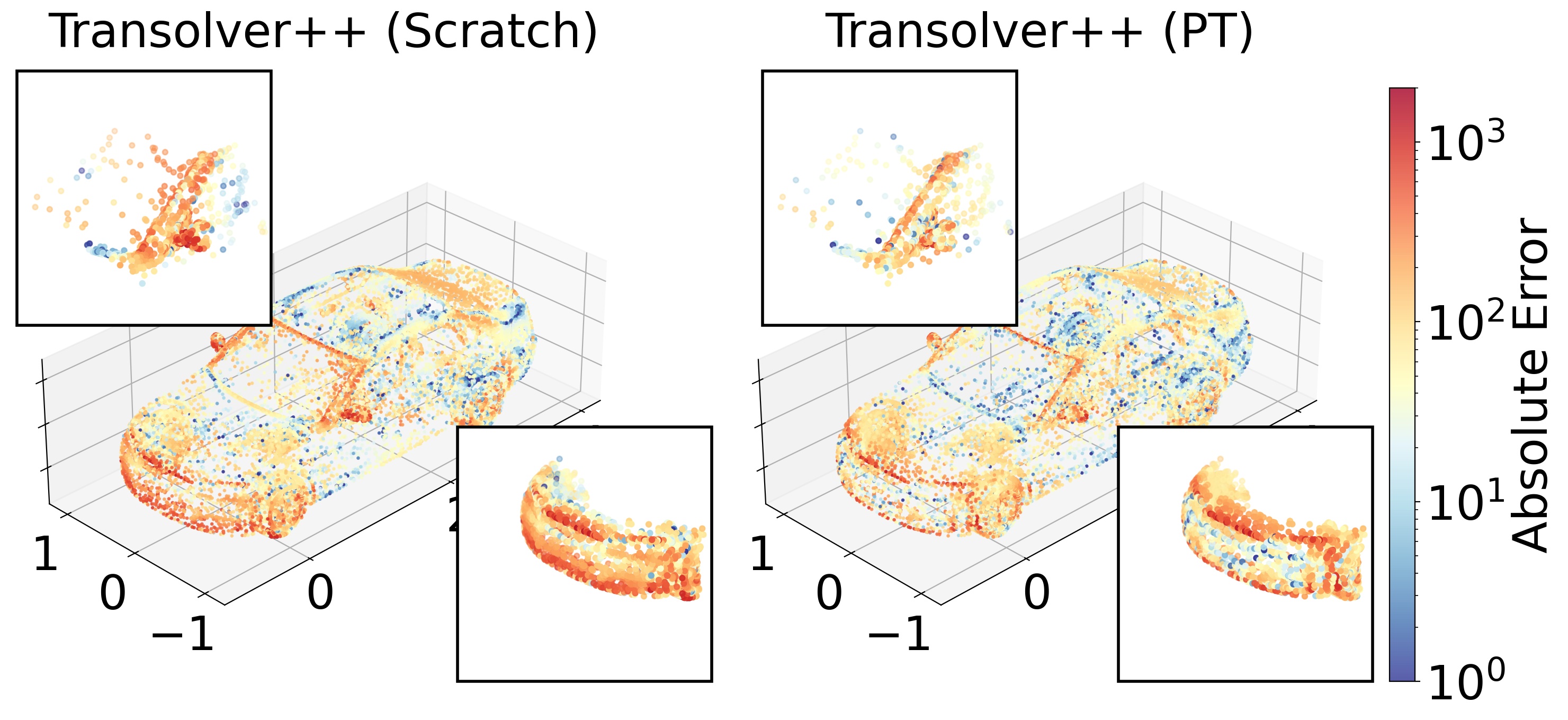} 
    \caption{DrivAerML surface pressure absolute error comparison. 16 training samples for Transolver++.}
    \label{fig:app_car_11}
\end{figure}

\begin{figure}[h!]
    \centering
    \includegraphics[width=0.8\textwidth]{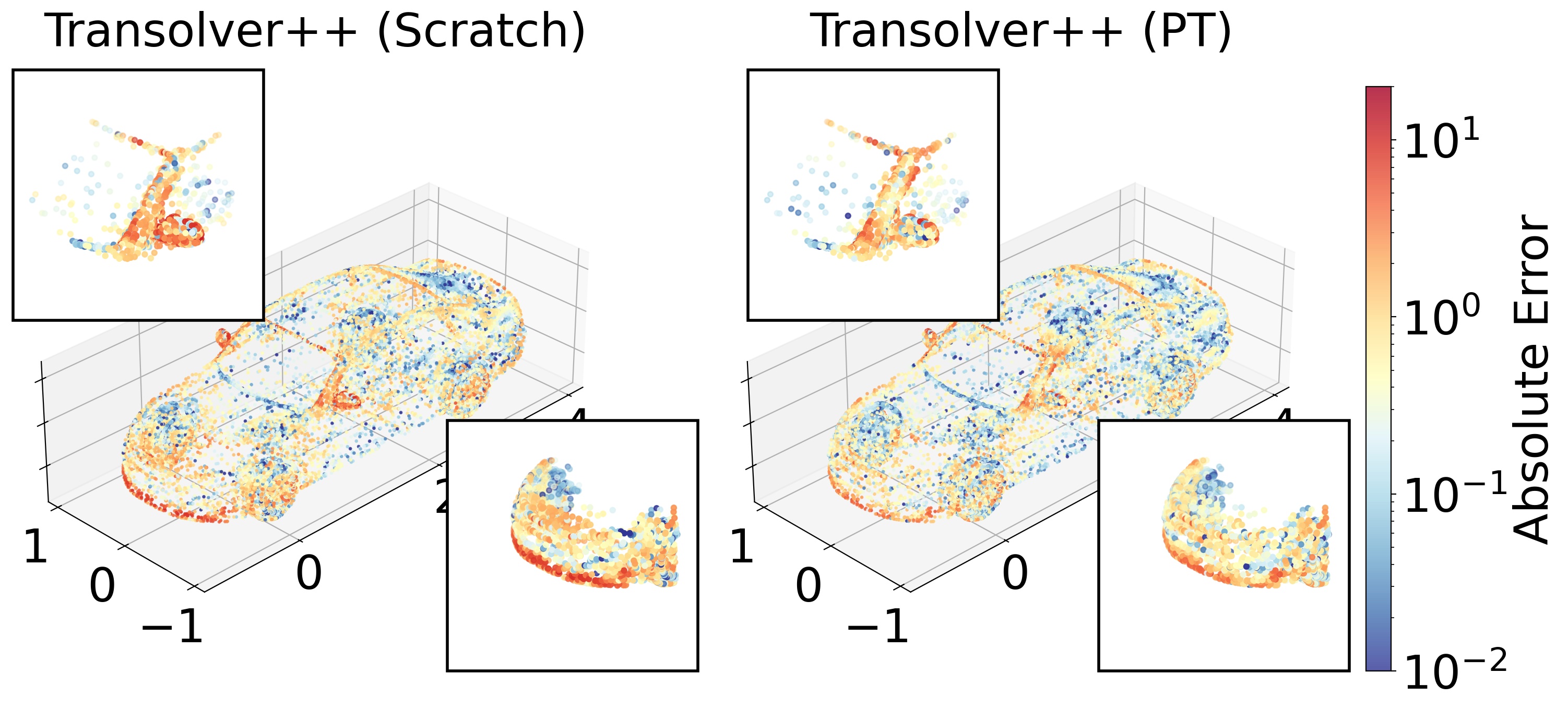} 
    \caption{DrivAerML surface wall shear stress absolute error comparison. 32 training samples for Transolver++.}
    \label{fig:app_car_12}
\end{figure}

\begin{figure}[h!]
    \centering
    \includegraphics[width=0.8\textwidth]{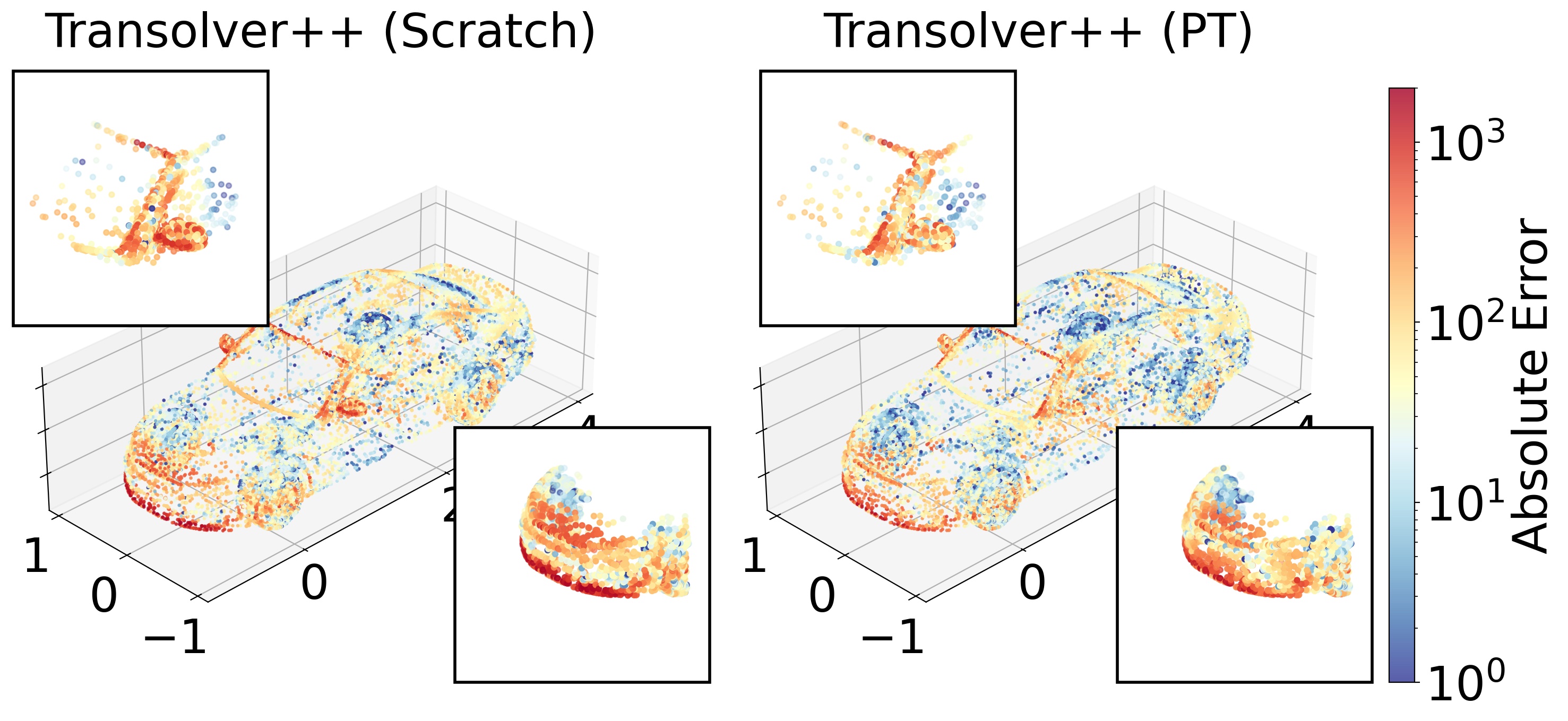} 
    \caption{DrivAerML surface pressure absolute error comparison. 32 training samples for Transolver++.}
    \label{fig:app_car_13}
\end{figure}

\begin{figure}[h!]
    \centering
    \includegraphics[width=0.8\textwidth]{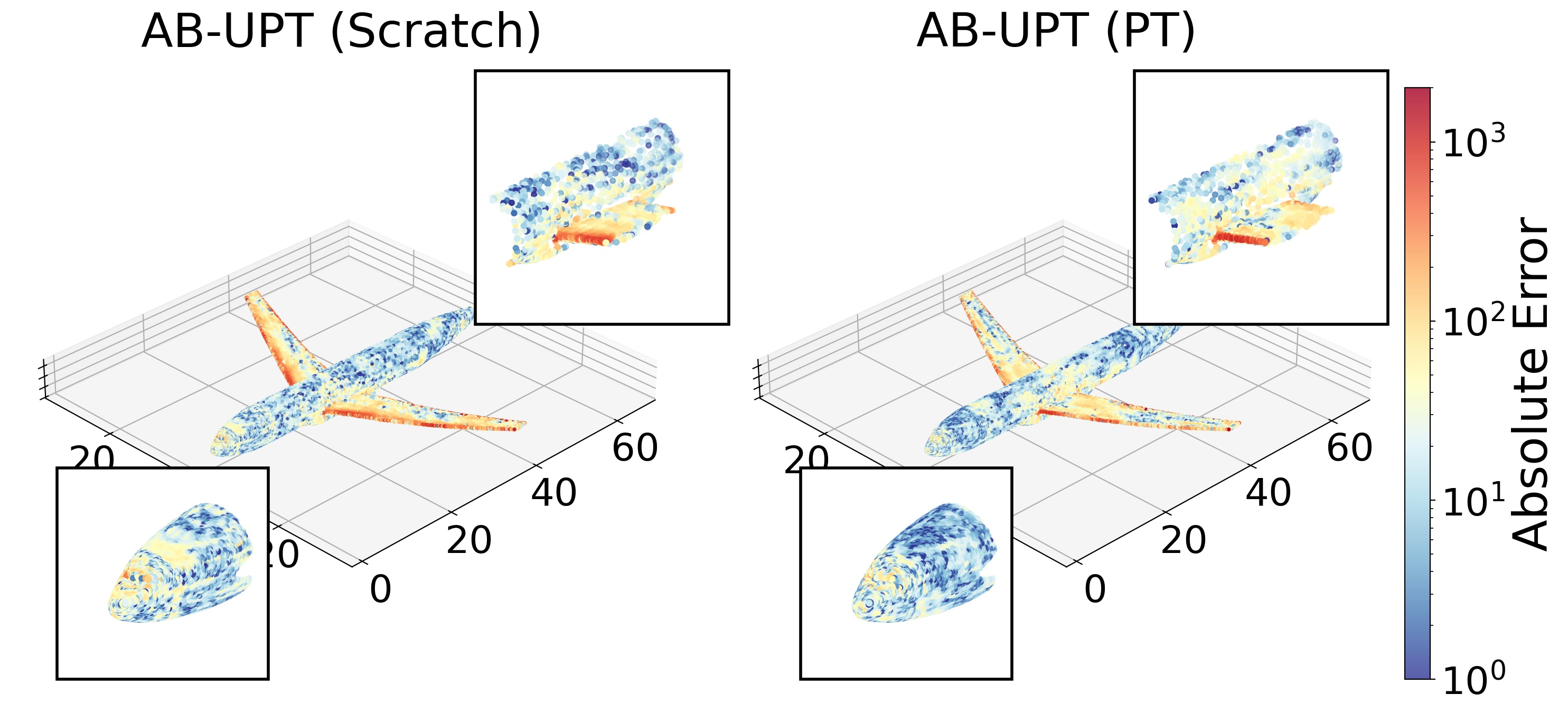} 
    \caption{SHIFT-Wing surface pressure absolute error comparison. 128 training samples for AB-UPT.}
    \label{fig:app_wing_1}
\end{figure}

\begin{figure}[h!]
    \centering
    \includegraphics[width=0.8\textwidth]{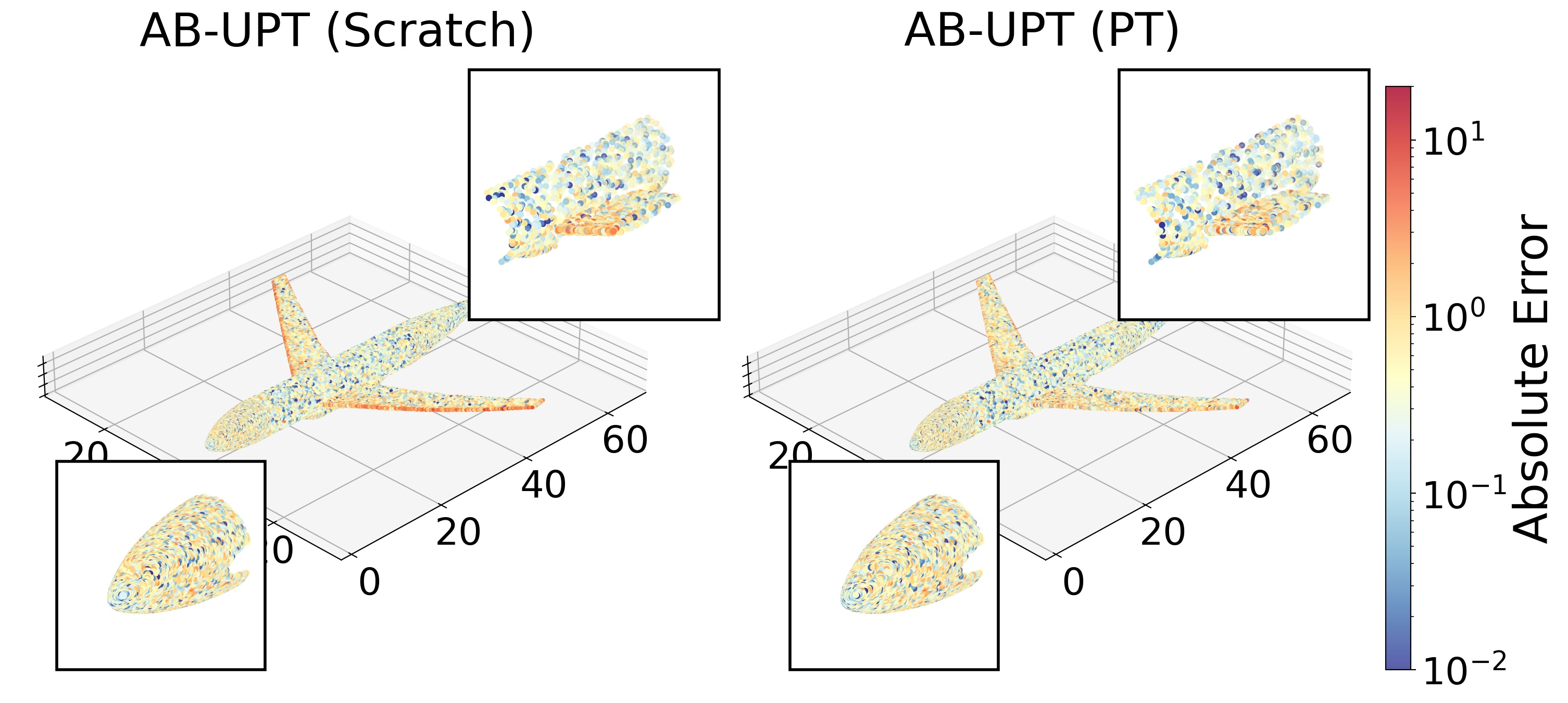} 
    \caption{SHIFT-Wing surface wall shear stress absolute error comparison. 128 training samples for AB-UPT.}
    \label{fig:app_wing_2}
\end{figure}

\begin{figure}[h!]
    \centering
    \includegraphics[width=0.8\textwidth]{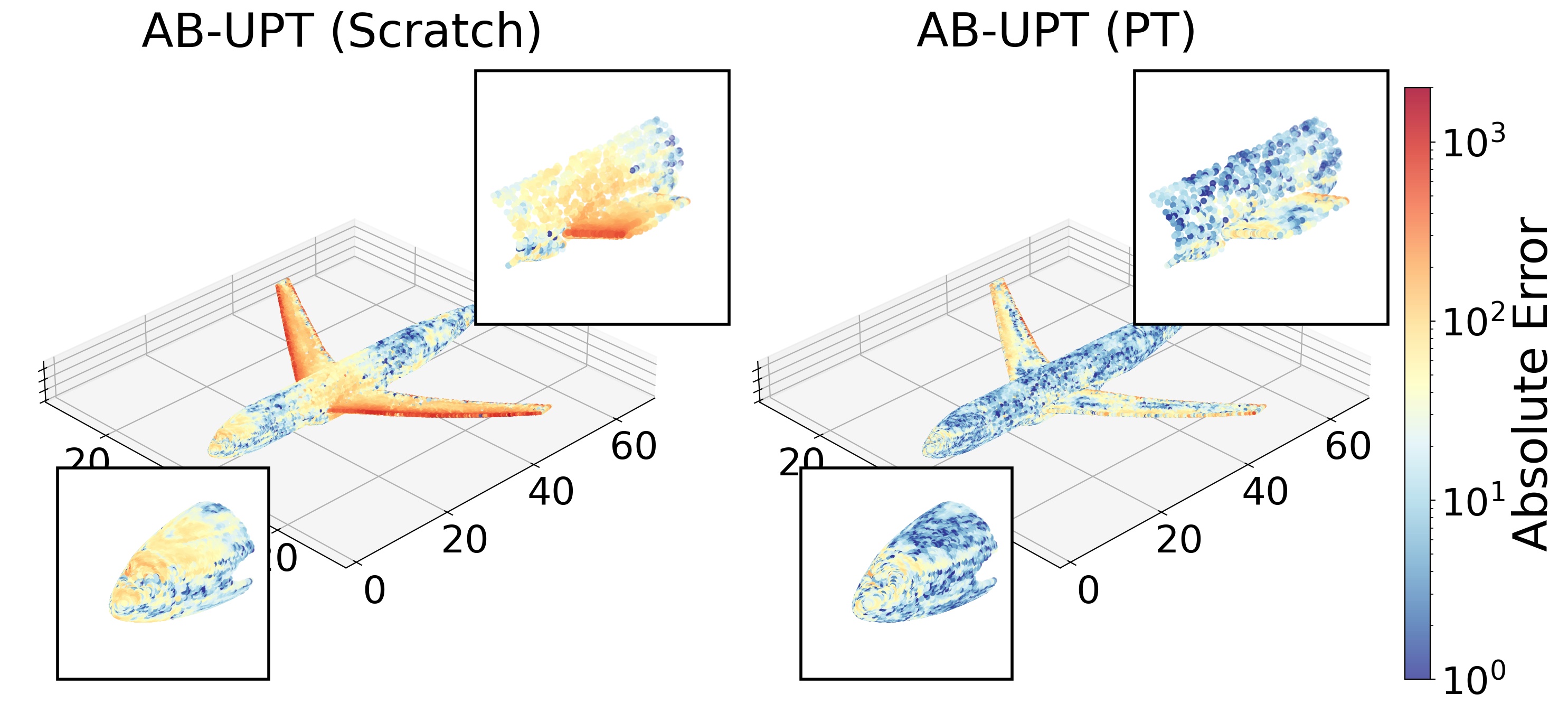} 
    \caption{SHIFT-Wing surface pressure absolute error comparison. 128 training samples for AB-UPT.}
    \label{fig:app_wing_3}
\end{figure}

\begin{figure}[h!]
    \centering
    \includegraphics[width=0.8\textwidth]{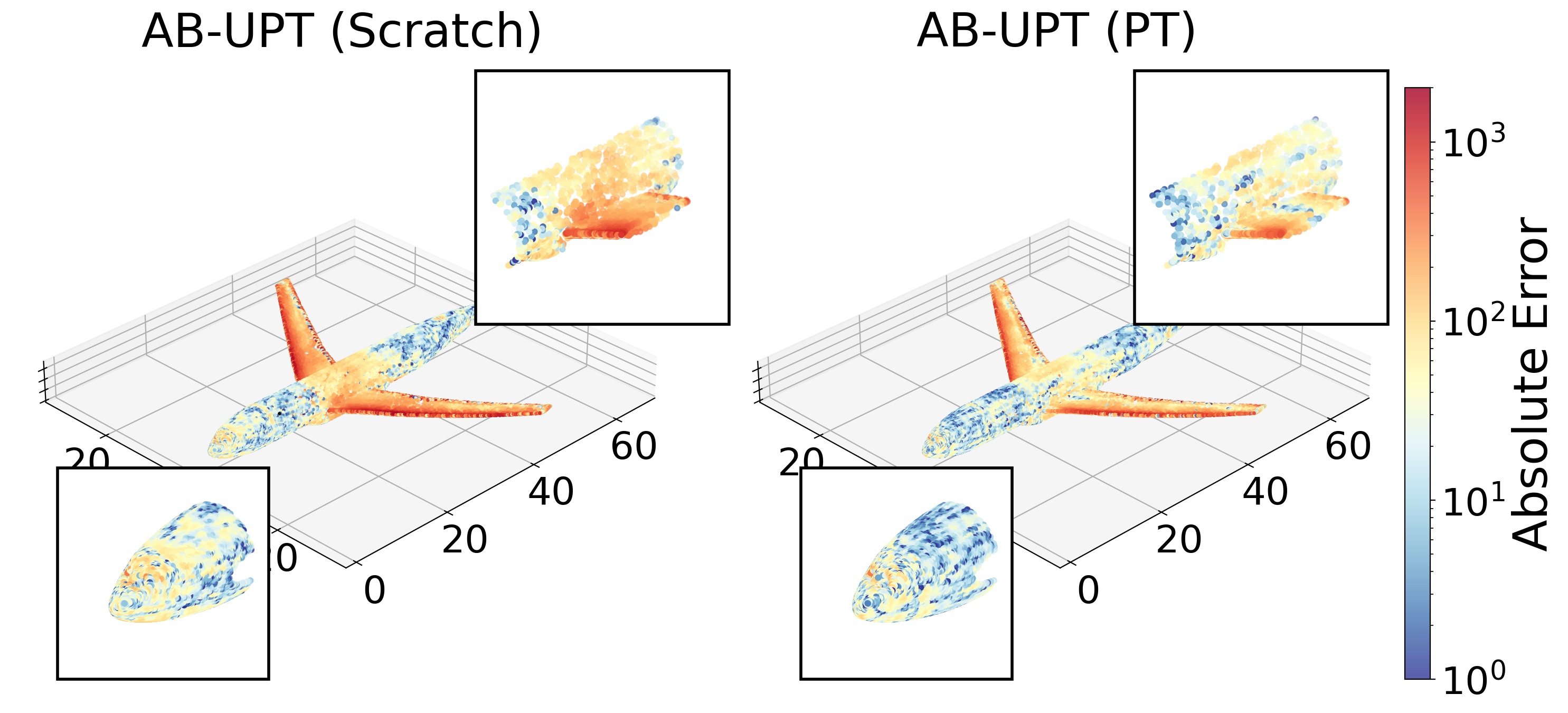} 
    \caption{SHIFT-Wing surface pressure absolute error comparison. 64 training samples for AB-UPT.}
    \label{fig:app_wing_4}
\end{figure}

\begin{figure}[h!]
    \centering
    \includegraphics[width=0.8\textwidth]{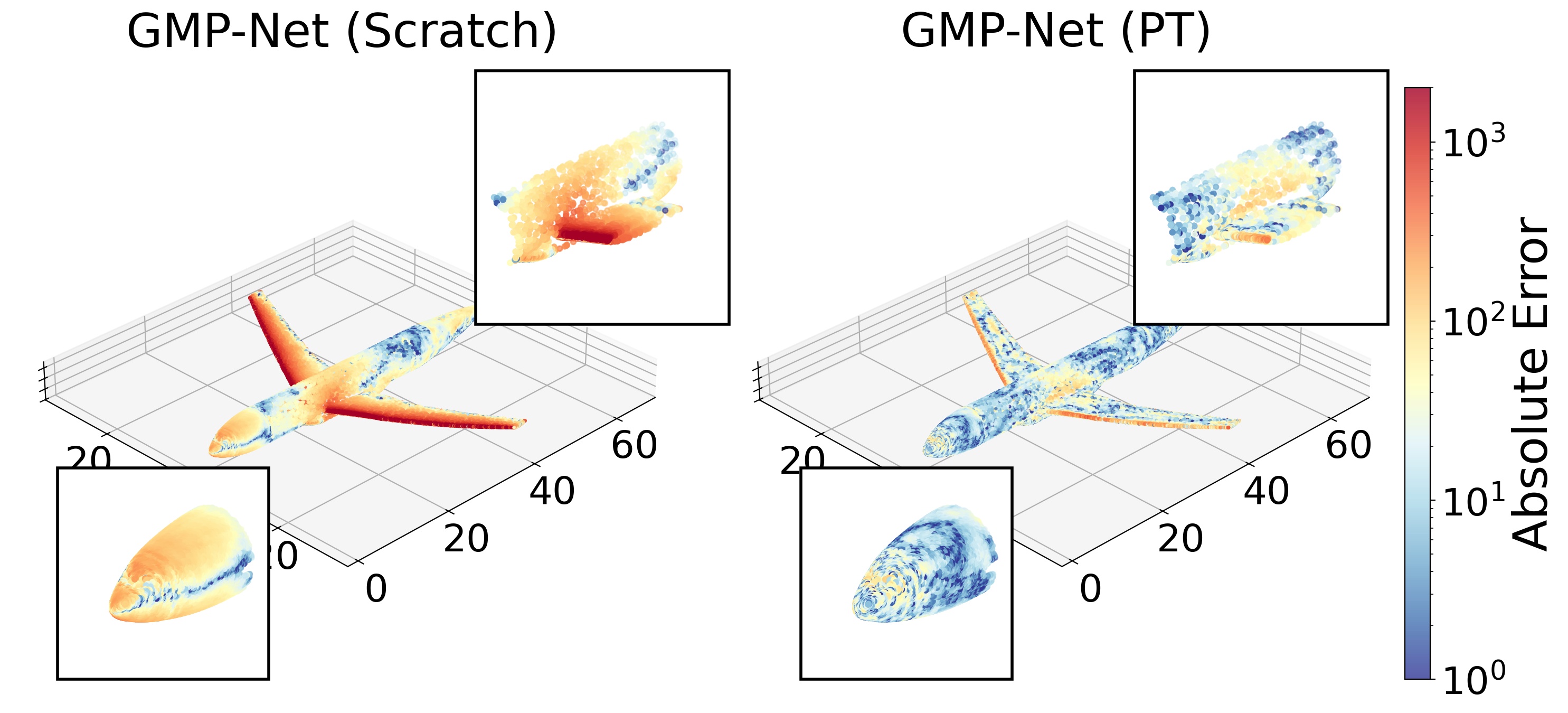} 
    \caption{SHIFT-Wing surface pressure absolute error comparison. 128 training samples for GMP-Net.}
    \label{fig:app_wing_5}
\end{figure}

\begin{figure}[h!]
    \centering
    \includegraphics[width=0.8\textwidth]{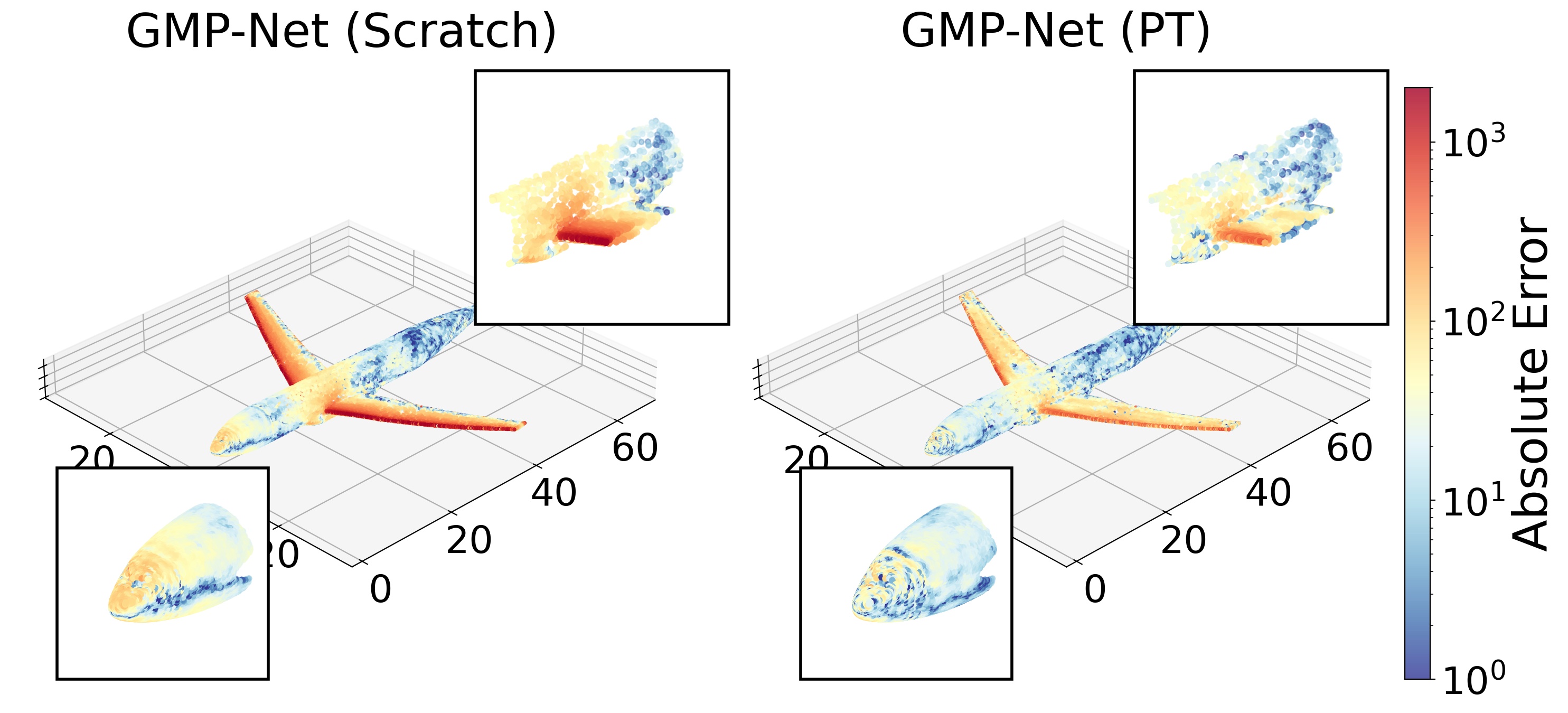} 
    \caption{SHIFT-Wing surface pressure absolute error comparison. 128 training samples for GMP-Net.}
    \label{fig:app_wing_6}
\end{figure}

\begin{figure}[h!]
    \centering
    \includegraphics[width=0.8\textwidth]{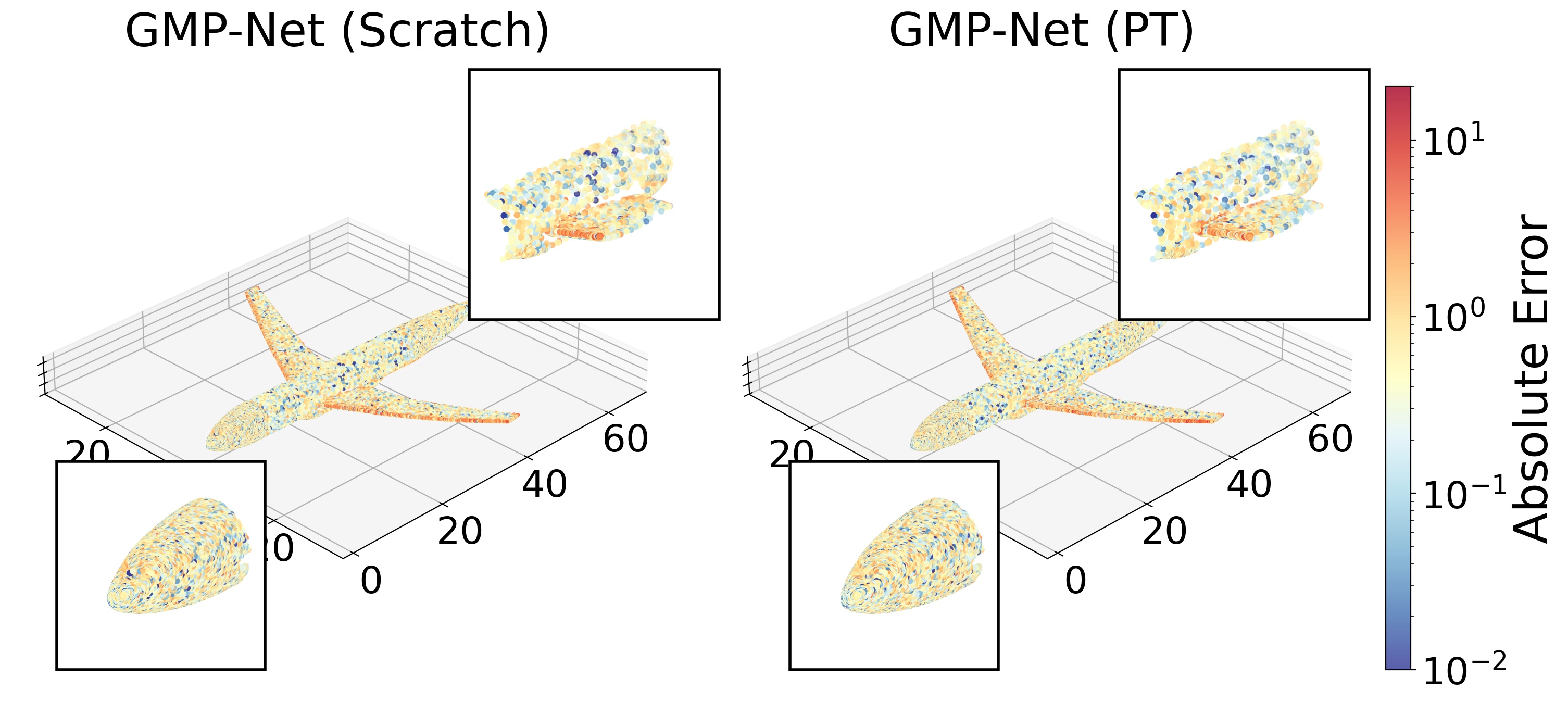} 
    \caption{SHIFT-Wing surface wall shear stress absolute error comparison. 16 training samples for GMP-Net.}
    \label{fig:app_wing_7}
\end{figure}

\begin{figure}[h!]
    \centering
    \includegraphics[width=0.8\textwidth]{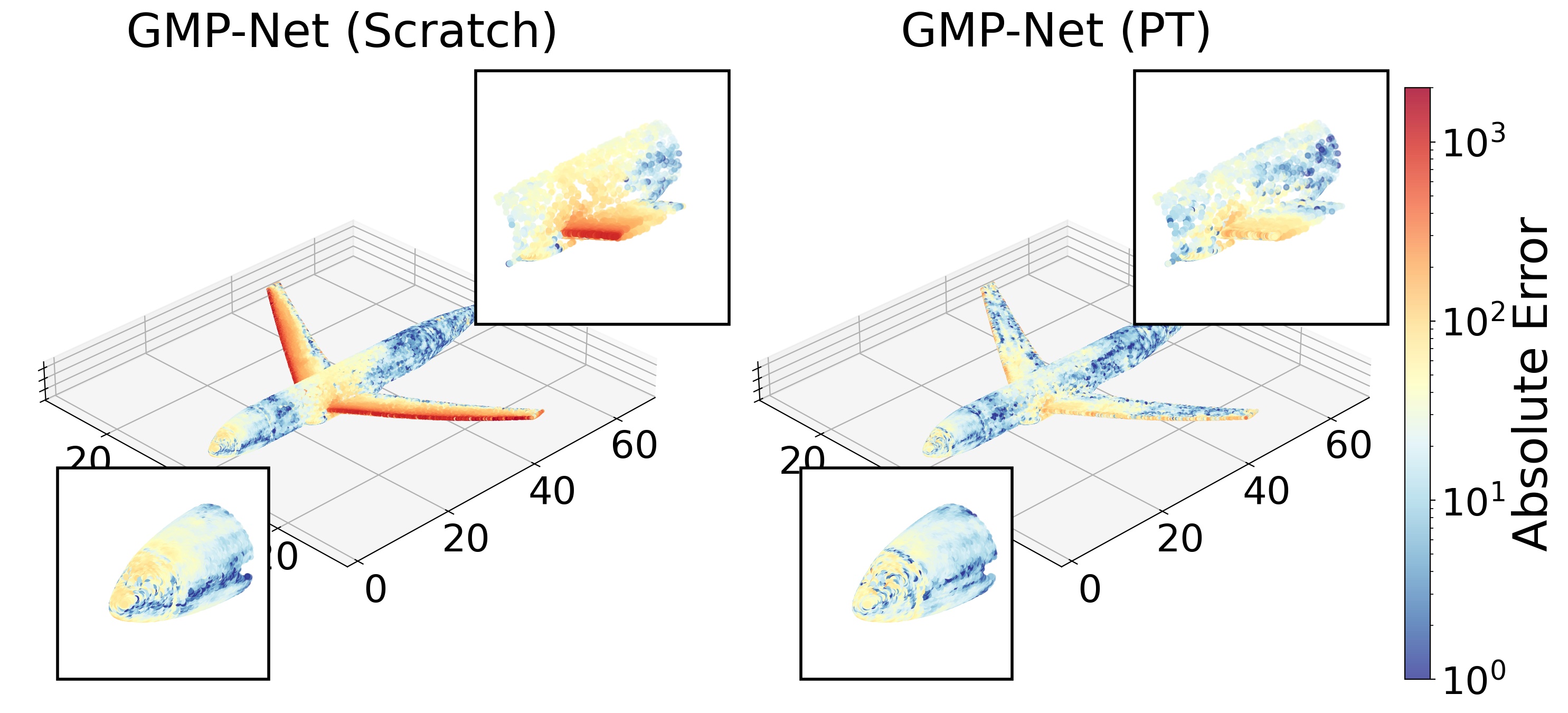} 
    \caption{SHIFT-Wing surface pressure absolute error comparison. 64 training samples for GMP-Net.}
    \label{fig:app_wing_8}
\end{figure}

\begin{figure}[h!]
    \centering
    \includegraphics[width=0.8\textwidth]{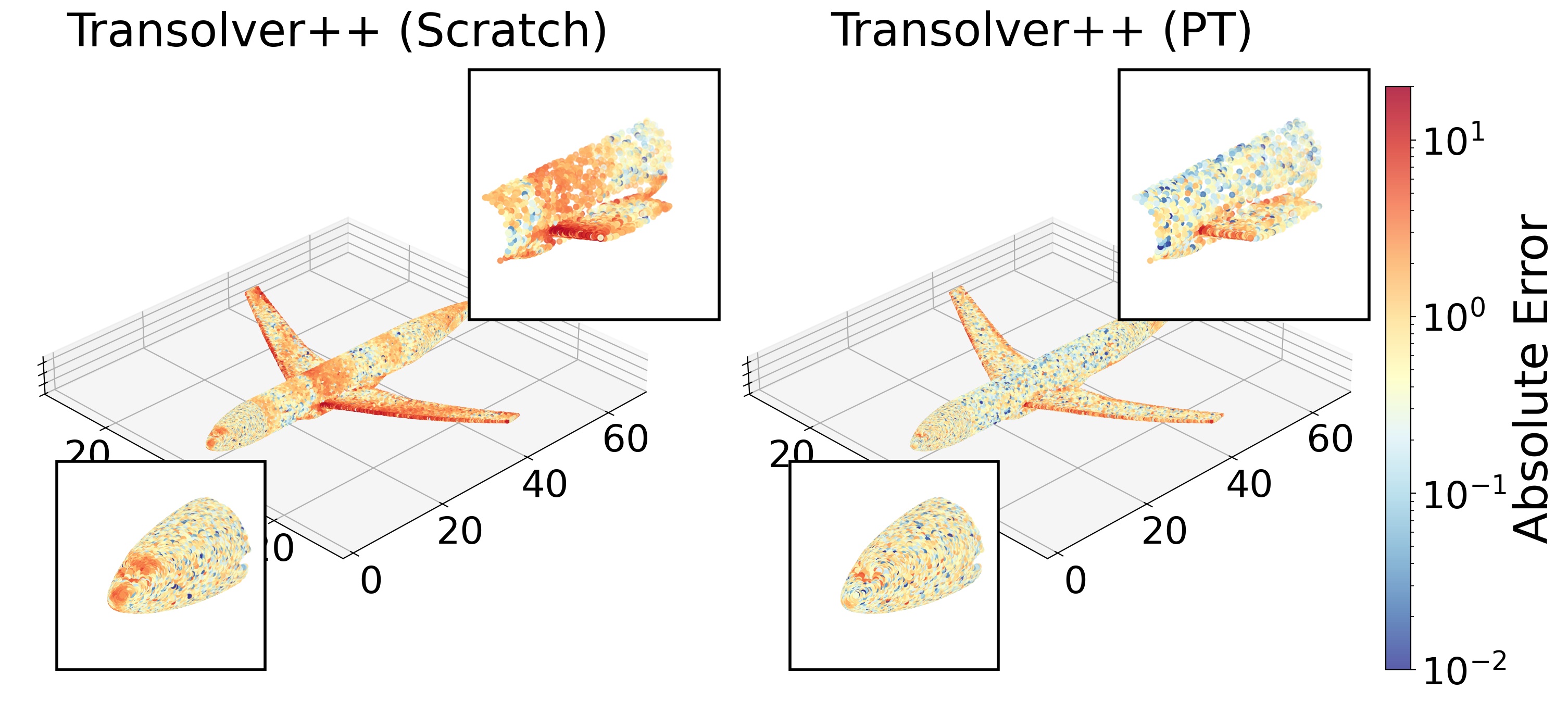} 
    \caption{SHIFT-Wing surface wall shear stress absolute error comparison. 16 training samples for Transolver++.}
    \label{fig:app_wing_9}
\end{figure}

\begin{figure}[h!]
    \centering
    \includegraphics[width=0.8\textwidth]{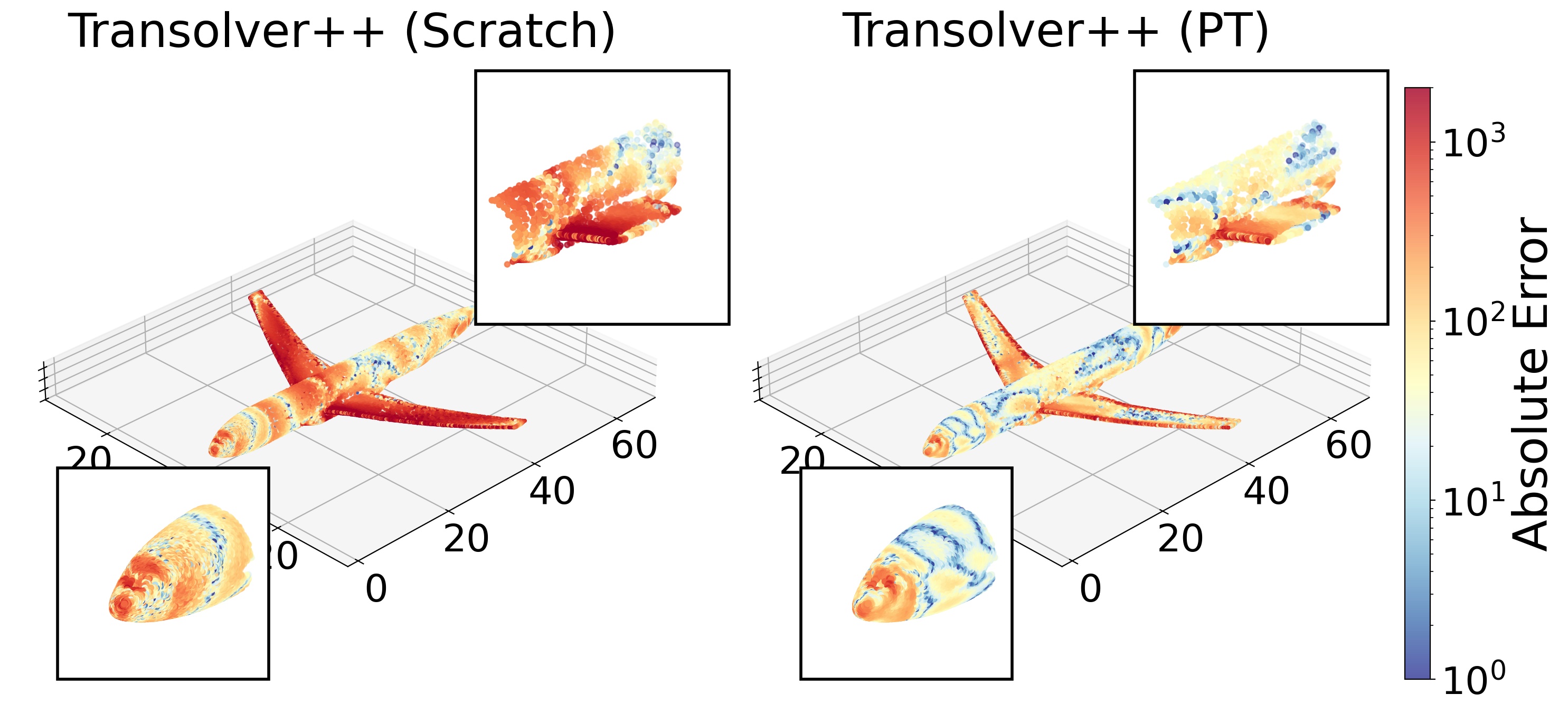} 
    \caption{SHIFT-Wing surface pressure absolute error comparison. 16 training samples for Transolver++.}
    \label{fig:app_wing_10}
\end{figure}

\begin{figure}[h!]
    \centering
    \includegraphics[width=0.8\textwidth]{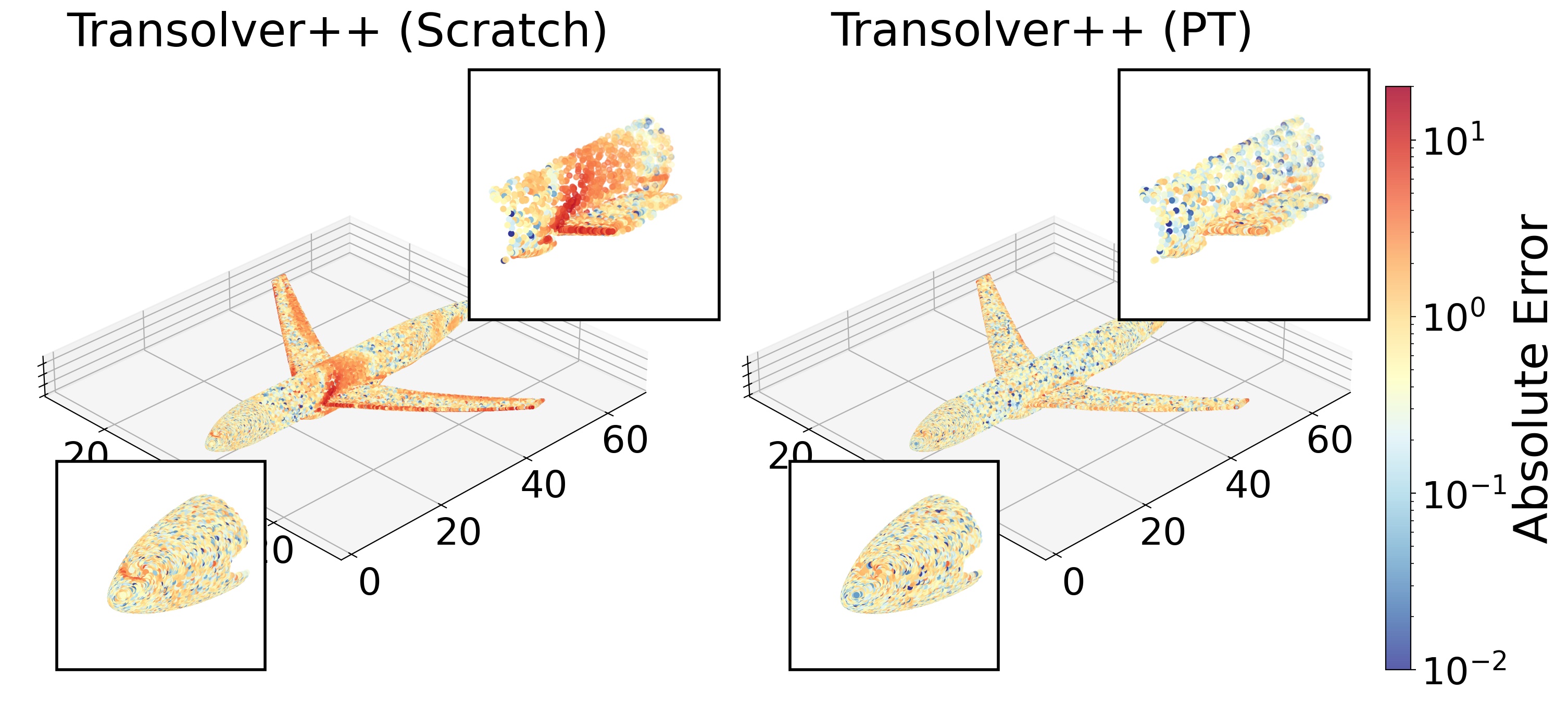} 
    \caption{SHIFT-Wing surface wall shear stress absolute error comparison. 32 training samples for Transolver++.}
    \label{fig:app_wing_11}
\end{figure}

\begin{figure}[h!]
    \centering
    \includegraphics[width=0.8\textwidth]{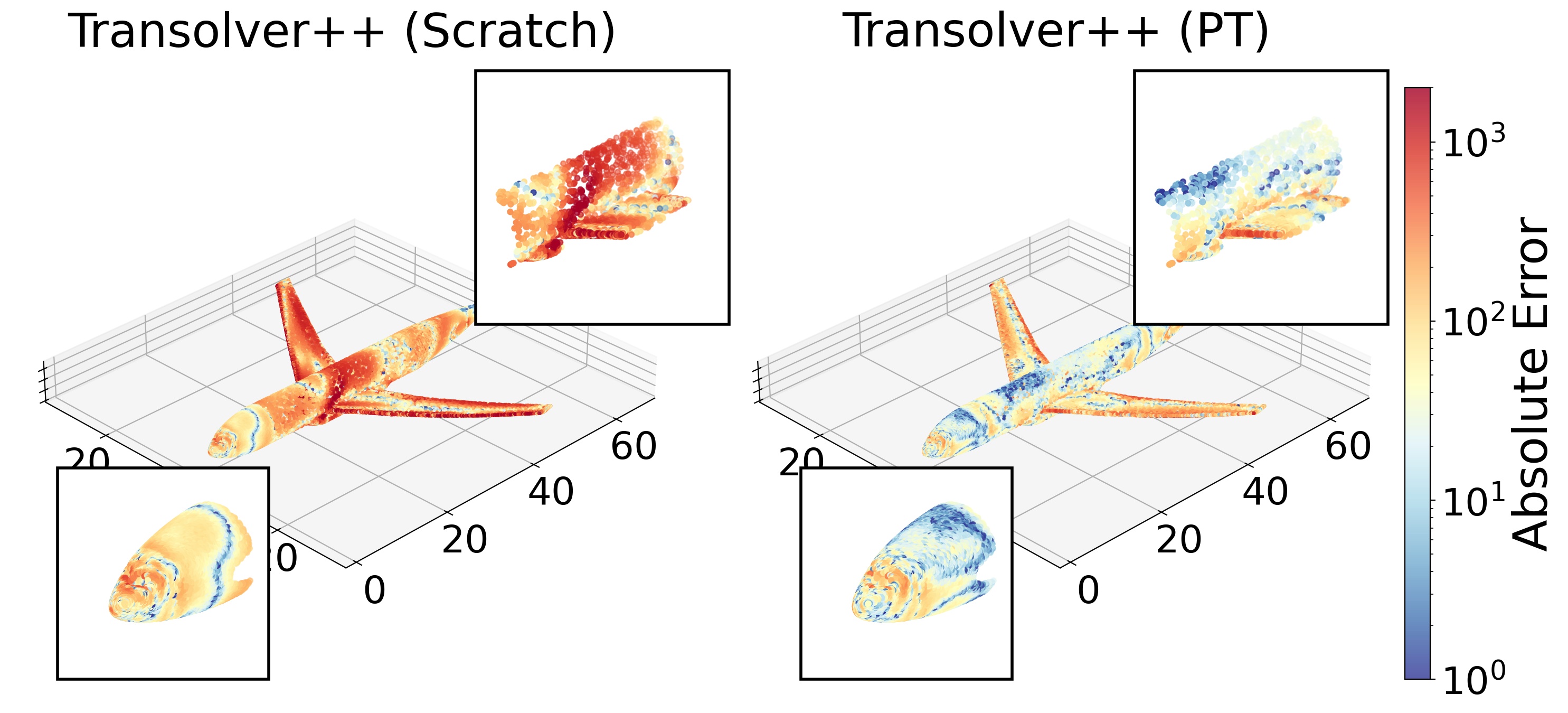} 
    \caption{SHIFT-Wing surfacepressure absolute error comparison. 32 training samples for Transolver++.}
    \label{fig:app_wing_12}
\end{figure}

\begin{figure}[h!]
    \centering
    \includegraphics[width=0.8\textwidth]{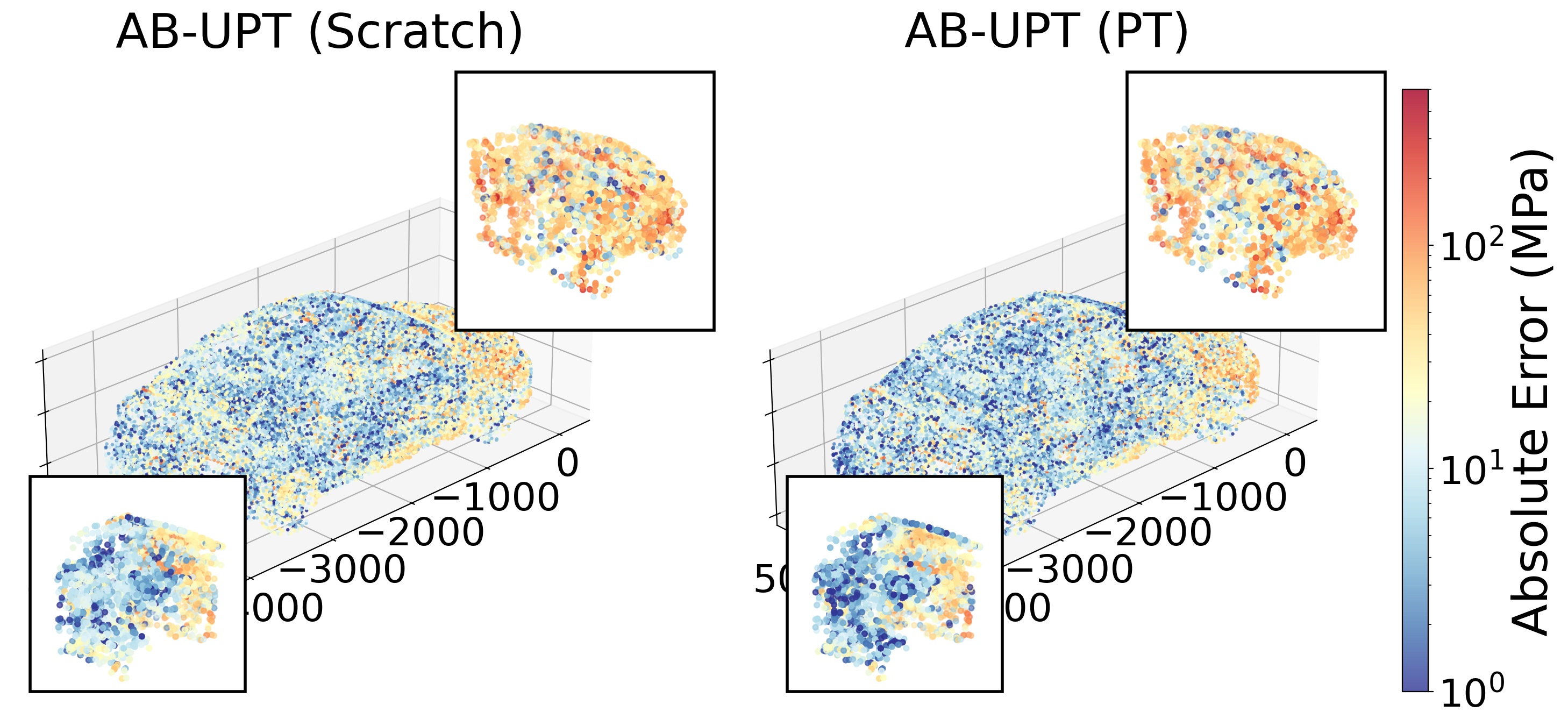} 
    \caption{SHIFT-Crash Von Mises stress absolute error comparison. 16 training samples for AB-UPT.}
    \label{fig:app_crash_1}
\end{figure}

\begin{figure}[h!]
    \centering
    \includegraphics[width=0.8\textwidth]{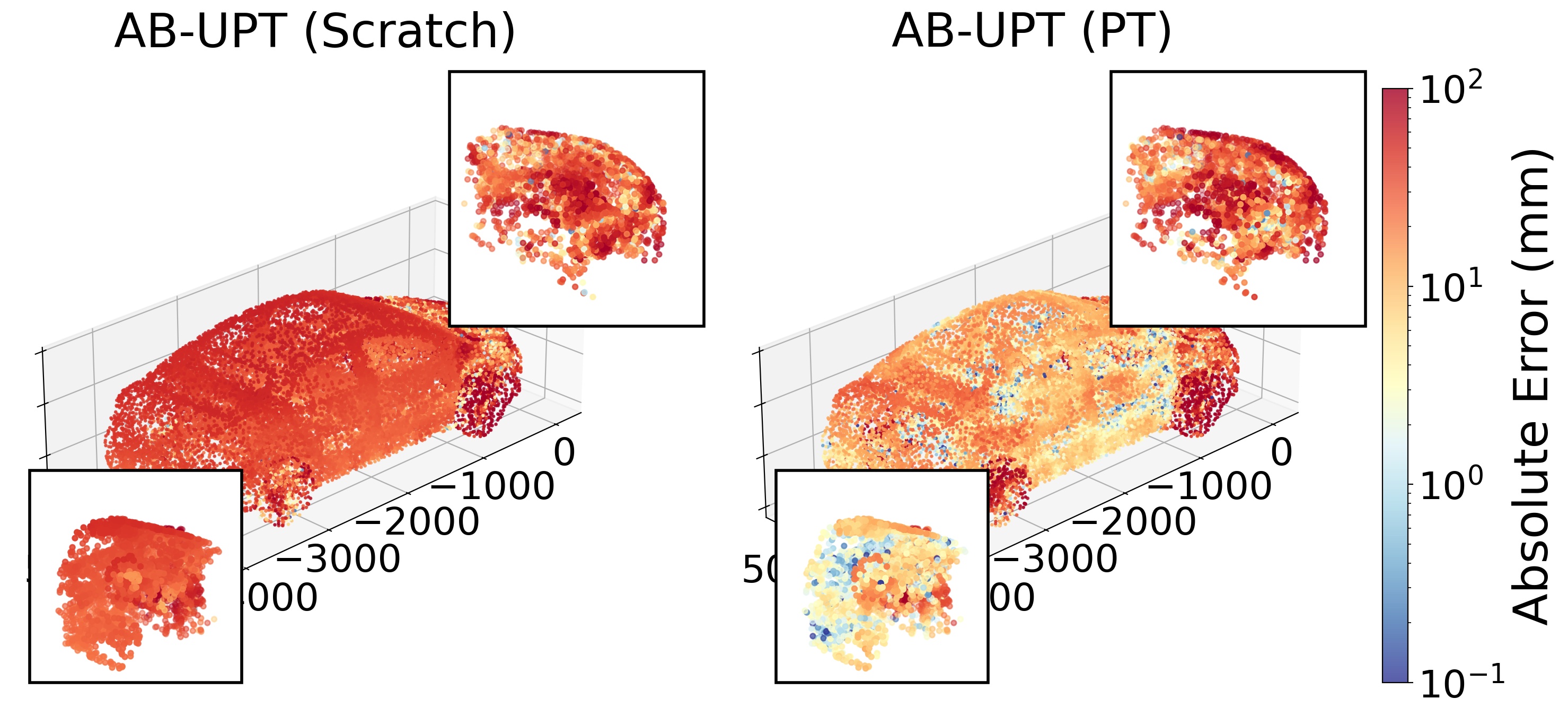} 
    \caption{SHIFT-Crash dissplacement absolute error comparison. 16 training samples for AB-UPT.}
    \label{fig:app_crash_2}
\end{figure}

\begin{figure}[h!]
    \centering
    \includegraphics[width=0.8\textwidth]{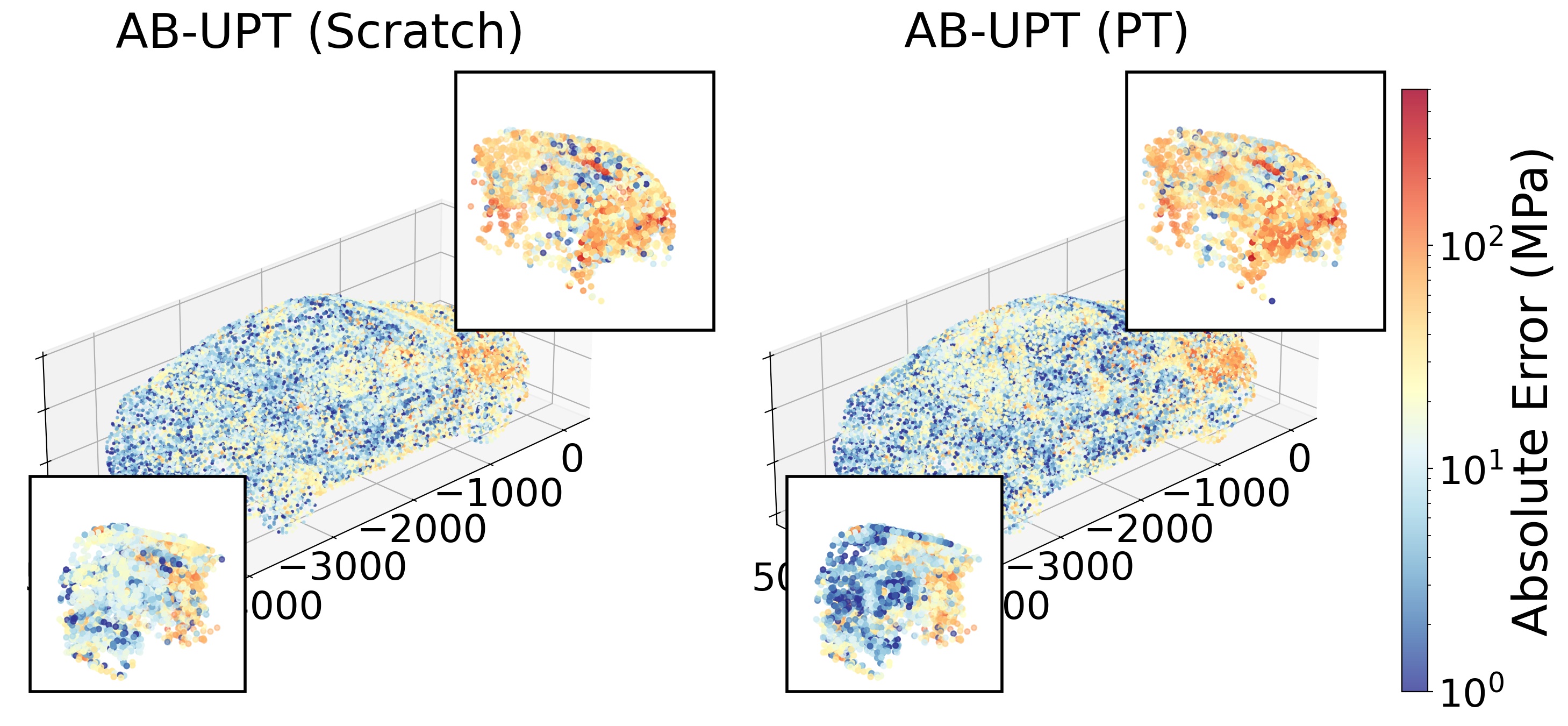} 
    \caption{SHIFT-Crash Von Mises stress absolute error comparison. 16 training samples for AB-UPT.}
    \label{fig:app_crash_3}
\end{figure}

\begin{figure}[h!]
    \centering
    \includegraphics[width=0.8\textwidth]{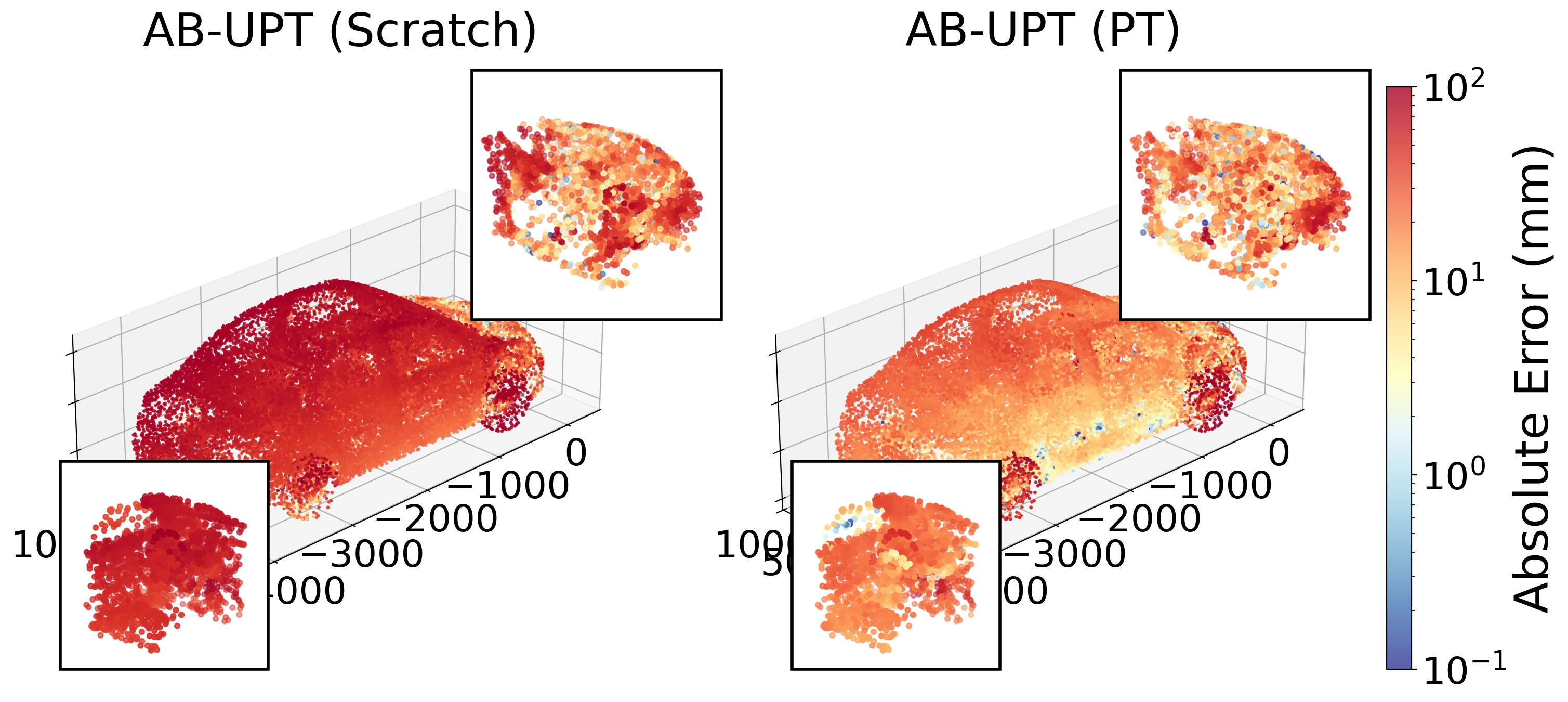} 
    \caption{SHIFT-Crash displacement absolute error comparison. 64 training samples for AB-UPT.}
    \label{fig:app_crash_4}
\end{figure}

\begin{figure}[h!]
    \centering
    \includegraphics[width=0.8\textwidth]{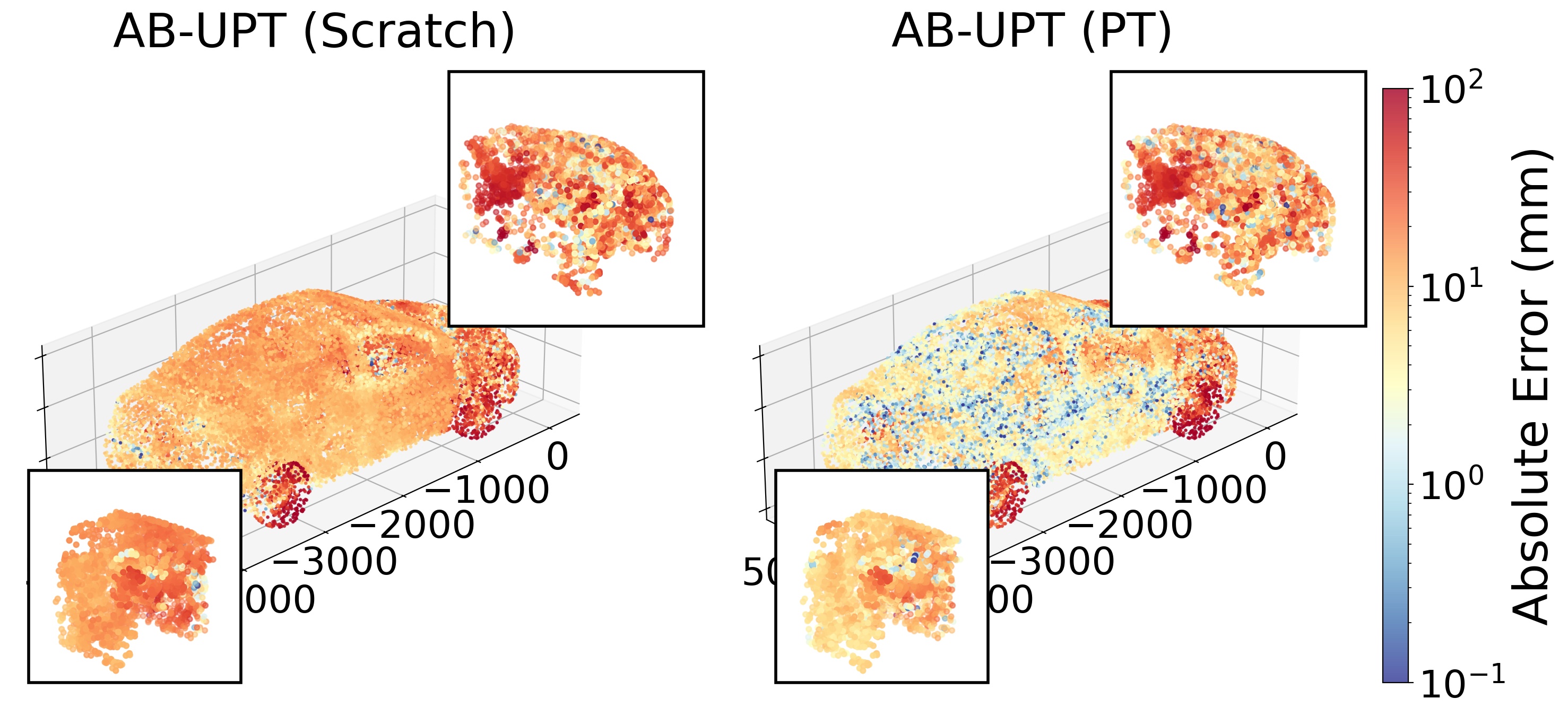} 
    \caption{SHIFT-Crash displacement absolute error comparison. 64 training samples for AB-UPT.}
    \label{fig:app_crash_5}
\end{figure}

\begin{figure}[h!]
    \centering
    \includegraphics[width=0.8\textwidth]{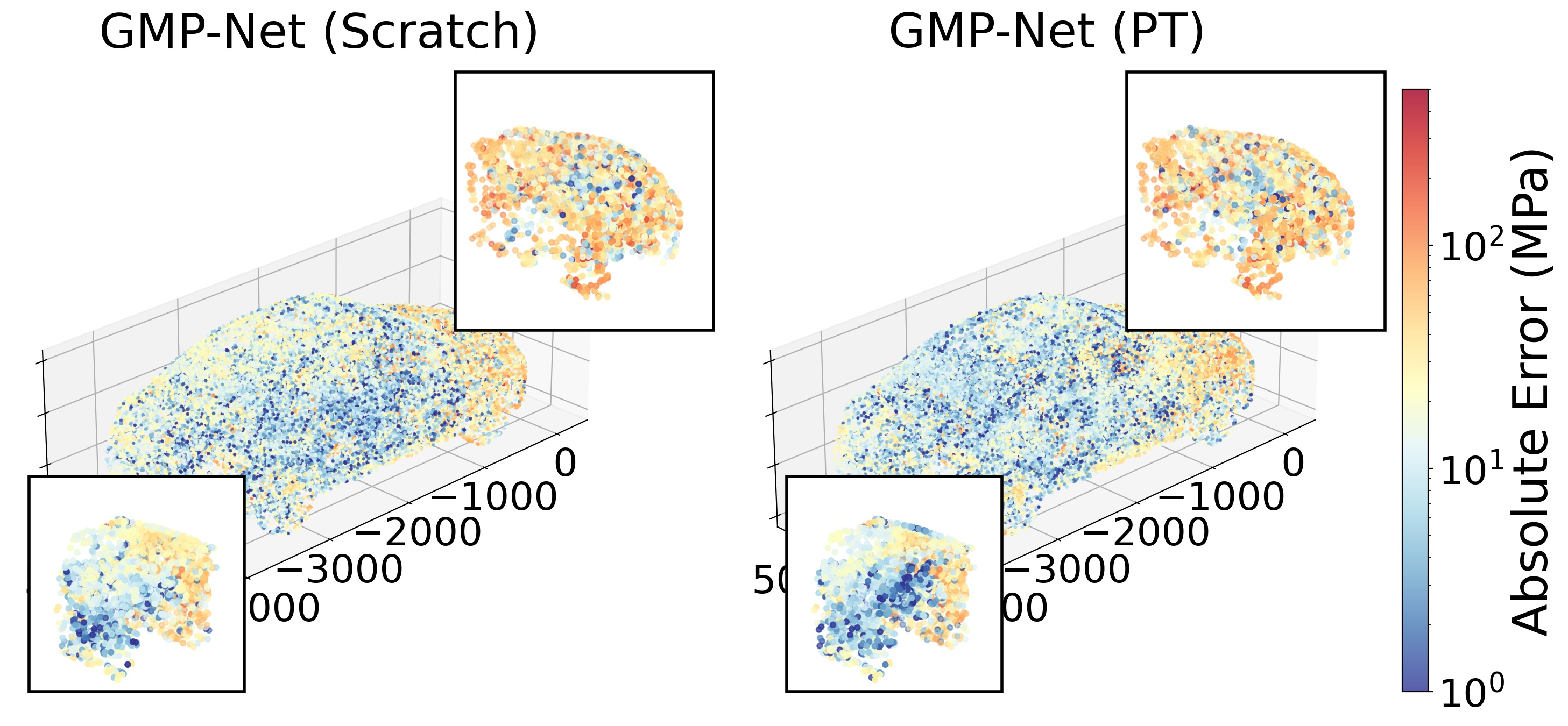} 
    \caption{SHIFT-Crash Von Mises stress absolute error comparison. 16 training samples for GMP-Net.}
    \label{fig:app_crash_6}
\end{figure}

\begin{figure}[h!]
    \centering
    \includegraphics[width=0.8\textwidth]{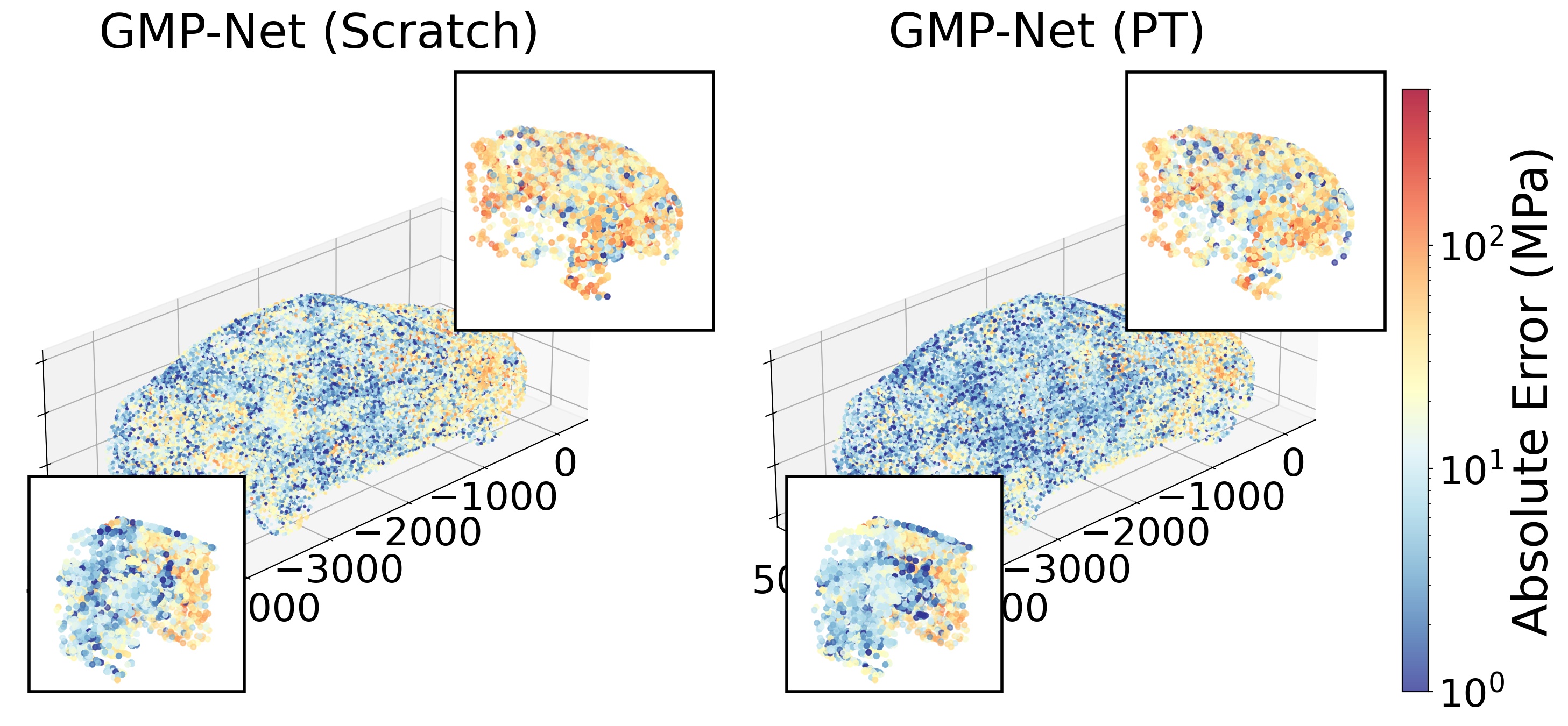} 
    \caption{SHIFT-Crash Von Mises stress absolute error comparison. 32 training samples for GMP-Net.}
    \label{fig:app_crash_7}
\end{figure}

\begin{figure}[h!]
    \centering
    \includegraphics[width=0.8\textwidth]{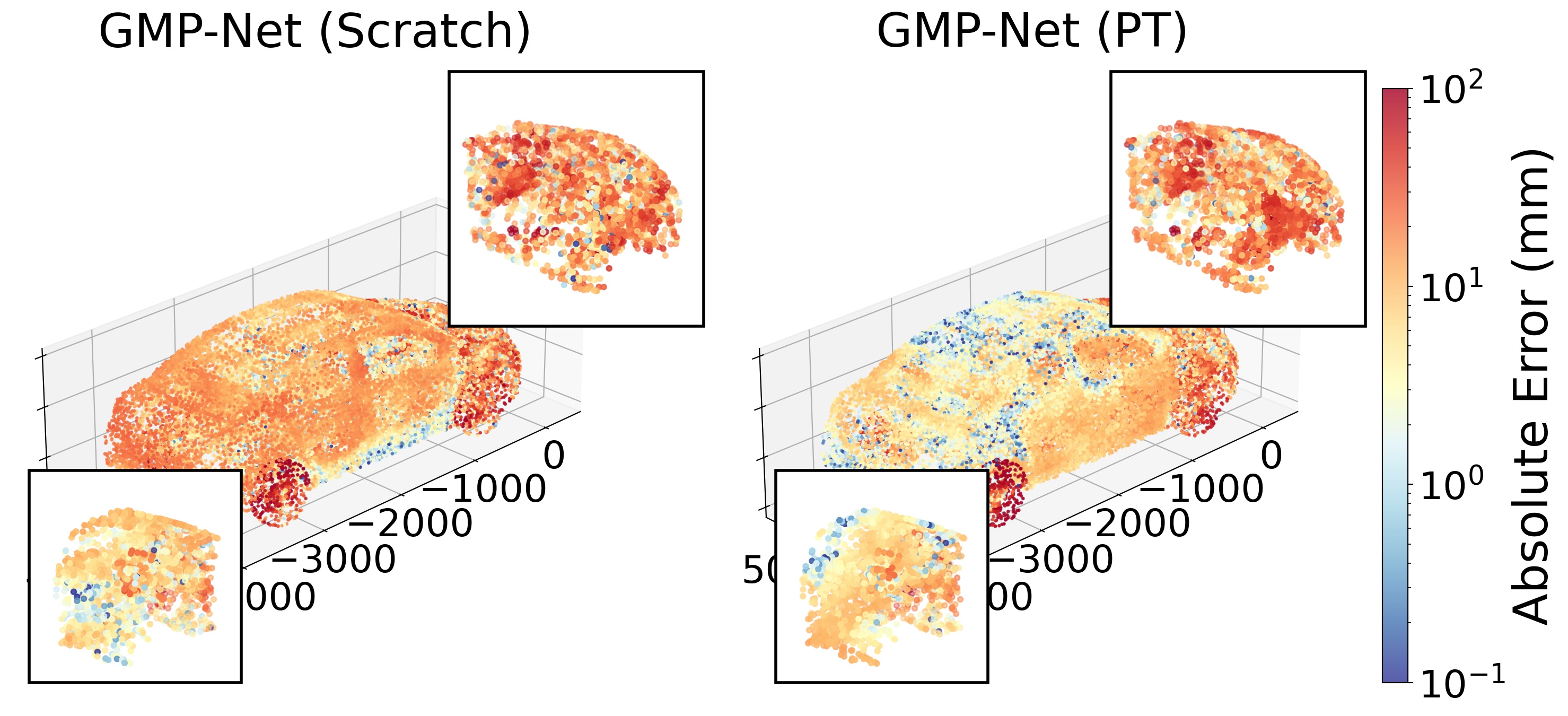} 
    \caption{SHIFT-Crash displacement absolute error comparison. 32 training samples for GMP-Net.}
    \label{fig:app_crash_8}
\end{figure}

\begin{figure}[h!]
    \centering
    \includegraphics[width=0.8\textwidth]{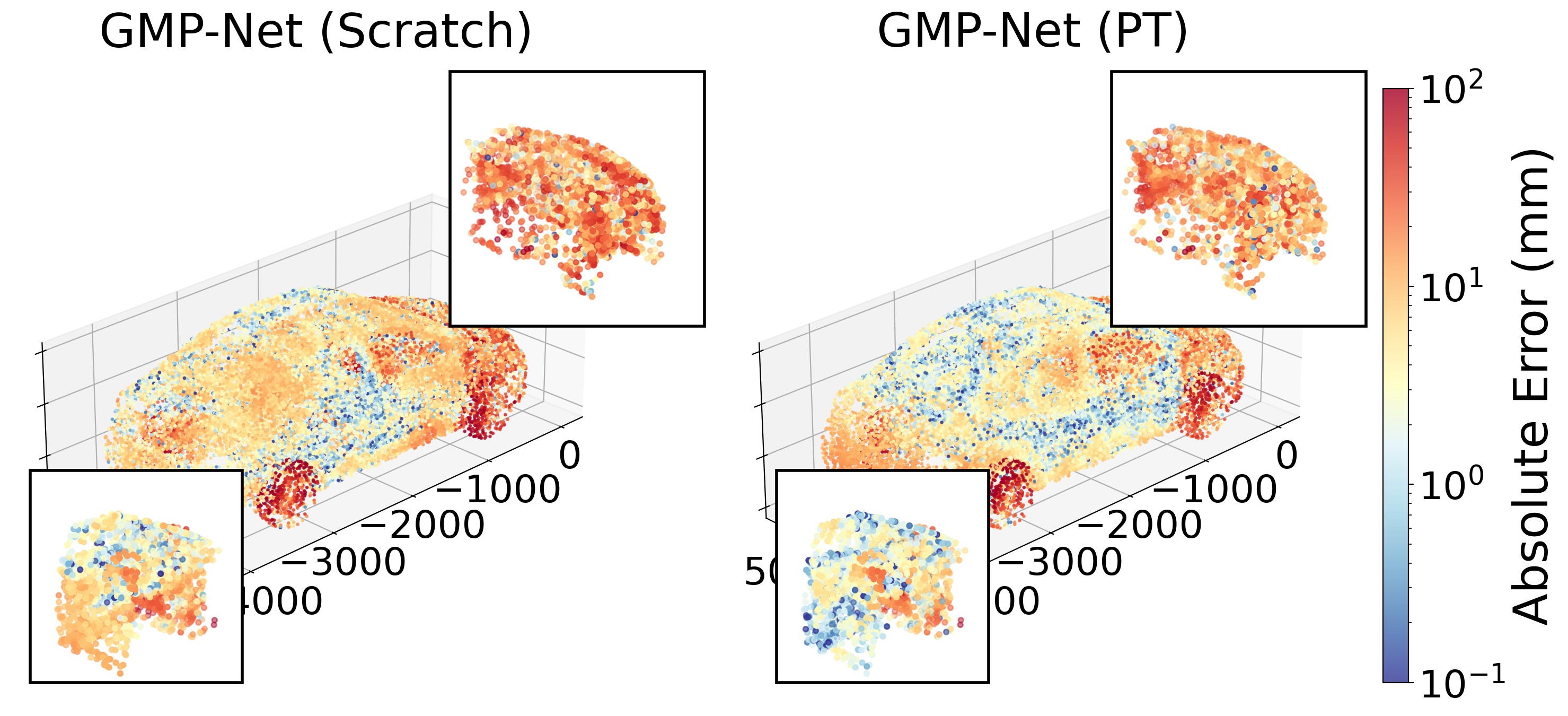} 
    \caption{SHIFT-Crash displacement absolute error comparison. 64 training samples for GMP-Net.}
    \label{fig:app_crash_9}
\end{figure}

\begin{figure}[h!]
    \centering
    \includegraphics[width=0.8\textwidth]{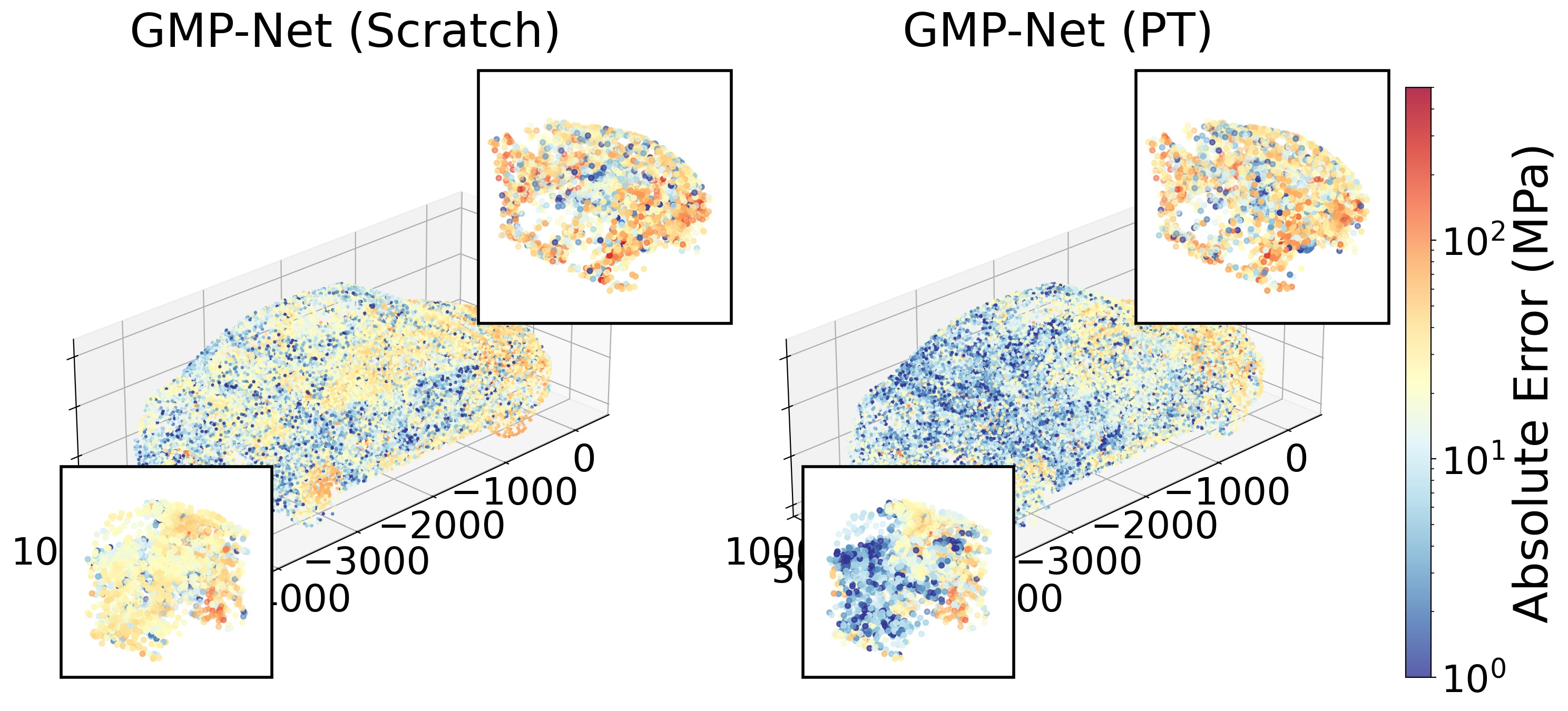} 
    \caption{SHIFT-Crash Von Mises stress absolute error comparison. 64 training samples for GMP-Net.}
    \label{fig:app_crash_10}
\end{figure}

\begin{figure}[h!]
    \centering
    \includegraphics[width=0.8\textwidth]{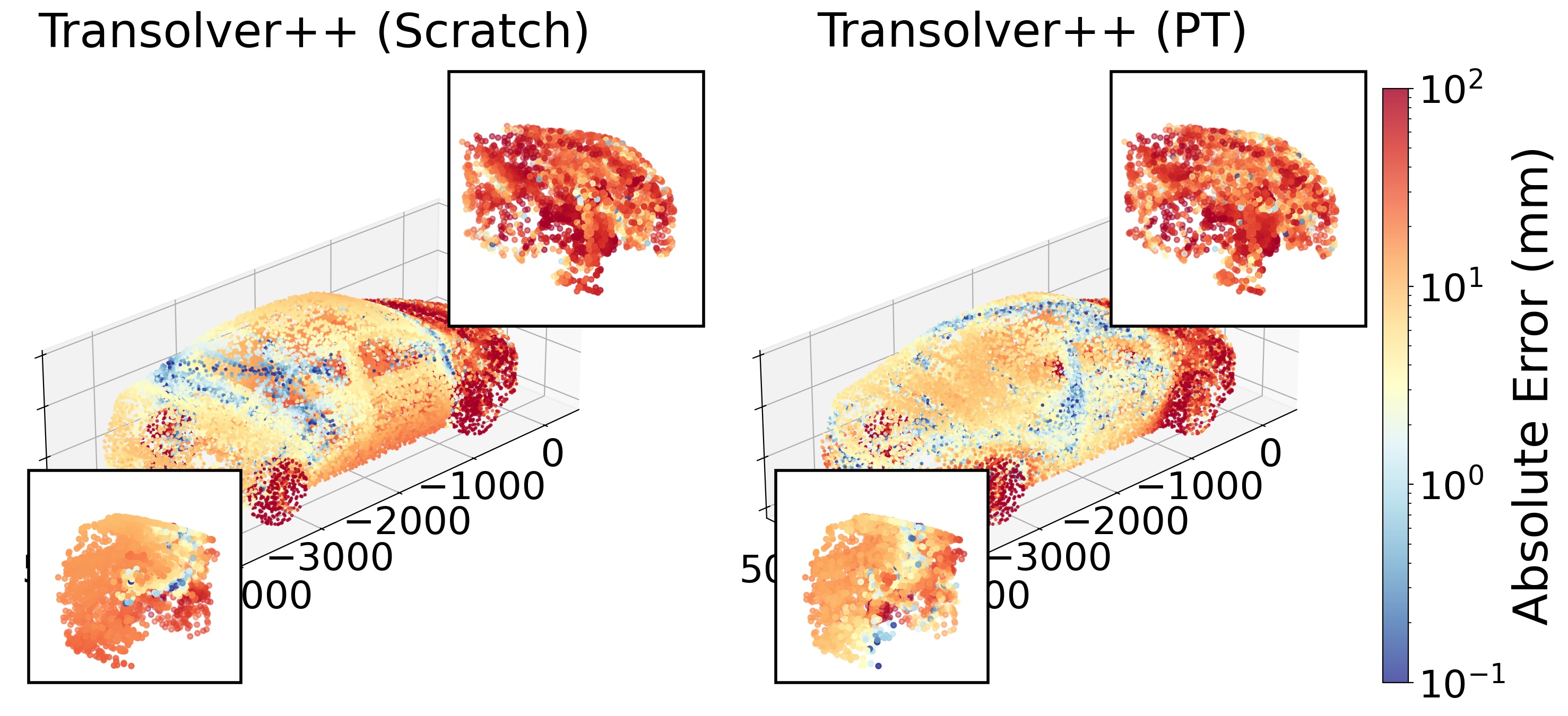} 
    \caption{SHIFT-Crash displacement absolute error comparison. 16 training samples for Transolver++.}
    \label{fig:app_crash_11}
\end{figure}

\begin{figure}[h!]
    \centering
    \includegraphics[width=0.8\textwidth]{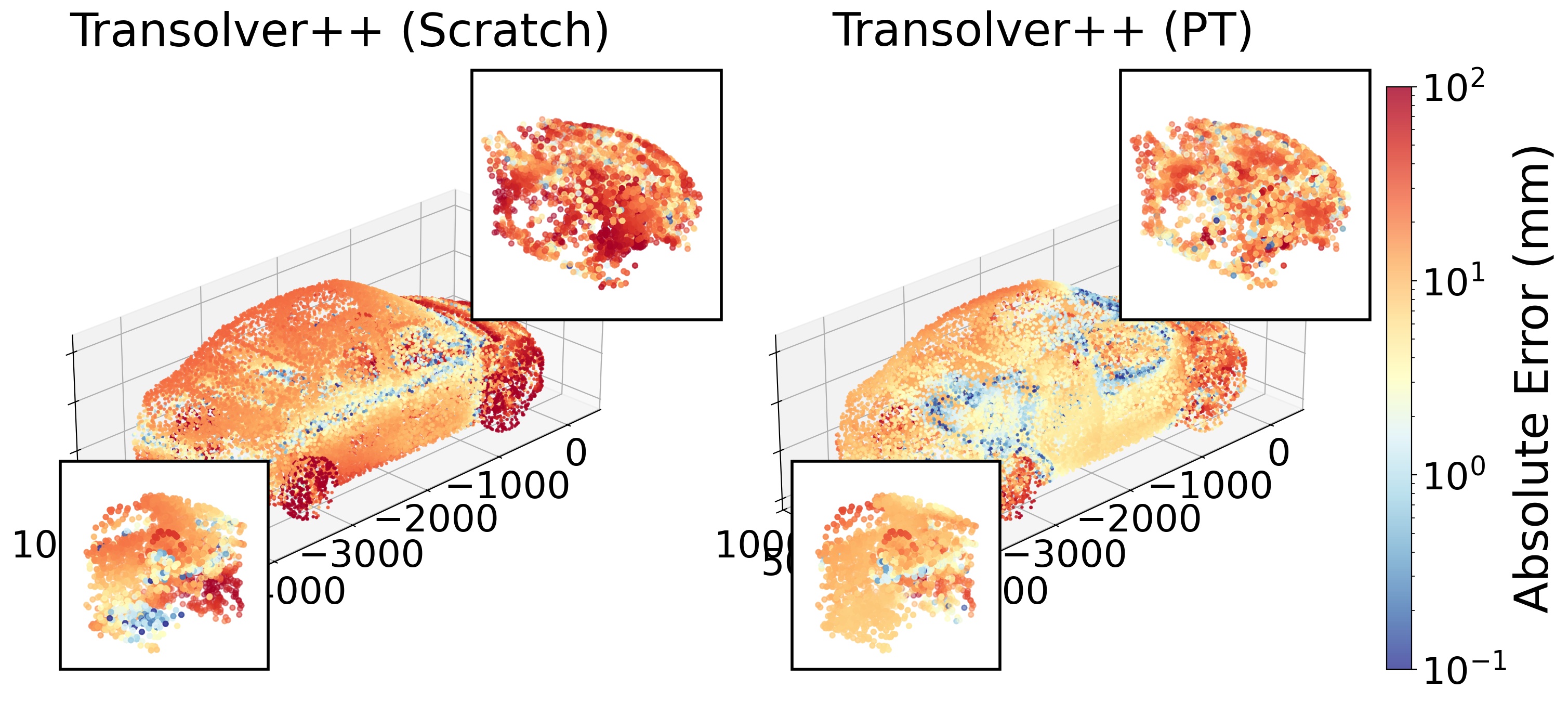} 
    \caption{SHIFT-Crash displacement absolute error comparison. 64 training samples for Transolver++.}
    \label{fig:app_crash_12}
\end{figure}

\begin{figure}[h!]
    \centering
    \includegraphics[width=0.8\textwidth]{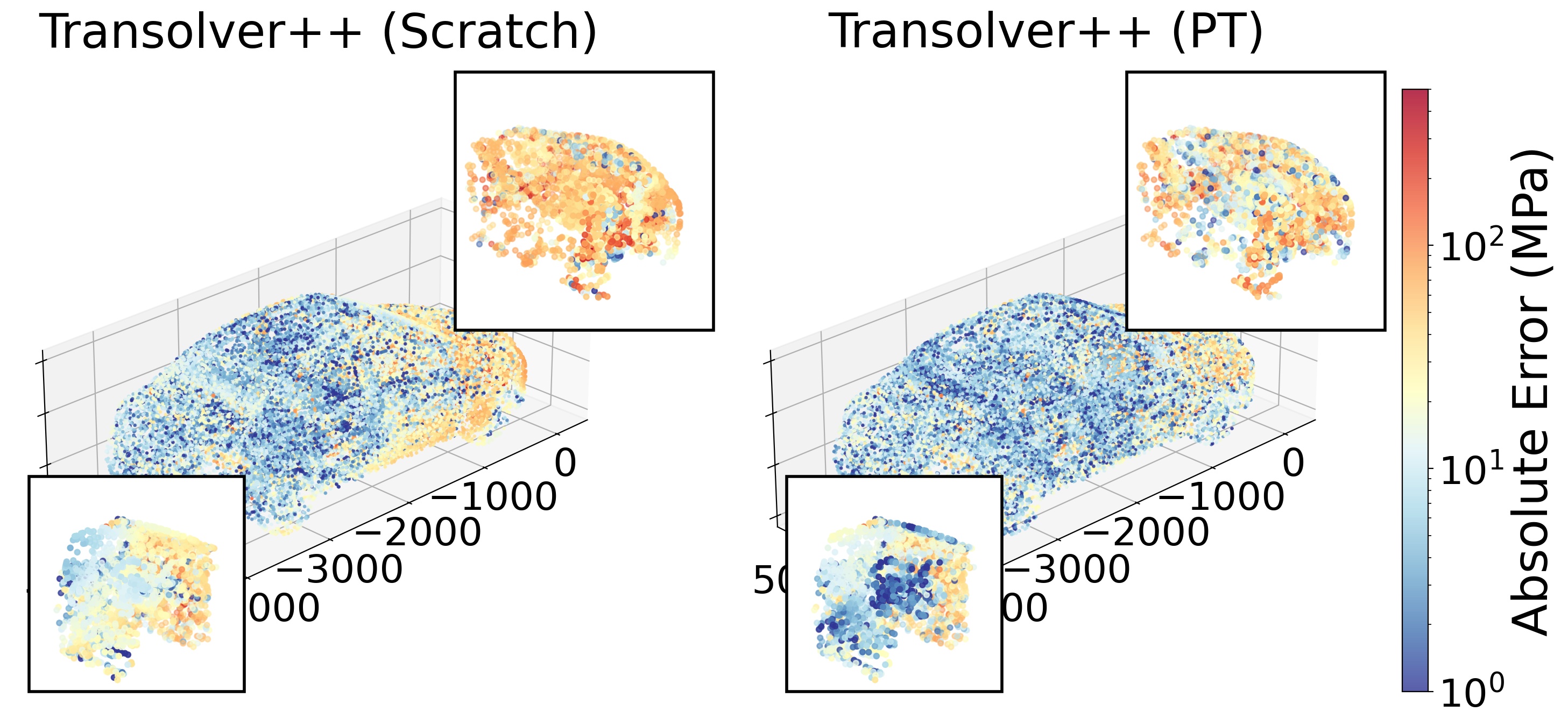} 
    \caption{SHIFT-Crash Von Mises stress absolute error comparison. 64 training samples for Transolver++.}
    \label{fig:app_crash_13}
\end{figure}

\begin{figure}[h!]
    \centering
    \includegraphics[width=0.8\textwidth]{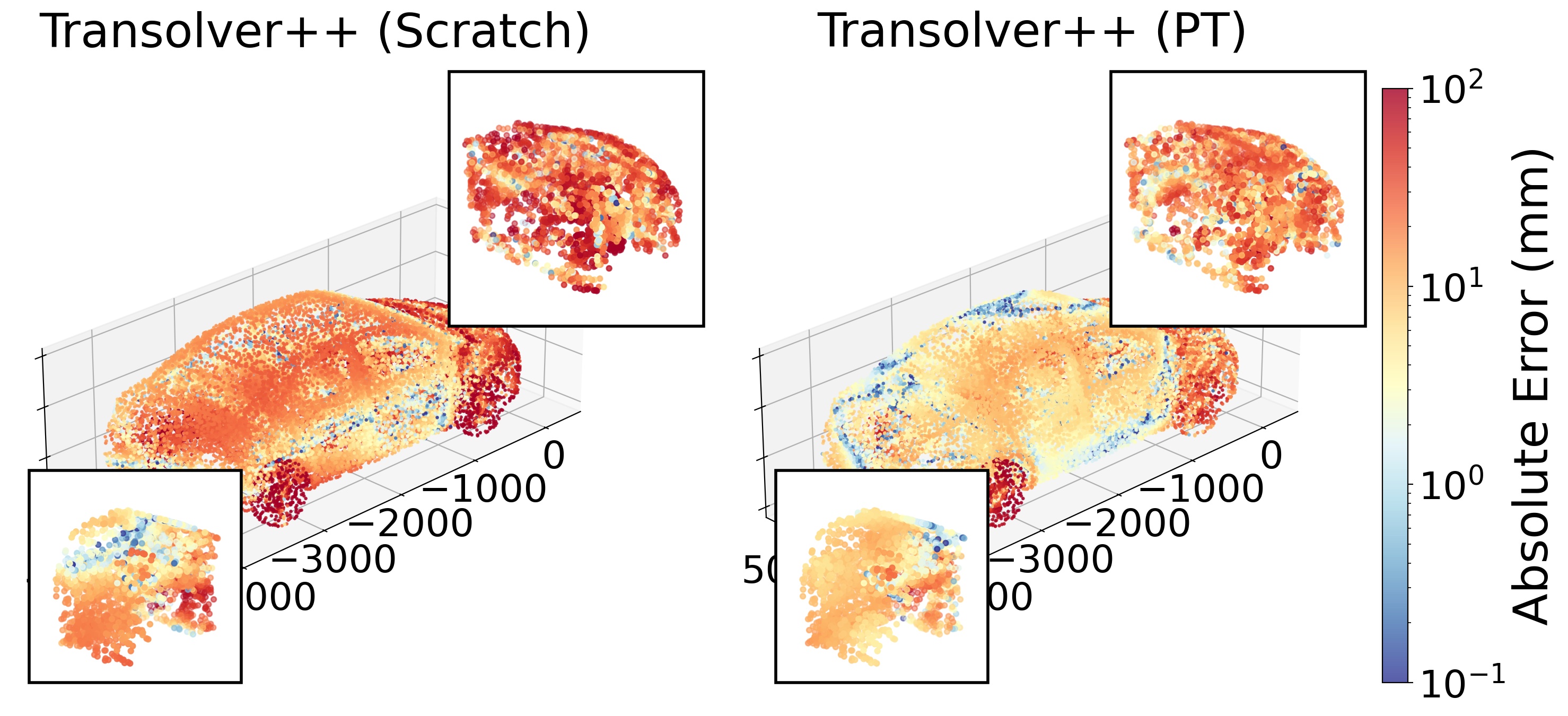} 
    \caption{SHIFT-Crash displacement absolute error comparison. 64 training samples for Transolver++.}
    \label{fig:app_crash_14}
\end{figure}

\begin{figure}[h!]
    \centering
    \includegraphics[width=0.8\textwidth]{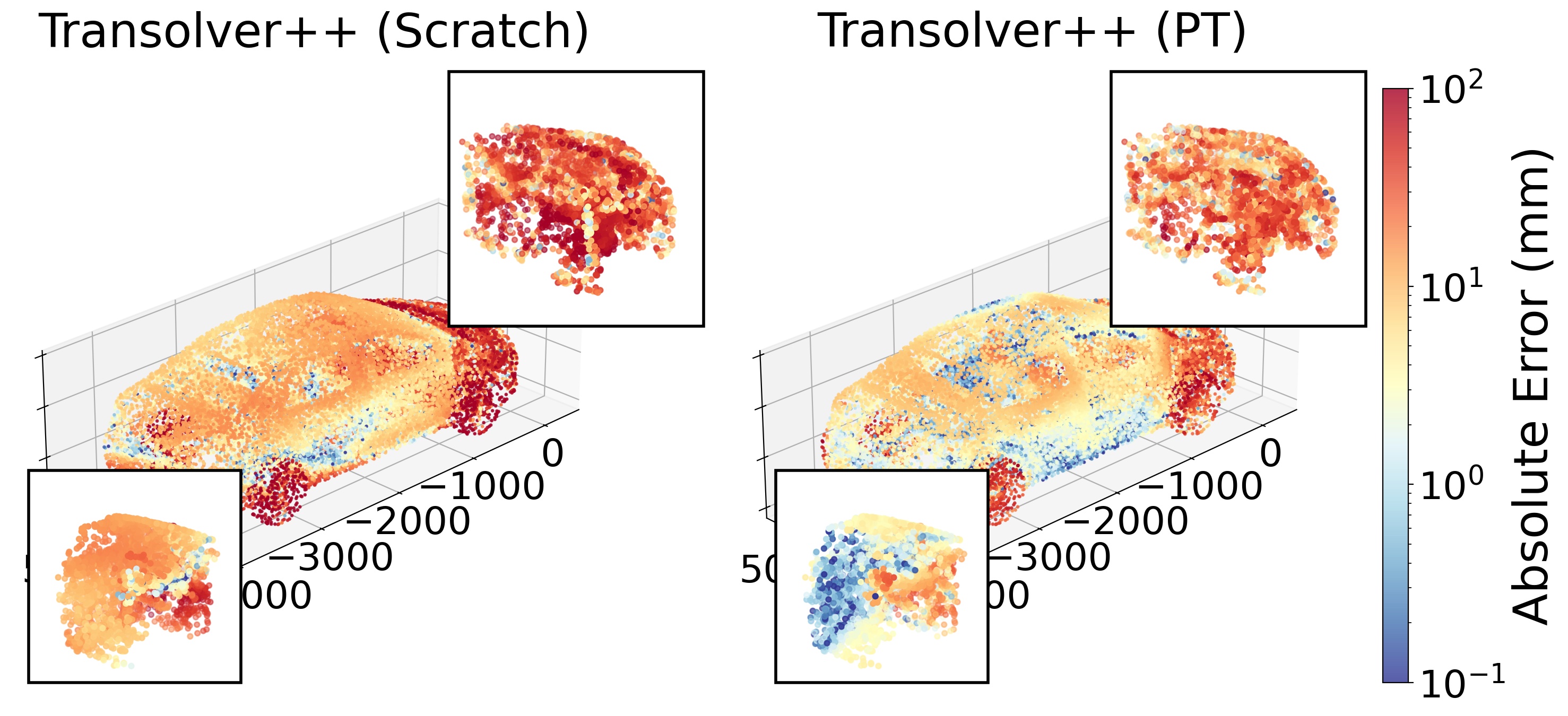} 
    \caption{SHIFT-Crash displacement absolute error comparison. 64 training samples for Transolver++.}
    \label{fig:app_crash_15}
\end{figure}

\begin{figure}[h!]
    \centering
    \includegraphics[width=0.8\textwidth]{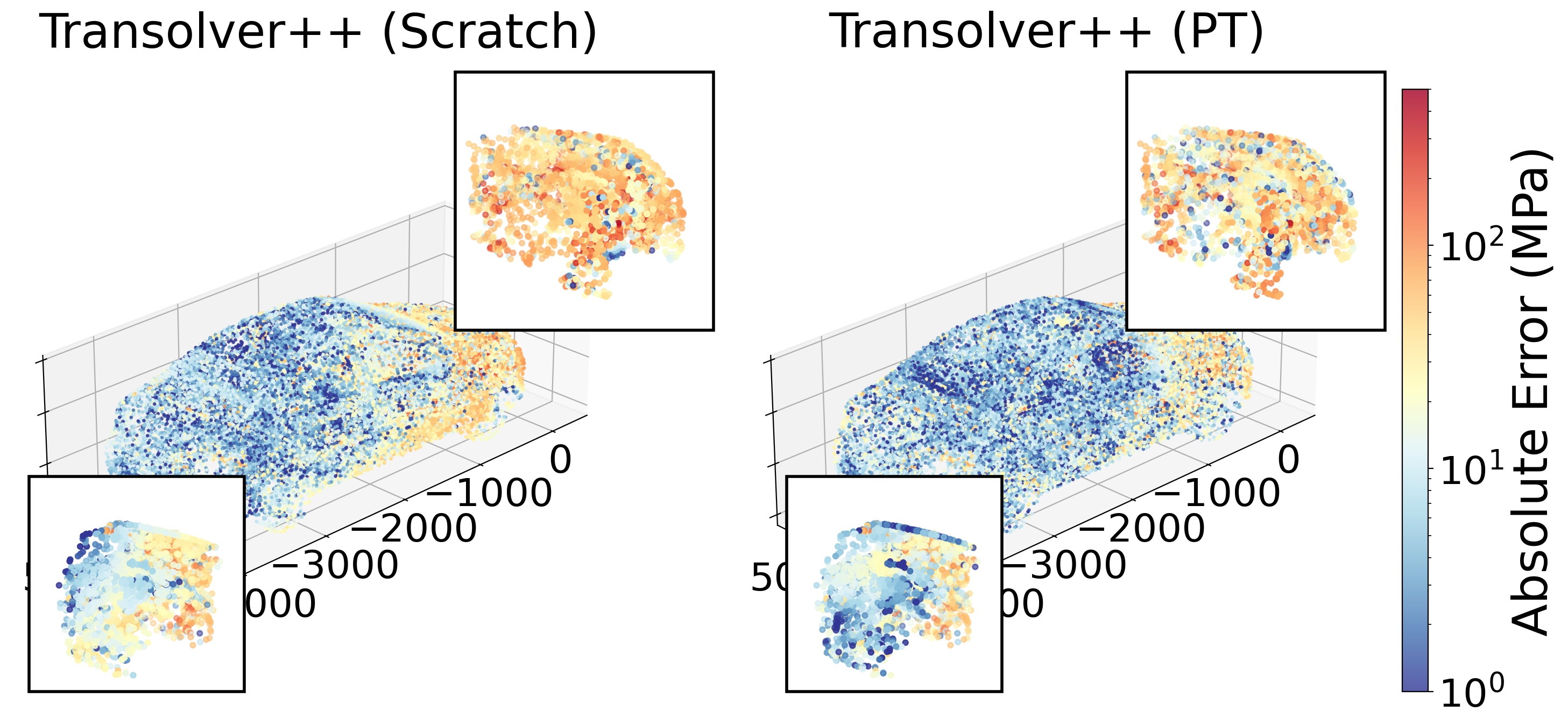} 
    \caption{SHIFT-Crash Von Mises stress absolute error comparison. 64 training samples for Transolver++.}
    \label{fig:app_crash_16}
\end{figure}

First, in Figures \ref{fig:app_drivaerml_smart}-\ref{fig:app_shiftcrash_transolver} we provide the fine-tuning loss evolutions for all the combinations of models and datasets.
Secondly, Figures \ref{fig:app_convergence} and \ref{fig:app_datasaving} show the improvement in convergence and data saving for all the previous combinations.
Finally, Figures \ref{fig:rollout_ks_error_app}-\ref{fig:app_crash_16} show visual results for steady and autoregressive experiments.

\clearpage

\clearpage

\end{document}